\documentclass[review]{elsarticle}

\usepackage{amsmath}
\usepackage{amsfonts}
\usepackage{amssymb}
\usepackage{amsthm}
\usepackage{bm}
\usepackage{indentfirst}
\usepackage{graphicx}
\usepackage{subfigure}
\usepackage{array}
\usepackage{fullpage}

\DeclareMathAlphabet{\mathsfsl}{OT1}{cmss}{m}{sl}

\newcommand{\PreserveBackslash}[1]{\let\temp=\\#1\let\\=\temp}
\newcolumntype{C}[1]{>{\PreserveBackslash\centering}p{#1}}
\newcolumntype{R}[1]{>{\PreserveBackslash\raggedleft}p{#1}}
\newcolumntype{L}[1]{>{\PreserveBackslash\raggedright}p{#1}}

\numberwithin{equation}{section}

\theoremstyle{definition}

\usepackage{anyfontsize}
\usepackage{enumerate}
\usepackage{esint}
\usepackage{etoolbox}
\usepackage{isomath}
\usepackage{mathrsfs}
\usepackage{mathtools}
\usepackage{multicol}
\usepackage{pgfplots}
\usepackage{scalerel}
\usepackage{stackengine}
\usepackage{tabulary}
\usepackage{tikz}

\makeatletter
\newcommand*\bdot{\mathpalette\bdot@{.65}}
\newcommand*\bdot@[2]{\mathbin{\vcenter{\hbox{\scalebox{#2}{$\m@th#1\bullet$}}}}}
\makeatother

\makeatletter
\newcommand*\bddot{\mathpalette\bddot@{.65}}
\newcommand*\bddot@[2]{\mathbin{\vcenter{\hbox{\scalebox{#2}
    {$\m@th#1\smash{{}_{\bullet}^{\bullet}}$}}}}}
\makeatother

\newcommand{\circled}[2][]{%
  \tikz[baseline=(char.base)]{%
    \node[shape = circle, draw, inner sep = .5pt]
    (char) {\phantom{\ifblank{#1}{#2}{#1}}};%
    \node at (char.center) {\makebox[0pt][c]{#2}};}}
\robustify{\circled}

\makeatletter
\newcommand{\opnorm}{\@ifstar\@opnorms\@opnorm}
\newcommand{\@opnorms}[1]{%
  \left|\mkern-1.5mu\left|\mkern-1.5mu\left|
   #1
  \right|\mkern-1.5mu\right|\mkern-1.5mu\right|
}
\newcommand{\@opnorm}[2][]{%
  \mathopen{#1|\mkern-1.5mu#1|\mkern-1.5mu#1|}
  #2
  \mathclose{#1|\mkern-1.5mu#1|\mkern-1.5mu#1|}
}
\makeatother

\stackMath
\newcommand\reallywidecheck[1]{%
\savestack{\tmpbox}{\stretchto{%
  \scaleto{%
    \scalerel*[\widthof{\ensuremath{#1}}]{\kern-.6pt\bigwedge\kern-.6pt}%
    {\rule[-\textheight/2]{1ex}{\textheight}}%WIDTH-LIMITED BIG WEDGE
  }{\textheight}%
}{0.5ex}}%
\stackon[1pt]{#1}{\scalebox{-1}{\tmpbox}}%
}

\usepackage[utf8]{inputenc}
\usepackage{bm}
\usepackage{mathrsfs}
\usepackage{amsmath}
\usepackage{amsfonts}
\usepackage{amsthm}
\usepackage{algorithm}
\usepackage[noend]{algpseudocode}
\usepackage{color}
\usepackage{setspace}
\usepackage{float}
\usepackage{import}
\usepackage{url}
\usepackage{multicol}
\usepackage{multirow}
\usepackage{subfigure}
\biboptions{sort&compress}
\usepackage[top=1in,bottom=1in,left=1in,right=1in]{geometry}
\newcommand{\real}{\mathbb{R}}
\newcommand{\complex}{\mathbb{C}}

\newcommand{\mcK}{\mathcal{K}}

\newcommand{\mcD}{\mathcal{D}}

\newcommand{\mcG}{\mathcal{G}}

\newcommand{\mcF}{\mathcal{F}}
\newcommand{\mcU}{\mathcal{U}}
\newcommand{\mcH}{\mathcal{H}}

\newcommand{\mcN}{\mathcal{N}}

\newcommand{\mcR}{\mathcal{R}}
\newcommand{\mcL}{\mathcal{L}}

\newcommand{\mbR}{\mathbb{R}}

\newcommand{\mbRd}{{\mathbb{R}^d}}

\usepackage{xcolor}

\def\omg{{\Omega}}

\def\omgbb{\mathcal{B}\mathcal{B}\Omega}

\def \bb{\bm{b}}

\def \Gb{\bm{G}}
\def \fb{\bm{f}}

\def \Fb{\bm{F}}
\def \gb{\bm{g}}
\def \ub{\bm{u}}
\def \Ub{\bm{U}}
\def \wb{\bm{w}}
\def \vb{\bm{v}}
\def \xb{\bm{x}}

\def \hb{\bm{h}}
\def \Hb{\bm{H}}

\def \pb{\bm{p}}

\def \tb{\bm{t}}
\def \rb{\bm{r}}

\def \qb{\bm{q}}

\def \yb{\bm{y}}
\def \cb{\bm{c}}
\def \Cb{\bm{C}}

\def \Xb{\bm{X}}

\newcommand{\vertii}[1]{{\left\vert\left\vert #1
    \right\vert\right\vert}}

\newtheorem{theorem}{Theorem}

\newtheorem{lemma}{Lemma}

\newtheorem{assumption}{Assumption}

\begin{document}

\begin{frontmatter}

\title{Learning Deep Implicit Fourier Neural Operators (IFNOs) with \\ Applications to Heterogeneous Material Modeling}

\address[yy]{Department of Mathematics, Lehigh University, Bethlehem, PA 18015, USA}
\address[cl]{School of Aerospace and Mechanical Engineering, The University of Oklahoma, Norman, OK 73019, USA}

\author[yy]{Huaiqian You}\ead{huy316@lehigh.edu}
\author[yy]{Quinn Zhang}\ead{quz222@lehigh.edu}
\author[cl]{Colton J. Ross}\ead{cjross@ou.edu}
\author[cl]{Chung-Hao Lee}\ead{ch.lee@ou.edu}
\author[yy]{Yue Yu\corref{cor1}}\ead{yuy214@lehigh.edu}
\cortext[cor1]{Corresponding author}

%\date{January 2020}

\begin{abstract}
Constitutive modeling based on continuum mechanics theory has been a classical approach for modeling the mechanical responses of materials. However, when constitutive laws are unknown or when defects and/or high degrees of heterogeneity are present, these classical models may become inaccurate. In this work, we propose to use data-driven modeling, which directly utilizes high-fidelity simulation and/or experimental measurements to predict a material's response without using conventional constitutive models. Specifically, the material response is modeled by learning the implicit mappings between loading conditions and the resultant displacement and/or damage fields, with the neural network serving as a surrogate for a solution operator. To model the complex responses due to material heterogeneity and defects, we develop a novel deep neural operator architecture, which we coin as the Implicit Fourier Neural Operator (IFNO). In the IFNO, the increment between layers is modeled as an integral operator to capture the long-range dependencies in the feature space. As the network gets deeper, the limit of IFNO becomes a fixed point equation that yields an implicit neural operator and naturally mimics the displacement/damage fields solving procedure in material modeling problems. 
To obtain an efficient implementation, we parameterize the integral kernel of this integral operator directly in the Fourier space and interpret the network as discretized integral (nonlocal) differential equations, which consequently allow for the fast Fourier transformation (FFT) and accelerated learning techniques for deep networks. 
%By identifying layers with time instants, IFNO can be interpreted as discretized integral (nonlocal) differential equations, and consequently allows for accelerated learning techniques for deep networks. 
%the nonlocal kernel network (NKN) is employed, which is resolution-independent and naturally embeds the material micromechanical properties and defects in the integrand. 
We demonstrate the performance of our proposed method for a number of examples, including hyperelastic, anisotropic and brittle materials. As an application, we further employ the proposed approach to learn the material models directly from digital image correlation (DIC) tracking measurements, and show that the learned solution operators substantially outperform the conventional constitutive models in predicting displacement fields.
\end{abstract}


\end{frontmatter}

%\begin{document}
%\maketitle

\tableofcontents

\section{Introduction}

In science and engineering, predicting and monitoring heterogeneous material responses are ubiquitous in many applications \cite{zohdi2002toughening,wriggers1998computational,kok2018anisotropy,bostanabad2018computational,su2006guided,AFOSR2014,talreja2015modeling,soric2018multiscale,pijaudier2013damage,mourlas2019accurate,markou2021new}. In these materials, the microstructure, in terms of the geometric distribution of phases, constituent properties, and interfacial bonding attributes influences the deformation and failure behavior, which needs to be accurately captured to guarantee reliable and trustworthy predictions and inform decision making. Conventionally, constitutive models based on continuum mechanics have been commonly employed for modeling heterogeneous material responses. When the material microstructures are known, constitutive models in conjunction with other field equations (e.g., balance of linear momentum) are often built in the form of partial differential equations (PDEs), and the material responses are obtained by approximating the PDE solutions with classical numerical methods such as finite elements.

However, fundamental challenges are still present in utilizing the constitutive models and numerical simulations to provide a comprehensive physical and functional description of heterogeneous material responses \cite{lindgren2016us}. First, in the constitutive modeling theory the choice of governing laws (such as the strain energy density function) is often determined \textit{a priori} and the free parameters are often tuned to obtain agreement with experimental stress-strain data. This fact makes the rigorous calibration and validation process challenging. Second, although new experimental technologies and testing procedures have been designed to observe much smaller microstructure patterns and monitor defects in a faster manner \cite{bostanabad2018computational,lindgren2013state,hdbk2009nondestructive,achenbach2000quantitative,jones2015probing,pan2018review,shukla2020physics}, it remains difficult to fully quantify the microstructure and responses for individual material samples, due to variability and measurement noises from different microstructure geometries, properties, and operating environments. In addition to these challenges, many microstructure characterization methods require the use of destructive methods that could alter the observed microstructural properties, such as optical clearing and histological processing \cite{misfeld2007heart,rieppo2008practical}. Therefore, in the application scenarios where the material response of a particular material sample is of interests, such as the non-destructive evaluation and damage prediction problems, conventional constitutive models may suffer from errors stemmed from its functional form assumption and the measurement noises, leading to limited predictivity.

To address these challenges, data-driven computing has been considered as an alternative to the conventional constitutive modeling. In recent years, there has been significant progress in the development of deep neural networks (NNs), focusing on learning the hidden physics of a complex system \cite{ghaboussi1998autoprogressive,ghaboussi1991knowledge,carleo2019machine,karniadakis2021physics,zhang2018deep,cai2022physics,pfau2020ab,he2021manifold,besnard2006finite,ibanez2017data,ibanez2018manifold,stainier2019model,kirchdoerfer2016data,heider2020so,fuhg2021physics}. Among these works, several studies that use neural networks 
%in conjunction with standard numerical methods 
in modeling heterogeneous materials have been conducted \cite{wang2018multiscale,he2020physics,tartakovsky2020physics,liu2019deep,yang2019derivation,garbrecht2021interpretable}. In \cite{he2020physics,tartakovsky2020physics}, physics-informed NN models \cite{raissi2019physics} were developed, where the material responses were modeled as the solution of a {\it known} PDE by a deep NN with weights and biases learned together with the PDE's unknown parameter fields (e.g., permeability). In \cite{garbrecht2021interpretable}, a symbolic regression method \cite{bongard2007automated,schmidt2009distilling,udrescu2020ai,bomarito2021development} was developed to learn the microstructure-dependent plasticity from data, where the constitutive models were generated using interpretable machine learning as symbolic expressions. In \cite{wang2018multiscale}, data-driven approaches were employed for the homogenization procedure, where the information from multiple sub-scales can be used to sequentially generate the macroscopic prediction in a cost-efficient manner. In \cite{yang2019derivation,liu2019deep}, representative volume elements (RVE) were employed to build the material law for heterogeneous materials, and a homogenized model was then discovered based on the RVE database. To the authors' best knowledge, most of the state-of-the-art NN developments for heterogeneous material modeling either focus on the homogenized behavior of the material or rely on (partially) known physics laws, which limits their applicability to problems where the unknown heterogeneous behavior of each individual sample is of interest.

More recently, the use of NNs has been extended to learning maps between inputs of a dynamical system and its state, so that the network serves as a surrogate for a solution operator \cite{lu2019deeponet,lu2021learning,li2020neural,li2020multipole,li2020fourier}. This approach, which can be referred as {\it neural operators}, finds applicability when the constitutive laws are unknown. Representative works in this direction include the integral neural operator architectures \cite{li2020neural,li2020multipole,li2020fourier,you2022nonlocal,Ong2022,gupta2021multiwaveletbased} and the DeepONet architectures \cite{lu2019deeponet,lu2021learning,goswami2022physics}. Comparing with the classical NNs, the most notable advantages of neural operators are resolution independence and generalizability to different input instances. The former implies that the accuracy of the prediction is invariant with respect to the resolution of input parameters such as loading conditions and material properties. This fact is in stark contrast with the classical finite-dimensional approaches that build the NN models between finite-dimensional Euclidean spaces, so that their accuracy is tied to the resolution of input \cite{guo2016convolutional,zhu2018bayesian,adler2017solving,bhatnagar2019prediction,khoo2021solving}. Furthermore, being generalizable with respect to different input parameter instances renders another computing advantage: once the neural operator is trained, solving for a new instance of the input parameter only requires a forward pass of the network. This unique property is in contrast with traditional PDE-constrained optimization techniques \cite{de2015numerical} and some other NN models that directly parameterize the solution \cite{raissi2019physics,weinan2018deep,bar2019unsupervised,smith2020eikonet,pan2020physics}, as all these methods only approximate the solution for a {\it single instance of the input}. In \cite{yin2022simulating,goswami2022physics,yin2022interfacing}, neural operators have been successfully applied to model the unknown physics law of homogeneous materials. In \cite{li2020neural,li2020multipole,li2020fourier,lu2021comprehensive}, neural operators are employed as a solution surrogate for the Darcy's flow in a heterogeneous porous medium, when the microstructure field is known.

In this work, we propose to advance the current data-driven methods on heterogeneous material modeling by designing deep neural operators to model heterogeneous material responses without using any predefined constitutive models or microstructure measurements. Specifically, through learning the solution operator directly from high-fidelity simulation and/or experimental measurements, we integrate material identification, modeling procedures, and material response prediction. The material microstructure properties are learned implicitly from the data and naturally embedded in the network parameters. The heterogeneous material responses can thus be obtained without assumptions on microstructure or governing laws. To capture the complex and possibly nonlinear material responses, deep NNs are necessary to learn multiple levels of abstraction for representations of the raw input data. To achieve this goal, we pursue a new integral neural operator architecture that, 1) is stable in the limit of deep layers with fixed memory costs, 2) has guaranteed universal approximation capability, and 3) is independent of the input resolution and generalizable to unseen input function instances. Our proposed architecture can be interpreted as a data-driven surrogate of the fixed point procedure, in the sense that the increment of fixed point iterations are modeled as increment between layers. As such, a forward pass through a very deep network is analogous to obtaining the PDE solution as an implicit problem, and the universal approximation capability is guaranteed as far as there exists a convergent fixed point equation\footnote{Here, we point out that the idea of using constant parameters across layers and formulating the NNs as a fixed point equation was also proposed in implicit networks \cite{el2021implicit,bai2019deep,winston2020monotone,bai2020multiscale,fung2021jfb} such that the deep network can be trained with fixed memory costs.}. To further accelerate the learning, we identify iterative layers with time instants such that the proposed network can be interpreted as discretized autonomous integral (non-local) differential equations, and consequently allows for the shallow-to-deep initialization technique \cite{haber2018learning,modersitzki2009fair,you2022nonlocal} where optimal parameters learned on shallow networks are considered as (quasi-optimal) initial guesses for deeper networks. Since the proposed architecture is built as a modification of the Fourier Neural Operator method (FNO), it also parameterizes the integral kernel directly in the Fourier space and utilizes the fast Fourier transformation (FFT) to efficiently evaluate the integral operator. As such, our network inherits the advantages of FNOs on \textit{resolution independence} and \textit{superior efficiency}. Because it preserves the similar properties to both the implicit neural networks and the FNOs, we refer to our proposed network as implicit Fourier neural operators (IFNOs).

We summarize our major contributions as follows.
\begin{enumerate}
    \item We introduce a novel deep neural operator by parameterizing the layer increment as an integral operator, referred to as IFNO, which learns the mapping between loading conditions and material responses as a solution operator while preserving the accuracy across resolutions.
    \item By resembling the network architecture as a fixed point method, the IFNOs can be interpreted as a numerical solver for an implicit problem with unknown material properties/microstructure, and the universal approximation property is guaranteed as far as there exists a converging fixed point equation for this implicit problem.
    \item By identifying the layers with time instants, the IFNOs can also be interpreted as discretized nonlocal time-dependent equations, which allows for accelerated learning techniques for deep networks, such as the shallow-to-deep technique \cite{haber2018learning}.
    \item In a variety of complex material response learning tasks, the IFNOs demonstrate not only stability but also improved accuracy in the deep network limit: in complex learning tasks, the IFNOs outperform the best FNOs with reduced memory costs and halved prediction errors.
    \item Our proposed method integrates material identification, modeling procedures, and material response prediction into one learning framework, which makes it particularly promising for learning complex material responses without explicit constitutive models and/or microstructure measurements. To demonstrate this capability, we learn the mechanical responses of a latex glove sample directly from digital image correlation (DIC) tracking measurements. Comparing with the conventional constitutive models, our method reduces the prediction error by 10 times.
\end{enumerate}

The remainder of this paper is organized as follows. In Section \ref{sec:background}, we introduce three integral neural operator architectures that inspired our work and highlight their advantages and limitations. In Section \ref{sec:ifno}, we introduce the IFNOs as inspired by an implicit problem solver, and discuss its universal approximation capability. In Section \ref{sec:experiments}, we show the stability and convergence of the IFNOs for a number of benchmarks, including heterogeneous, hyperelastic, anisotropic and brittle fracture material problems, that illustrate the efficacy of our network compared to the baseline networks. Next, in Section \ref{sec:dic} we further demonstrate the applicability of our data-driven approach to learn the unknown mechanical responses directly from DIC tracking measurements, providing evidence that the scheme yields accurate predictions for practical engineering problems. In Section \ref{sec:conclusion}, we provide a summary of our achievements and concluding remarks. In the appendix, we provide 
%an overview of peridynamics -- the high-fidelity model we used to generate the brittle fracture material dataset, and 
additional numerical results.

\section{Background and Related Work}\label{sec:background}

This section provides the necessary background for the rest of the paper by formally stating the problem of neural operator learning, providing succinct reviews on the three integral neural operator learning approaches recently proposed in the literature that inspired the proposed IFNOs,  and highlighting their properties, as summarized in Table \ref{tab:comparison}.

\begin{table}
\begin{center}
{\small\begin{tabular}{ c | c | c | c | c | c }
\hline
 Model & Layer-Independent & Efficiency&Continuous in& Stability in &Ref\\
       & Parameters  & Through FFT&  Depth (Time) & Deep Networks&\\ \hline
GKN & \checkmark & -- & -- & -- & \cite{li2020neural,li2020multipole}\\
NKN & \checkmark & -- & \checkmark & \checkmark & \cite{you2022nonlocal} \\ 
FNO & -- & \checkmark & -- & -- & \cite{li2020fourier}\\
\hline
IFNO & \checkmark & \checkmark & \checkmark & \checkmark &  \\ \hline
\end{tabular}}
\end{center}
\caption{List of the properties for the graph kernel networks (GKNs), nonlocal kernel networks (NKNs), Fourier neural operators (FNOs), and the proposed implicit Fourier neural operators (IFNOs).}
\label{tab:comparison}
\end{table}

%%%%%%%%%%%%%%%%%%%%%%%%%%%%%%%%%%%%%%%%%%%%%%%%%
\subsection{Problem statement: Learning solution operators}

The main application considered in this work is the modeling of complex material responses under different loading conditions. Formally, consider a $s$-dimensional body occupying the domain $\omg\subset\mathbb{R}^s$ ($s=1,2$ or $3$), which deforms under external loading. Without prior knowledge of the material properties or constitutive laws, our ultimate goal is to identify the {\it best surrogate solution operator}, that accurately predicts the material mechanical responses in terms of the resultant displacement field $\ub(\xb)$ and/or damage field given new and unseen material property or loading scenarios. In this context, different types of loading scenarios are considered, such as a displacement-type loading applied on the subject's boundary, a body force applied on the whole domain $\Omega$, a traction loading applied on part of its boundaries or a combination of the above. Denoting the whole boundaries of domain $\Omega$ as $\partial \Omega$, we consider general mixed boundary conditions: $\partial\Omega=\partial\Omega_D\bigcup \partial\Omega_N$ %
and $(\partial\Omega_D)^o\bigcap (\partial\Omega_N)^o=\emptyset$, where $\partial \Omega_D$ and $\partial \Omega_N$ are the Dirichlet and Neumann boundaries, respectively. To apply the displacement-type loading on the boundary, we assume that $\ub(\xb)=\ub_D(\xb)$ are provided on $\partial\Omega_D$, while the traction $\tb(\xb)$ is applied on the boundary $\partial\Omega_N$.

In this work, we propose to learn the surrogate solution operator as a mapping between functions, namely, the microstructure and/or loading and the resultant displacement/damage field, given a collection of observed function pairs. Mathematically, let $\mcK_{\bb}$ be the unknown differential operator associated with the momentum balance equation and $\mcN_{\bb}$ be the unknown operator associated with the traction, both depending on the material microstructure parameter field $\bb(\xb)$. Given a body force $\gb(\xb)$, the momentum balance equation and boundary conditions write:
\begin{equation}\label{eqn:pde}
\begin{aligned} 
\mcK_{\bb}[\ub](\xb)=\gb(\xb),\quad&\xb\in \omg,\\
\ub(\xb)=\ub_D(\xb),\quad&\xb\in\partial \omg_D,\\
\mcN_{\bb}[\ub](\xb)=\tb(\xb),\quad&\xb\in\partial \omg_N.\\
\end{aligned}
\end{equation}
To solve the displacement field, we consider the problem of learning a general solution operator, with its input being a concatenated vector function $\fb(\xb)$ of $\xb$, $\bb(\xb)$, $\gb(\xb)$, $\ub_D(\xb)$, $\tb(\xb)$ and its output being the displacement field $\ub(\xb)$, for all $\xb\in\omg$. Here, we notice that $\ub_D(\xb)$ and $\tb(\xb)$ are only defined on the displacement boundary $\partial\omg_D$ and the traction boundary $\partial\omg_N$, respectively. To make them well-defined on the whole domain, we employ the zero-padding strategy proposed in \cite{lu2021comprehensive}, namely, we define $\fb(\xb):=[\xb,\bb(\xb),\gb(\xb),\tilde{\ub}_D(\xb),\tilde{\tb}(\xb)]$ where
\begin{equation}\label{eqn:pad}
\tilde{\ub}_D(\xb)=\left\{\begin{array}{cc}
    \ub_D(\xb), & \text{ if }\xb\in\partial\omg_D \\
    0, &  \text{ if }\xb\in\omg\backslash\partial\omg_D
\end{array}\right.,\qquad \tilde{\tb}(\xb)=\left\{\begin{array}{cc}
    \tb(\xb), & \text{ if }\xb\in\partial\omg_N \\
    0, &  \text{ if }\xb\in\omg\backslash\partial\omg_N
\end{array}\right..    
\end{equation}
In what follows, we denote the input and output function spaces as $\mcF=\mcF(\omg;\real^{d_F})$ and $\mcU=\mcU(\omg;\real^{d_u})$, respectively. Let $\{\fb_j,\ub_j\}_{j=1}^N$ be a set of observations where the input $\{\fb_j\}\subset\mcF$ is a sequence of independent and identically distributed random fields from a known probability distribution $\mu$ on $\mcF$, and $\mcG^\dag[\fb_j](\xb)=\ub_j(\xb)\in\mcU$, possibly noisy, is the output of the solution map $\mcG^\dag:\mcF\to\mcU$. With neural operator learning, we aim to build an approximation of $\mcG^\dag$ by constructing a nonlinear parametric map
$$
\mcG[\cdot\,;\,\theta]:\mcF\times\Theta\rightarrow\mcU,
$$
in the form of a neural network (NN), for some finite-dimensional parameter space $\Theta$. Here, $\theta\in\Theta$ is the set of parameters in the network architecture to be inferred by solving the following minimization problem
\begin{equation}\label{eqn:opt}
\min_{\theta\in\Theta}\mathbb{E}_{\fb\sim\mu}[C(\mcG[\fb;\theta],\mcG^\dag[\fb])]\approx \min_{\theta\in\Theta}\sum_{j=1}^N[C(\mcG[\fb_j;\theta],\ub_j)],
\end{equation}
where $C$ denotes a properly defined cost functional $C:\mcU\times\mcU\rightarrow\real$. Although $\fb_j$ and $\ub_j$ are (vector) functions defined on a continuum, with the purpose of doing numerical simulations, we assume that they are defined on a discretization of the domain defined as $\chi=\{\xb_1,\cdots,\xb_M\}\subset \omg$. With such a discretization to establish learning governing laws, a popular choice of the cost functional $C$ is the mean square error, i.e.,  
%the difference between $G(\fb_j;\theta)$ and $\ub_j$ in the $l^2$ norm defined on $D_j$:
$$C(\mcG[\fb_j;\theta],\ub_j):=\sum_{\xb_i\in \chi}\vertii{\mcG[\fb_j;\theta](\xb_i)-\ub_j(\xb_i)}^2.$$
In this context, we have formulated the material response modeling problem as to learn the solution operator $\mcG$ of an unknown PDE system from data. 
To emphasize the importance and challenges of learning the solution operator rather than a particular solution $\ub$, we notice that when the operators $\mcK_{\bb}$ and $\mcN_{\bb}$ are known, existing methods, ranging from the classical discretization of PDEs with known coefficients to modern machine learning (ML) approaches such as the basic version of physics-informed neural networks \cite{raissi2019physics}, lead to finding the solution $\ub\in\mcU$ for a single instance of the material parameter and loading $\fb\in\mcF$. However, when constitutive laws are unknown or when defects and/or high degrees of heterogeneity are present such that the classical constitutive models may become inaccurate, the operators $\mcK_{\bb}$ and $\mcN_{\bb}$ can not be predefined. 

Thus, our goal is to provide a {\it neural operator}, i.e., an approximated solution operator $\mcG[\cdot;\theta]:\fb\rightarrow \ub$ that delivers solutions of the system for any input $\fb$. This is a more challenging task for several reasons. First, in contrast to the classical NN approaches where the solution operator is parameterized between finite-dimensional Euclidean spaces \cite{guo2016convolutional,zhu2018bayesian,adler2017solving,bhatnagar2019prediction,khoo2021solving}, the neural operators are built as mappings between infinite-dimensional spaces, and they are resolution independent. As the consequence, \textit{no further modification or tuning will be required for different resolutions} in order to achieve the same level of solution accuracy \cite{li2020neural,li2020fourier,you2022nonlocal}. Second, for every new instance of material microstructure and/or loading scenarios $\fb$, the neural operators require only a forward pass of the network, which implies that the optimization problem \eqref{eqn:opt} \textit{only needs to be solved once and the resulting NN can be utilized to solve for multiple instances of the input parameter}. This property is in contrast to the classical numerical PDE methods \cite{leveque2007finite,zienkiewicz1977finite,karniadakis2005spectral} 
%\MD{\bf [maybe here we should cite PDE constrained optimization approaches rather then forward solvers]} 
and some ML approaches  \cite{raissi2019physics,weinan2018deep,bar2019unsupervised,smith2020eikonet,pan2020physics}, where the optimization problem needs to be solved for every new instance of the input parameter of a known governing law. 
Finally, of fundamental importance is the fact that the neural operators can find solution maps regardless of the presence of an underlying PDE and only require the observed data pairs $\{(\fb_j,\ub_j)\}_{j=1}^N$. Therefore, learning a data-driven neural operators would be particularly promising when the mechanical responses are provided by experimental measurements such as the displacement tracking data from DIC (see Section \ref{sec:dic}) or molecular dynamics simulations \cite{kim2019peri,you2022data} for which the material governing equations are not available.

%%%%%%%%%%%%%%%%%%%%%%%%%%%%%%%%%%%%%%%

\subsection{Three relevant integral neural operator architectures}\label{sec:3ino}

We now discuss the network architecture of three relevant integral neural operator learning methods, namely, the GKNs \cite{li2020neural,li2020multipole},  NKNs \cite{you2022nonlocal}, and FNOs \cite{li2020fourier}. To provide a consistent description of all three networks and illustrate their connections with the proposed IFNO architecture, we describe each model following a formulation similar to the one presented in \cite{you2022nonlocal}.

\paragraph{Lifting Layer} In integral neural operator models, we first lift the input $\fb(\cdot)\in\mcF$ to a representation (feature) $\hb(\cdot,0)$ that corresponds to the first network layer (also known as the lifting layer, see, e.g., \cite{kovachki2021neural}). In this section, we identify the first argument of $\hb$ with space (the set of nodes) and the second argument with time (the set of layers). Given an input vector field $\fb(\xb):\real^s\to\mbR^{d_F}$, we define the first network layer as 
$$\hb(\xb,0)=\mathcal{P}[\fb](\xb):=P(\xb)\fb(\xb)+\pb(\xb).$$
Here, $P(\xb)\in\real^{d\times d_F}$ and $\pb(\xb)\in\real^{d}$ define an affine pointwise mapping. In practice, $P(\xb)$ and $\pb(\xb)$ are often taken as constant parameters, i.e., $P(\xb)\equiv P$ and $\pb(\xb)\equiv \pb$.

\paragraph{Iterative Kernel Integration Layers} Then, we formulate the NN architecture in an iterative manner: 
\begin{equation}\label{eqn:layer}
\hb(\cdot,l\Delta t)=\mathcal{L}_l[\hb(\cdot,(l-1)\Delta t)],\quad l=1,\cdots,L,
%\hb(\cdot,0)\rightarrow \hb(\cdot,\Delta t)\rightarrow\hb(\cdot,2\Delta t)\rightarrow \cdots \rightarrow \hb(\cdot,T),
\end{equation}
where $\hb(\cdot,j\Delta t)$, $j=0,\cdots,L:=T/\Delta t$, is a sequence of functions representing the values of the network at each hidden layer, taking values in $\real^{d}$. $\mathcal{L}_1,\cdots,\mathcal{L}_{L}$ are the nonlinear operator layers defined via the action of the sum of a local linear operator (i.e., a nonlocal integral kernel operator) and a bias function. Within each layer, we treat the nodes within a layer as a continuum so that we have an infinite number of nodes, i.e., a layer has an infinite width. As such, each layer representation can be seen by a function of the continuum set of nodes $\omg\subset\real^s$. Then, we denote the $l$-th network representation by $\hb(\xb,l\Delta t):\real^s\times \mathbb N^+\to\mbRd$, or, equivalently, $\hb(\xb,l\Delta t)=\hb(\xb,t):\real^s\times(0,T]\to\mbRd$. Here, $l=0$ (or equivalently, $t=0$) denotes the first hidden layer, whereas $t=L\Delta t$ (or $t=T$) for the last hidden layer. The use of the symbol $t$ stems from the relationship that can be established between the network update and a time stepping scheme.
% For integral neural operators, the layer update rule in \eqref{eqn:layer} is assumed to take the form:
% \begin{equation}\label{eq:ino}
% \hb(\xb,(l+1)\Delta t)=A_1\hb(\xb,l\Delta t)+A_2\sigma\left(R\hb(\xb,l\Delta t)+\int_\omg k(\xb,\yb,\fb(\xb),\fb(\yb);\vb)\hb(\yb,l\Delta t) d\yb + \cb\right),
% \end{equation}
% where $A_1$ and $A_2$ are fixed coefficients to be specified later,  and 
% $\sigma$ is an activation function which is taken as the popular rectified linear unit function in this paper:
% \begin{displaymath}
% \text{ReLU}(x):=\left\{\begin{array}{cc}
%      0,& \text{ for }x\leq 0; \\
%      x,& \text{ for }x>0.
% \end{array}\right.
% \end{displaymath}

\paragraph{Projection Layer} Third, the output $\ub(\cdot)\in\mcU$ is obtained through a projection layer. In particular, we project the last hidden layer representation $\hb(\cdot,T)$ onto $\mcU$ as:
$$\ub(\xb)=\mathcal{Q}[\hb(\cdot,T)](\xb):=Q_2(\xb)\sigma(Q_1\hb(\xb,T)+\qb_1(\xb))+\qb_2(\xb).$$
Here, $Q_1(\xb)\in\real^{d_{Q}\times d}$, $Q_2(\xb)\in\real^{d_{u}\times d_Q}$, $\qb_1(\xb)\in\real^{d_Q}$ and $\qb_2(\xb)\in\real^{d_u}$ are the appropriately sized matrices and vectors that are part of the parameter set that we aim to learn. $\sigma$ is an activation function. Unless otherwise stated, in this work we choose $\sigma$ to be the popular rectified linear unit (ReLU) function:
\begin{equation}\label{eqn:relu}
 \text{ReLU}(x):=\left\{\begin{array}{cc}
      0,& \text{ for }x\leq 0; \\
      x,& \text{ for }x>0.
 \end{array}\right.
\end{equation}
Similarly as for the lifting layer, $Q_1(\xb)$, $Q_2(\xb)$, $\qb_1(\xb)$ and $\qb_2(\xb)$ are also often taken as constant parameters, which will be denoted as $Q_1$, $Q_2$, $\qb_1$ and $\qb_2$, respectively.

To sum up, the integral neural operators can be written as mappings of the form:
\begin{equation}
    \mcG[\fb;\theta]=\mathcal{Q}\circ\mathcal{L}_L\circ\mathcal{L}_{L-1}\circ\cdots\circ\mathcal{L}_1\circ \mathcal{P}[\fb].
\end{equation}
The architectures of the GKNs, FNOs, NKNs, and our IFNOs mainly differ in the design of their iterative layer update rules in \eqref{eqn:layer}, which will be elaborated in more detail for each method below. We also summarize their benefits and limitations in Table \ref{tab:comparison}, to highlight that the proposed IFNOs are designed in such a way that all the benefits of these approaches are preserved, while the limitations are overcome.

%%%
\paragraph{Graph Kernel Networks (GKNs)} As the first integral neural operator, the GKNs introduced in \cite{li2020neural} have the foundation in the representation of the solution of a PDE by the Green's function. In the GKNs, it is assumed that the iterative kernel integration part is invariant across layers, i.e.,
$$\mathcal{L}_1=\mathcal{L}_2=\cdots=\mathcal{L}_L:=\mathcal{L}^{GKN},$$
with the update of each layer network given by
\begin{equation}\label{eq:gkn}
\hb(\xb,(l+1)\Delta t)=\mathcal{L}^{GKN}[\hb(\xb,l\Delta t)]:=\sigma\left(W\hb(\xb,l\Delta t)+\int_\omg \kappa(\xb,\yb,\fb(\xb),\fb(\yb);\vb)\hb(\yb,l\Delta t) d\yb + \cb\right).
\end{equation}
Here, $\sigma$ is an activation function, $W\in\real^{d\times d}$ and $\cb\in\real^d$ are the learnable tensors, and $\kappa\in\real^{d\times d}$ is a tensor kernel function that takes the form of a (usually shallow) NN whose parameters $\vb$ are to be learned. The GKN resembles the original ResNet block \cite{He2016Resnet}, where the usual discrete affine transformation is substituted by a continuous integral operator. Therefore, the learnt network parameters are resolution-independent: the learned $W$, $\cb$, and $\vb$ are close to optimal even when used with different resolutions, i.e., with different partitions/discretizations of the domain $\omg$. However, despite its advantage on resolution-independence, in the presence of complex learning tasks the applicability of the GKNs may become compromised by two factors. First, in the most general version of the GKNs, the integral in \eqref{eq:gkn} is realized through a message passing graph neural network architecture on a fully-connected graph. Therefore, the GKNs are generally much more expensive than other integral neural operators, say, FNOs, making the GKNs less favorable for large-scale problems. Second, although single-layer and shallow GKNs have been shown to be successful in learning governing equations, e.g., the Darcy \cite{li2020neural} and Burgers \cite{li2020multipole} equations, it was found in \cite{you2022nonlocal} that the GKNs may become unstable when the number of its layers increases. As the GKN becomes deeper, either there is no gain in accuracy or increasing values of the loss function occur.

\paragraph{Nonlocal Kernel Networks (NKNs)} As a deeper and stabilized modification of the GKNs, the NKNs are introduced in \cite{you2022nonlocal} to handle both learning governing equations and classifying images tasks. The NKN stems from the interpretation of the neural network as a discrete nonlocal diffusion reaction equation that, in the limit of infinite layers, is equivalent to a parabolic nonlocal equation. Therefore, its stability in the deep layer limit can be analyzed via nonlocal vector calculus. In the NKNs, the iterative kernel integration is also assumed to be layer-independent. Differs from the GKNs where the next layer representation is defined via a nonlinear operator, the increment of each layer network representation is defined as a nonlinear operator in the NKNs. In particular, the network hidden layer update is given as 
%$\hb(\xb,(l+1)\Delta t)=\mathcal{L}^{NKN}(\hb(\xb,l\Delta t))$, where:
{\begin{align}
\nonumber&\hb(\xb,(l+1)\Delta t)=\mathcal{L}^{NKN}[\hb(\xb,l\Delta t)]\\
&~~:=\hb(\xb,l\Delta t)+ {\Delta t}\left(\int_\omg \kappa(\xb,\yb,\fb(\xb),\fb(\yb);\vb)(\hb(\yb,l\Delta t)-\hb(\xb,l\Delta t)) d\yb-W(\xb;\wb)\hb(\xb,l\Delta t)+\cb\right).\label{eq:nkn}
\end{align}}
As for the GKNs, the kernel tensor function $\kappa\in\real^{d\times d}$ is modeled by a NN parameterized by $\vb$. The reaction term $W\in\real^{d\times d}$ is modeled by another NN parameterized by $\wb$. The NKN architecture preserves the continuous, integral treatment of the interactions between nodes that characterizes the GKNs, and hence enables resolution independence with respect to the inputs. On the other hand, by modeling the layer representation increment and identifying the number of layers with the number of time steps in a time-discretization scheme, the training of deep NNs in the NKNs is accelerated via the shallow-to-deep technique \cite{ruthotto2019deep}. In particular, it is obvious to see that by diving both sides of \eqref{eq:nkn} by $\Delta t$, the term $(\hb(\cdot,(l+1)\Delta t)-\hb(\cdot,l\Delta t))/\Delta t$ corresponds to the discretization of a first-order derivative so that this architecture can be interpreted as a nonlinear differential equation in the limit of deep layers, i.e., as $\Delta t\to 0$. Thus, the optimal parameters ($\vb$, $\wb$ and $\cb$) of a shallow network are interpolated and will be reused in a deeper one as initial guesses. In \cite{you2022nonlocal}, it is found that the NKNs generalize well to different resolutions and stays stable when the network is getting deeper.

Similarly to the GKNs, since the building blocks of the NKNs are integral operators characterized by space dependent kernels with minimal assumptions, they come at the price of a higher computational cost compared to other networks whose kernels have a convolutional structure (e.g., the standard CNN and FNO). Hence, the NKNs are computationally more expensive than the FNOs, and generally less favorable in large-scale learning tasks.%This is due to the fact that both $k$ and $R$ are neural networks. 

\paragraph{Fourier Neural Operators (FNOs)} The Fourier neural operator (FNO) was first proposed in \cite{li2020fourier}, where the integral kernel $\kappa$ is parameterized in the Fourier space. In particular, the FNO drops the dependence of kernel $\kappa$ on the input $\bb$ and assumes that $\kappa(\xb,\yb;\vb):=\kappa(\xb-\yb;\vb)$. The integral operator in \eqref{eq:gkn} then becomes a convolution operator so that $\kappa$ can be parameterized directly in the Fourier space. The corresponding $l-$th layer update is then given by
\begin{equation}\label{eq:fno}
\hb(\xb,(l+1)\Delta t)=\mathcal{L}^{FNO}_{l+1}[\hb(\xb,l\Delta t)]:=\sigma\left(W_l\hb(\xb,l\Delta t)+\mathcal{F}^{-1}[\mathcal{F}[\kappa(\cdot;\vb_l)]\cdot \mathcal{F}[\hb(\cdot,l\Delta t)]](\xb)+ \cb_l(\xb)\right),
\end{equation}
where $\mathcal{F}$ and $\mathcal{F}^{-1}$ denote the Fourier transform and its inverse, respectively. In practice, $\mathcal{F}$ and $\mathcal{F}^{-1}$ are computed using the the FFT algorithm and its inverse to each component of $\hb$ separately, with the highest modes truncated and keeping only the first $k$ modes. $\cb_l(\xb)$ defines a pointwise bias, which is often taken as a constant bias $\cb_l(\xb)\equiv\cb_l$ (see, e.g., \cite{kovachki2021neural,kovachki2021universal}). Therefore, $\mathcal{F}[\hb(\cdot,l\Delta t)]$ has the shape $d\times k$, and the trainable parameters for each hidden layer will be $\cb_l\in\real^d$, $W_l\in\real^{d\times d}$, and $\mathcal{F}[\kappa(\cdot;\vb_l)]:=R_l\in\complex^{d\times d\times k}$. {Here, we use $W_l$, $\cb_l$ and $\vb_l$ to highlight the fact that in the FNOs, each layer has different parameters (i.e., different kernels, weights and biases).} This is different from the layer-independent kernel in the GKNs and NKNs, and makes the total number of trainable parameters in the FNOs as $DOF^{FNO}:=[d(1+d_F)]+[L(d+d^2+2d^2k)]+[d_Q(d+d_u+1)+d_u]$. Here, the first part is the number of parameters associated with the lifting layer, the second part is associated with the $L$ iterative kernel integration layers, and the last part comes from the projection layer. As the network gets deeper, the second part dominates the total number of parameters, and therefore, the number of trainable parameters in the FNOs grows almost linearly with the increase of $L$.

Comparing with the GKN and NKN, the FNO has superior efficiency because one can use the FFT to compute \eqref{eq:fno}. Moreover, in \cite{kovachki2021universal}, Kovachiki et al. have proved that with sufficiently large depth $L$, the FNOs are universal in the sense that they can approximate any continuous operator to a desired accuracy. However, the number of trainable parameters in the FNOs increases as the network gets deeper, which makes the training process of the FNOs more challenging and potentially prone to over-fitting. In \cite{you2022nonlocal}, it was found that when the network gets deeper, the training error decreases in the FNO while the test error becomes much larger than the training error, indicating that the network is overfitting the training data. Furthermore, if one further increases the number of hidden layer $L$, training the FNOs becomes challenging due to the vanishing gradient phenomenon. 
%In contrast, NKNs trained with the shallow-to-deep initialization are robust to the vanishing gradient phenomenon, and not subject to overfitting issues. 
On the other hand, as reported in \cite{lu2021comprehensive}, the vanilla version of the FNO is generally restricted to simple geometries and structured data. Although the FNO has superior efficiency and is a theoretically proved universal approximator, its application is generally limited to the cases when the data is structured and less complex such that a shallow network would be sufficient.

\section{Implicit Fourier Neural Operators (IFNOs)}\label{sec:ifno}

To overcome the limitations of the architectures mentioned in Section \ref{sec:3ino}, we propose Implicit Fourier Neural Operators (IFNOs), an efficient, deep, and stable integral neural operator for solution operator learning problems. In particular, we first formulate the solution operator as an implicitly defined mapping, and then propose to model it as a fixed point, not via an explicit mapping. Based on this idea, we provide the hidden layer network formulation for the IFNO and illustrate the shallow-to-deep training technique. While the former reduces the number of trainable parameters and memory cost, the latter aims to resolve the difficulty of network training in the limit of deep layers. Finally, we discuss the expressiveness of the IFNOs by showing that as far as there exists a converging fixed point equation for the target implicit problem, the IFNOs would be universal. In the present study, we assume that the datum are structured so the FFT can be employed, and we also note that when the problem domain $\omg$ and the discretization $\chi$ are not structured, one might employ the nonlinear mapping extension technique developed in \cite{lu2021comprehensive} to obtain a structured datum so that the IFNO, based on the following discussion, is still applicable.

%%%%%%%%%%%%%%%%%%%%%%%%%%%%%%%%%%%%%%%%%%%%%%%
\subsection{The network architecture}

\begin{figure}[t]
	\centering
	\includegraphics[width=0.9\columnwidth]{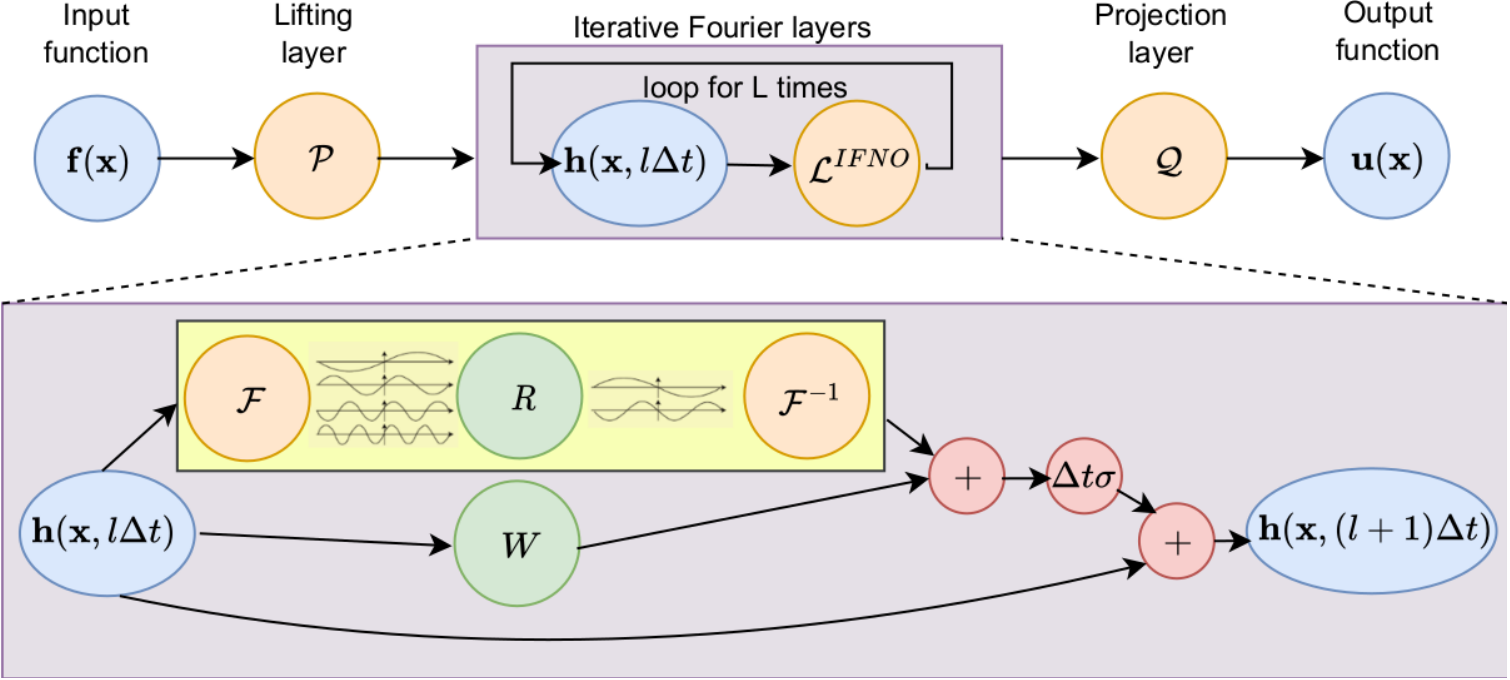}
	\caption{The architecture of IFNO: start from input $\fb(\xb)$, then 1) Lift to a high dimensional feature space by the lifting layer $\mathcal{P}$ and obtain the first hidden layer representation $\hb(\xb,0)$; 2) Apply $L$ iterative layers with the formulation proposed in \eqref{eq:IFNO}; 3) Project the last hidden layer representation $\hb(\xb,L\Delta t)$ back to the target dimension by a shallow network $\mathcal{Q}$.}
\label{fig:domain}
\end{figure}

We now propose an IFNO for the solution of the problem outlined in \eqref{eqn:pde}. To see a guiding principle for our architecture, let us consider the following boundary displacement example:
\begin{equation}\label{eqn:u_D}
\begin{aligned} 
\mcK_{\bb}[\ub](\xb)=\gb(\xb),\quad&\xb\in \omg,\\
\ub(\xb)=\ub_D(\xb),\quad&\xb\in\partial \omg,\\
\end{aligned}
\end{equation}
where $\mcK_{\bb}$ is a differential operator depending on the (possibly nonlinear) material constitutive law, and $\ub_D$ is the prescribed displacement on the boundary. Given a discretization of the domain defined as $\chi=\{\xb_1,\cdots,\xb_M\}$, the desired network output, or equivalently the numerical solution of \eqref{eqn:pde}, is then $\Ub=[\Ub_1,\Ub_2,\cdots,\Ub_M]\approx[\ub(\xb_1),\cdots,\ub(\xb_M)]$. Here, we assume, without loss of generality, that the first $\beta$ number of points are on $\partial\omg$, and therefore the solution $\Ub$ on these points is prescribed by the displacement boundary condition $\ub_D$. With proper discretization methods, such as the finite difference method, for the differential operator $\mcK_{\bb}$ and an instance of input vector $\Fb=[\bb(\xb_1),\cdots,\bb(\xb_M),\gb(\xb_{1}),\cdots,\gb(\xb_M),\ub(\xb_1),\cdots,\ub(\xb_\beta),\mathbf{0},\cdots,\mathbf{0}]$, the numerical solution $\Ub$ is determined from the following implicit system of equations:
\begin{equation}\label{eqn:R}
\mcH(\Ub;\Fb):=\left[\begin{array}{c}
\Ub_1-\ub_D(\xb_1)\\
\vdots\\
\Ub_\beta-\ub_D(\xb_\beta)\\
\mcK_{\bb}^h(\Ub)-\Gb\\
\end{array}\right]=\mathbf{0}.   
\end{equation}
Herein, $\Gb:=[\gb(\xb_{\beta+1}),\cdots,\gb(\xb_M)]$ is the loading term, and $\mcK_{\bb}^h$ is the discretized operator. To solve for $\Ub$ from the nonlinear system in \eqref{eqn:R}, one can employ fixed-point iteration methods, such as its special case -- the Newton-Raphson method. In particular, with an initial guess of the solution (denoted as $\Ub^0$), the process is repeated to produce successively better approximations to the roots of \eqref{eqn:R} following:
\begin{equation}\label{eqn:newton}
 \Ub^{l+1}=\Ub^l-(\nabla \mcH(\Ub^l;\Fb))^{-1}\mcH(\Ub^l;\Fb):=\Ub^l+\mcR(\Ub^l,\Fb),
\end{equation}
until a sufficiently precise value is reached. Here, we noticed that for each implicit problem, there are infinite numbers of the corresponding fixed point equations, and \eqref{eqn:newton} is just one example. In fact, the fixed point method solves the implicit system as long as there exists one fixed point equation with a convergent and unique solution.

Guided by the representation in \eqref{eqn:newton}, we argue that the desired network output is more aptly described implicitly, and propose to develop a network architecture to model the operator $\mcR$ and mimic the fixed point method by design. 
%In fact, the idea of using fixed-point methods was also proposed in implicit networks \cite{el2021implicit,bai2019deep,winston2020monotone,bai2020multiscale} as an efficient training procedure. 
Using the same notations of Section \ref{sec:background}, we propose the following iterative network update formulation
{\begin{align}
\nonumber\hb(\xb,(l+1)\Delta t)=&\mcL^{IFNO}[\hb(\xb,l\Delta t)]\\
:=&\hb(\xb,l\Delta t)+ {\Delta t}\sigma\left(W\hb(\xb,l\Delta t)+\mathcal{F}^{-1}[\mathcal{F}[\kappa(\cdot;\vb)]\cdot \mathcal{F}[\hb(\cdot,l\Delta t)]](\xb)+ \cb(\xb)\right).\label{eq:IFNO}
\end{align}}
Note that although the FFT is still applied to each component of $\hb$ separately with the highest modes truncated as for the FNOs, the hidden layer parameters are taken to be layer-independent, which is distinctly different from the FNOs. Following the conventions in the FNOs, we also take the bias $\cb_l(\xb)$ as a constant bias ($\cb(\xb)\equiv\cb$) in all the subsequent numerical tests. Therefore, the set of trainable parameters in our IFNOs are $P\in\real^{d\times d_F}$ and $\pb\in\real^{d}$ for the lifting layer, $Q_1\in\real^{d_{Q}\times d}$, $Q_2\in\real^{d_{u}\times d_Q}$, $\qb_1\in\real^{d_Q}$ and $\qb_2\in\real^{d_u}$ for the projection layer, and $\cb\in\real^d$, $W\in\real^{d\times d}$ and $\mathcal{F}(\kappa(\cdot;\vb))=R\in\complex^{d\times d\times k}$ for the hidden layers. The total number of trainable parameters is $DOF^{IFNO}:=[d(1+d_F)]+[d+d^2+2d^2k]+[d_Q(d+d_u+1)+d_u]$, which is independent of the number of hidden layers $L$, alleviating the major bottleneck of the overfitting issue encountered by the original FNOs with a deeper network. Moreover, this feature also enables the straightforward application of the shallow-to-deep initialization technique. 
%and reduces the computational effort and memory allocation. 

As the layer becomes deep ($\Delta t\rightarrow 0$), \eqref{eq:IFNO} can be seen as an analog of a discretized ordinary differential equations (ODEs). This allows us to exploit the shallow-to-deep learning technique described in Section \ref{sec:background} for the NKNs. Similarly to in \eqref{eq:nkn}, we can reinterpret the network update as the time discretization of a differential equation and use the optimal parameters obtained with $L$ layers as the initial guesses for deeper networks. Specifically, let $W$, $\cb$ and $R$ be the optimal network parameters obtained by training an IFNO of depth $L$. For further improving the accuracy of the network, we can increase the number of layers (or equivalently, time steps), and train a new network of depth $\widetilde{L}>L$. The idea of the shallow-to-deep technique is to perform interpolation in time (or across layers) over the optimal parameters obtained at depth $L$ and to scale them in such a way that the final time of the differential equation remains unchanged. In our specific setting, due to the fact that the network parameters are not time dependent, this technique simply corresponds to the initialization of the (deeper) $\widetilde L$-layer network by $W$, $\cb$ and $R$.

As a further note, we point out that although the idea of using repeated hidden layers has not been explored for the FNOs, resembling fixed-point methods is not new for neural networks. In \cite{el2021implicit,bai2019deep,winston2020monotone,bai2020multiscale}, implicit networks are introduced as an analog to a forward pass through an ``infinite depth'' network, without storing the intermediate quantities of the forward pass for back-propagation, and hence can be trained using constant memory costs with respect to depth. One can see that our IFNO architecture requires only constant memory cost, similar to implicit networks. Moreover, it preserves the continuous, integral treatment of the interactions between nodes that characterizes integral neural operators. Therefore, the IFNO provides a new and efficient implicit-type neural operator architecture -- that is why it is named ``implicit''. 

Table \ref{tab:comparison} summarizes relevant properties of the IFNOs in comparison with other integral neural operators. 
%These statements are confirmed and illustrated by the numerical tests as will be reported in Section \ref{sec:experiments}. 
In summary, being a resemblance of an implicit equation solver and stable in the limit of deep layers make the proposed IFNO's architecture a viable tool for modeling problems with complex material responses, since these problems can be considered as PDE solution operator learning tasks.

\subsection{Universal approximation properties}

In this section, we show that the IFNOs are universal solution finding operators, in the sense that they can approximate a fixed point method to a desired accuracy. Without loss of generality, we consider a 1D domain $\omg\subset\real$, 
%both the input and output are scalar functions, i.e., 
$\fb(\xb):=[\xb,\hat{\fb}(\xb)]\in\real^2$ and $\ub(\xb)\in\real$.
The function $\ub\in C(\omg)$ is evaluated at uniformly distributed nodes $\chi =\{\xb_1,\xb_2,\dots,\xb_M\}$. Let us denote $\Ub^* := 
[\ub(\xb_1),\ub(\xb_2),\dots,\ub(\xb_M)]$ as the solution we seek, $\Ub^0:=[\xb_1,\cdots,\xb_M]$ as the initial guess, $\Cb=[\cb(\xb_1),\cdots,\cb(\xb_M)]$ as the collection of pointwise bias vectors $\cb(\xb_i)$, and $\Fb := [\hat{\fb}(\xb_1),\hat{\fb}(\xb_2),\dots,\hat{\fb}(\xb_M)]$ as the loading vector. We aim to show that for any desired accuracy $\varepsilon>0$, one can find a sufficiently large $L>0$ and a set of parameters $\theta_{\varepsilon}=\{P,\pb,Q_1,Q_2,\qb_1,\qb_2,\Cb,W,R\}$, such that the resultant IFNO model satisfies
$$\vertii{\mathcal{Q}\circ(\mathcal{L}^{IFNO})^L\circ \mathcal{P}([\Ub^0, \Fb ]^{\mathrm{T}})-\Ub^*}\leq\varepsilon,\quad \forall \Fb\in\real^M.$$
Here, the matrix and vector parameters in the lifting and projection layers are taken as pointwise functions. With a slight abuse of notation, we denote $P\in\real^{dM\times d_FM}$ as the collection of the pointwise weight matrices at each discretization point in $\chi$, and a similar convention applies for other matrix and vector parameters in the lifting and projection layers. Hence, the dimension of all trainable parameters are: $\Cb\in\real^{dM}$, $W\in\real^{d\times d}$, $R\in\complex^{d\times d\times k}$, $P\in\real^{dM\times d_FM}$, $\pb\in\real^{dM}$, $Q_1\in\real^{d_QM\times dM}$, $Q_2\in\real^{d_u M}\times d_QM$, $\qb_1\in\real^{d_QM}$ and $\qb_2\in\real^{d_uM}$. With the assumption that $\fb(\xb)\in\real^2$ and $\ub(\xb)\in \real$, we note that $d_F=2$ and $d_u=1$. As will be seen in the proof below, we will further take $d_Q=d$. For the simplicity of notation, in this section we organize the feature vector $\Hb\in\real^{dM}$ in a way such that the components corresponding to each discretization point are adjacent, i.e., $\Hb=[\Hb(\xb_1),\cdots,\Hb(\xb_M)]$ and $\Hb(\xb_i)\in\real^{d}$. For simplicity, we further assume that the Fourier coefficient is not truncated, and all available frequencies will be used. We point out that under this circumstance, we have $k=M$ and the (discretized) iterative layer can be written as
\begin{align*}
\mcL^{IFNO}[\Hb(l\Delta t)]=&\Hb(l\Delta t)+ \Delta t\sigma\left(\tilde{W}\Hb(l\Delta t)+\text{Re}(\mathcal{F}_{\Delta x}^{-1}(R\cdot \mathcal{F}_{\Delta x}(\Hb(l\Delta t))))+\Cb\right)\\
=&\Hb(l\Delta t)+ {\Delta t}\sigma\left(V^{IFNO}\Hb(l\Delta t)+\Cb\right),
\end{align*}
with 
$$ V^{IFNO} := \text{Re}\begin{bmatrix}
    \sum\limits_{n=0}^{M-1} R_{n+1}+W & \sum\limits_{n=0}^{M-1} R_{n+1} \exp(\frac{2i\pi\Delta x n}{M}) &
    \dots & \sum\limits_{n=0}^{M-1} R_{n+1} \exp(\frac{2i\pi(M-1)\Delta x n}{M}) \\
     \sum\limits_{n=0}^{M-1} R_{n+1} \exp(\frac{2i\pi\Delta x n}{M}) 
     & \sum\limits_{n=0}^{M-1} R_{n+1}+W & \dots  & \sum\limits_{n=0}^{M-1} R_{n+1} \exp(\frac{2i\pi(M-2)\Delta x n}{M}) \\
     \vdots & \vdots & \ddots & \vdots \\
     \sum\limits_{n=0}^{M-1} R_{n+1} \exp(\frac{2i\pi(M-1)\Delta x n}{M})
      & \sum\limits_{n=0}^{M-1} R_{n+1} \exp(\frac{2i\pi(M-2)\Delta x n}{M}) &\dots&  \sum\limits_{n=0}^{M-1} R_{n+1}+W \\
    \end{bmatrix}.$$
Here, $R\in\complex^{M\times d\times d}$ with $R_i\in\complex^{d\times d}$ being the component associated with each discretization point $\xb_i\in\chi$, ${V^{IFNO}}\in\real^{dM\times dM}$, $\Cb\in\real^{dM}$, $\tilde{W}:=W\oplus W\oplus \cdots \oplus W$ is a $dM\times dM$ block diagonal matrix formed by $W\in\real^{d\times d}$,  $\mathcal{F}_{\Delta x}$ and $\mathcal{F}_{\Delta x}^{-1}$ denote the discrete Fourier transform and its inverse, respectively. By further taking $R_2=\cdots=R_M=W=0$, a $d\times d$ matrix with all its elements being zero, it suffices to show the universal approximation property for an iterative layer as follows:
$$\mcL^{IFNO}(\Hb(l\Delta t)):=\Hb(l\Delta t)+ {\Delta t}\sigma\left(\tilde{V}\Hb(l\Delta t)+\Cb\right)$$
where $\tilde{V}:=\mathbf{1}_{[M,M]}\otimes V$ with $V\in\real^{d\times d}$ and $\mathbf{1}_{[m,n]}$ being an $m$ by $n$ all-ones matrix.

Before stating our main theoretical results, we need the following assumptions on $\Ub^*$ and $\mathcal{R}$:
\begin{assumption}\label{asp:1}
There exists a fixed point equation, $\Ub=\Ub+\mcR(\Ub,\Fb)$ for the implicit problem \eqref{eqn:R}, such that $\mathcal{R}:\mathbb{R}^{2M} \mapsto \mathbb{R}^M$ is a continuous function satisfying $\mathcal{R}(\Ub^*,\Fb) = \mathbf{0}$ and $||\mathcal{R}(\hat{\Ub},\Fb)-\mathcal{R}(\tilde{\Ub},\Fb)||_{l^2(\mathbb{R}^M)} \leq m ||\hat{\Ub}-\tilde{\Ub}||_{l^2(\mathbb{R}^M)}$ for any two vectors $\hat{\Ub},\tilde{\Ub}\in\real^{M}$. Here, $m > 0$ is a constant independent of $\Fb$.
   \end{assumption}
  
\begin{assumption}\label{asp:2}
   With the initial guess $\Ub^0:=[\xb_1,\cdots,\xb_M]$, the fixed-point iteration 
    \begin{equation*}
        \Ub^{l+1} = \Ub^{l}  + \mathcal{R}(\Ub^l,\Fb), \quad l=0,1,\dots.
    \end{equation*}
    converges, i.e., for any given $\varepsilon > 0$, there exists an integer $L$ such that 
    \begin{equation*}
        ||\Ub^l - \Ub^*||_{l^2(\mathbb{R}^M)} \leq \varepsilon, \quad \forall l > L,
    \end{equation*}
for all possible input instances $\Fb\in\real^M$ and their corresponding solutions $\Ub^*$.
\end{assumption}

Next, we prove that the IFNOs are universal, i.e., give a fixed point method and solution $\Ub^*$ satisfying  Assumptions \ref{asp:1}-\ref{asp:2}, one can find an IFNO whose output approximates $\Ub^*$ to a desired accuracy, $\varepsilon>0$. To be more precise, we will prove the following theorem:
\begin{theorem}[Universal approximation] \label{thm:main}
Let $\Ub^* = [\ub(\xb_1),\ub(\xb_2),\dots,\ub(\xb_M)]$ 
be the ground-truth solution that satisfies Assumptions \ref{asp:1}-\ref{asp:2}, the activation function $\sigma$ for all iterative kernel integration layers be the ReLU function, and the activation function in the projection layer be the identity function. Then for any $\varepsilon > 0$, there exist sufficiently large layer number $L>0$ and feature dimension number $d>0$, such that one can find a parameter set $\theta_{\varepsilon}=\{P,\pb,Q_1,Q_2,\qb_1,\qb_2,\Cb,V\}$ with $P\in\real^{dM\times 2M},\pb\in\real^{dM},Q_1\in\real^{dM\times dM},Q_2\in\real^{M\times dM},\qb_1\in\real^{dM},\qb_2\in\real^{M},\Cb\in\real^{dM},V\in\real^{d\times d}$ with the corresponding IFNO model satisfies 
\begin{equation*}
    \vertii{\mathcal{Q}\circ(\mathcal{L}^{IFNO})^L\circ \mathcal{P}([\Ub^0, \Fb ]^{\mathrm{T}})-\Ub^*}\leq\varepsilon,\quad \forall \Fb\in\real^M. 
\end{equation*}
\end{theorem}

Before proceeding to the proof of this main theorem, we first show the approximation property of a shallow neural network:
\begin{lemma}\label{lemma:1}
Given a continuous function ${\mathcal{T}}: 
\mathbb{R}^{2M} \mapsto \mathbb{R}^M$, and a non-polynomial and continuous activation function $\sigma$, for any constant $\varepsilon>0$ there exists a shallow neural network model $\hat{\mathcal{T}}:= S\sigma \left(B\Xb +A \right)$ such that 
\begin{equation*} 
    ||\mathcal{T}(\Xb) - \hat{\mathcal{T}}(\Xb)||_{l^2(\mathbb{R}^{M})} \leq \varepsilon, \quad \forall \Xb\in\real^{2M},
\end{equation*}
for sufficiently large feature dimension $\tilde{d}>0$. 
%where $\hat{\mathcal{R}}$ is defined as 
%\begin{equation}\label{eqn:Nh}
%    \hat{\mathcal{R}}(\xb) := S\sigma \left(R\xb +\bm{a} \right),
%\end{equation}
Here, $S\in \mathbb{R}^{M \times \tilde{d}M}$, $B \in \mathbb{R}^{\tilde{d}M \times 2M}$, and $A \in \mathbb{R}^{\tilde{d}M}$ are matrices/vectors which are independent of $\Xb$.
\end{lemma}

%\begin{equation}\label{eqn:ifno}
%  \hb(\xb,t+\Delta t) = \hb(\xb,t) + \underbrace{\Delta t \sigma \left(R\hb(\xb,t) + \int_D k(\xb-\yb;\vb)\hb(\yb,t)\mathrm{d}\yb +\cb \right)}_{\text{$\mathbf{\mathcal{R}}(\hb(\xb,t))$}}, 
%\end{equation}
%it can be viewed as a fixed-point iteration process with the goal to
%find $\hb$ such that $\mathbf{\mathcal{R}}(\hb) = \mathbf{0}$. 

%Next we show that any continuous function $\mathbf{\mathcal{R}}(\hb): 
%\mathbb{R}^d \mapsto \mathbb{R}^d$ can be approximated by 
%\begin{equation}\label{eqn:Nh}
%    \hat{\mathcal{R}}(\hb) = W\sigma \left(R\hb + \int_D k(\xb-\yb;\vb)\hb(\yb)\mathrm{d}\yb +\cb \right)
%\end{equation}
%to any degree of accuracy. For the simplicity of the notation, we will use $K(\xb)\hb$ to replace $\int_D k(\xb-\yb;\vb)\hb(\yb)\mathrm{d}\yb$ in the following proof. 

%\begin{theorem}
%Given a continuous function $\mathcal{R}:\mathbb{R}^d \mapsto\mathbb{R}^d$, and $\sigma$ is non-polynomial and continuous. There exist $\mathcal{N}$, such that 
%\begin{equation*}
%    ||\mathcal{R}(\hb) - \hat{\mathcal{R}}(\hb)||_{L^2(\mathbb{R}^d)} \leq \varepsilon 
%\end{equation*}
%\end{theorem}
\begin{proof}
As shown in \cite{pinkus1999approximation}, when $\sigma$ is non-polynomial and continuous, $\text{span}(\sigma(\rb\cdot\Xb  + a))$ is dense in $C(\mathbb{R}^{2M})$, where $\rb \in \mathbb{R}^{1\times 2M}$, and $a \in \mathbb{R}$.
Therefore, denoting $\mathcal{T}(\Xb) = [\mathcal{T}_1(\Xb),\dots,\mathcal{T}_M(\Xb)]$, for each $\mathcal{T}_i(\Xb)\in\real$ there exist $\bm{s}_i \in \mathbb{R}^{1\times \tilde{d}}, B_i \in \mathbb{R}^{\tilde{d} \times 2M}$, and $\bm{a}_i \in \mathbb{R}^{\tilde{d}}$, such that 
\begin{equation*}
    |\mathcal{T}_i(\Xb) - \bm{s}_i \sigma(B_i\Xb+\bm{a}_i)| \leq \frac{\varepsilon}{\sqrt{M}}, \quad \forall \Xb\in\real^{2M}.
\end{equation*}
Let 
\begin{equation*}
    S: = \begin{bmatrix}
    \bm{s}_1 & \mathbf{0} & \dots & \mathbf{0} \\
    \mathbf{0} & \bm{s}_2 & \dots & \mathbf{0} \\
    \vdots & \vdots & \ddots & \vdots  \\
    \mathbf{0} & \mathbf{0} & \dots & \bm{s}_M \\ 
    \end{bmatrix}\in\real^{M\times\tilde{d}M},\quad
    B := \begin{bmatrix}
        B_1 \\ 
        B_2 \\
        \vdots \\
        B_M
    \end{bmatrix}\in\real^{\tilde{d}M\times 2M},\quad
    A =\begin{bmatrix}
        \bm{a}_1 \\ 
        \bm{a}_2 \\
        \vdots \\
        \bm{a}_M \\
    \end{bmatrix}\in\real^{\tilde{d}M},
\end{equation*}
we then obtain
\begin{align*}
     ||\mathcal{T}(\Xb) -  S\sigma \left(B\Xb +A \right)||_{l^2(\mathbb{R}^M)} &= \sqrt{\sum_{i=1}^{M}|\mathcal{T}_i(\Xb) - \bm{s}_i \sigma(B_i\Xb+\bm{a}_i)|^2} \leq \sqrt{M \times \frac{\varepsilon^2}{M}} = \varepsilon,
\end{align*}
for all $\Xb\in\real^{2M}$.
\end{proof}

% Next, we prove that IFNOs are universal, i.e., give any fixed point method and solution $\Ub^*$ satisfying the Assumptions \ref{asp:1}-\ref{asp:2}, one can find an IFNO whose output approximates $\Ub^*$ to desired accuracy. To be more precise, we have the following theorem:
% \begin{theorem}[Universal approximation] Let $\Ub^* = (\ub(\xb_1),\ub(\xb_2),\dots,\ub(\xb_M))$ 
% be the ground-truth solution that satisfies Assumptions \ref{asp:1}-\ref{asp:2}, and $\sigma$ be the ReLU function as defined in \eqref{eqn:relu}. Then for any $\varepsilon > 
% 0$, there exists an IFNO, whose output is denoted as 
% $\Ub_{IFNO}$, such that 
% \begin{equation*}
%     ||\Ub_{IFNO} - \Ub^*||_{l^2(\mathbb{R}^M)} < \varepsilon.
% \end{equation*}
% \end{theorem}

We now proceed to the proof of Theorem \ref{thm:main}:
\begin{proof}
Since $\Ub^*$ satisfies Assumptions \ref{asp:1}-\ref{asp:2}, for any $\varepsilon > 0$, we first pick a sufficiently large integer $L$ such that $||\Ub^L - \Ub^*||_{l^2(\mathbb{R}^M)} \leq \frac{\varepsilon}{2}$. As shown in Lemma \ref{lemma:1}, %allows us to approximate function $\mathcal{R}$ to any desired accuracy. 
%Therefore, 
with sufficiently large feature dimension $\tilde{d}>0$, one can find $S \in \mathbb{R}^{M \times (\tilde{d}-1)M}$, $B \in \mathbb{R}^{(\tilde{d}-1)M \times 2M}$, and $A \in \mathbb{R}^{(\tilde{d}-1)M}$, such that $\hat{\mathcal{R}}(\Ub,\Fb):=S\sigma(B[\Ub,\Fb]^{\mathrm{T}} + A)$ satisfies
\begin{equation*}
     ||\mathcal{R}(\Ub,\Fb) -\hat{\mathcal{R}}(\Ub,\Fb)||_{l^2(\mathbb{R}^{M})} = ||\mathcal{R}(\Ub,\Fb) - S\sigma(B[\Ub,\Fb]^{\mathrm{T}} + A) ||_{l^2(\mathbb{R}^M)} \leq \frac{m\varepsilon}{2(1+m)^L},
\end{equation*}
where $m$ is the contraction parameter of $\mathcal{R}$, as defined in Assumption \ref{asp:1}.
%, and 
%$$\Vb_{\Ub,\Fb}:=[\ub(\xb_1),\Fb(\xb_1),\cdots,\ub(\xb_M),\Fb(\xb_M)]$$
%is a re-organized vector depending on $\Ub$ and $\Fb$. 
By this construction, we know that $S$ has independent rows. Hence, there exists the right inverse of $S$, which we denote as $S^{+} \in \mathbb{R}^{(\tilde{d}-1)M \times M}$, such that 
\begin{align*}
    SS^+ = I_M, \quad S^+S  := \tilde{I}_{(\tilde{d}-1)M}, 
    %:= E\oplus\cdots\oplus E, 
    %\begin{bmatrix}
%    I_M & \mathbf{0} \\
%    \mathbf{0} & \mathbf{0} \\
%    \end{bmatrix},
\end{align*}
where $I_M$ is the $M$ by $M$ identity matrix, $\tilde{I}_{(\tilde{d}-1)M}$ is a $(\tilde{d}-1)M$ by $(\tilde{d}-1)M$ block matrix with each of its element being either $1$ or $0$. Hence, for any vector $Z\in\real{(\tilde{d}-1)M}$, we have $\sigma(\tilde{I}_{(\tilde{d}-1)M}Z)=\tilde{I}_{(\tilde{d}-1)M}\sigma(Z)$. Moreover, we note that $S$ has a very special structure: from the $((i-1)(\tilde{d}-1)+1)$-th to the $(i(\tilde{d}-1))$-th column of $S$, all nonzero elements are on its $i$-th row. Correspondingly, we can also choose $S^+$ to have a special structure: from the $((i-1)(\tilde{d}-1)+1)$-th to the $(i(\tilde{d}-1))$-th row of $S^+$, all nonzero elements are on its $i$-th column. Hence, when multiplying $S^+$ with $\Ub$, there will be no entanglement between different components of $\Ub$. That means, $S^+$ can be seen as a pointwise weight function. %formed by a matrix $E:=[1,0,\cdots,0]\in \real^{(d-1)\times 1}$.
%, and $H$ is an invertible matrix that performs multiple row permutation operations.

We now construct the IFNO as follows. In this construction, we choose the feature dimention as $d := \tilde{d}M$. With the input $[\Ub^0,\Fb] \in \mathbb{R}^{2M}$, for the lift layer we set 
$$P := \mathbf{1}_{[M,1]}\otimes\begin{bmatrix}
S^+ & \mathbf{0}\\
\mathbf{0} & I_M\\
    \end{bmatrix}=\underbrace{\begin{bmatrix}
    % S^+ & \mathbf{0}  \\
    % \mathbf{0} & I_M  \\
    % S^+ & \mathbf{0}  \\
    % \mathbf{0} & I_M \\ 
    % \vdots & \vdots \\
    % S^+ & \mathbf{0}  \\
    % \mathbf{0} & I_M \\
S^+ & \mathbf{0}&S^+ & \mathbf{0} & \cdots & S^+ & \mathbf{0}\\
\mathbf{0} & I_M &\mathbf{0} & I_M& \cdots&\mathbf{0} & I_M\\
    \end{bmatrix}^{\mathrm{T}}}_{\text{repeated for }M \text{ times}}\in \real^{dM\times 2M},\;\text{ and }\;\pb := \mathbf{0}\in \real^{dM}.$$
As such, the initial layer of feature is then given by 
    %$$\Hb^0 =\mathcal{P}([\Ub^0,\Fb]^{\mathrm{T}})=[S^+\Ub^0, \Fb ,S^+\Ub^0, \Fb,\dots,S^+\Ub^0, \Fb]^{\mathrm{T}} \in \real^{dM}.$$
    $$\Hb^0 =\mathcal{P}([\Ub^0,\Fb]^{\mathrm{T}})=\mathbf{1}_{[M,1]}\otimes[S^+\Ub^0, \Fb ]^{\mathrm{T}} \in \real^{dM}.$$
Here, we point out that $P$ and $\pb$ can be seen as pointwise weight and bias functions, respectively.

Next we construct the iterative layer $\mathcal{L}^{INFO}$, by setting
$$V := 
    \begin{bmatrix}
    \tilde{I}_{(
    \tilde{d}-1)M}B/M\\
    0 \\
    \end{bmatrix}
    \begin{bmatrix}
    S/\Delta t  & \mathbf{0}\\
    \mathbf{0} & I_M/\Delta t\\
    \end{bmatrix},\;\tilde{V}:=\mathbf{1}_{[M,M]}\otimes V,\;
    % \begin{bmatrix}
    % V&\cdots&V\\
    % \vdots&\vdots&\vdots\\
    % V&\cdots&V\\
    % \end{bmatrix},
    \text{ and }\bm{C} := \mathbf{1}_{[M,1]}\otimes\begin{bmatrix}
    \tilde{I}_{(\tilde{d}-1)M}A/\Delta t \\
    \mathbf{0} \\
%    \vdots \\
%    \tilde{I}_{(\tilde{d}-1)M}A/\Delta t \\
%    \mathbf{0} \\
    \end{bmatrix}.$$
% $$V = 
%     \begin{bmatrix}
%     \tilde{I}_{(
%     \tilde{d}-1)M}B/M\\
%     0 \\
%     \vdots \\
%     \tilde{I}_{(\tilde{d}-1)M}B/M\\
%     0 \\
%     \end{bmatrix}
%     \begin{bmatrix}
%     S/\Delta t  & \mathbf{0} & \dots &S/\Delta t  & \mathbf{0} \\
%     \mathbf{0} & I_M/\Delta t & \dots & \mathbf{0} & I_M/\Delta t\\
%     \end{bmatrix},
%     \text{ and }\bm{C} = \begin{bmatrix}
%     \tilde{I}_{(\tilde{d}-1)M}A/\Delta t \\
%     \mathbf{0} \\
%     \vdots \\
%     \tilde{I}_{(\tilde{d}-1)M}A/\Delta t \\
%     \mathbf{0} \\
%     \end{bmatrix}.$$ 
Note that $\tilde{V}$ falls into the formulation of $V^{IFNO}$, by letting $R_1 =V$  %\begin{bmatrix}
%         \tilde{I}_{(\tilde{d}-1)M}B/M \\
%         0 \\
%     \end{bmatrix}\begin{bmatrix}
%         S/\Delta t & 0 \\
%         0 & I_M/\Delta t \\
%     \end{bmatrix}, 
 and $R_2=R_2=\cdots=R_{M} = W=0$.      
For the $l+1$-th layer of feature vector, we then arrive at 
    \begin{align*}
        \Hb&((l+1)\Delta t) = \Hb(l\Delta t)+ {\Delta t}\sigma\left(\tilde{V}\Hb(l\Delta t)+\Cb\right)\\
    = &\Hb(l\Delta t) + \left(I_M\otimes\begin{bmatrix}
        S^+S  & \mathbf{0} \\
        \mathbf{0} & I_M \\ 
        \end{bmatrix}\right)\sigma \left(\left(\mathbf{1}_{[M,1]}\otimes
        \begin{bmatrix}
    B/M \\
    \mathbf{0} \\
    \end{bmatrix}\right)
    \left(\mathbf{1}_{[1,M]}\otimes\begin{bmatrix}
        S & \mathbf{0}\\
        \mathbf{0} & I_M\\
        \end{bmatrix}\right) \Hb(l\Delta t) + \mathbf{1}_{[M,1]}\otimes\begin{bmatrix}
        A \\
        \mathbf{0}\\
        \end{bmatrix} \right),        
    % = &\Hb(l\Delta t) + \begin{bmatrix}
    %     S^+S  & \mathbf{0} & \dots & \mathbf{0} & \mathbf{0}\\
    %     \mathbf{0} & I_M & \dots & \mathbf{0} & \mathbf{0} \\
    %     \vdots & \vdots & \ddots & \mathbf{0} & \mathbf{0} \\
    %     \mathbf{0} & \mathbf{0} & \dots & S^+S  & \mathbf{0} \\
    %     \mathbf{0} & \mathbf{0} & \dots & \mathbf{0}& I_M  \\ 
    %     \end{bmatrix}\sigma \left(
    %     \begin{bmatrix}
    % B \\
    % \mathbf{0} \\
    % \vdots \\
    % B \\
    % \mathbf{0} \\
    % \end{bmatrix}
    % \begin{bmatrix}
    %     S & \mathbf{0}  & \dots &  S & \mathbf{0}  \\
    %     \mathbf{0} & I_M &\dots & \mathbf{0} & I_M\\
    %     \end{bmatrix} \Hb(l\Delta t) + \begin{bmatrix}
    %     A \\
    %     \mathbf{0}\\
    %     \vdots  \\
    %     A \\
    %     \mathbf{0} \\
    %     \end{bmatrix} \right),
    \end{align*}
where $\Hb(l\Delta t) = [\hat{\hb}_1^{l\Delta t},  \hat{\hb}_2^{l\Delta t},\dots, \hat{\hb}_{2M-1}^{l\Delta t},  \hat{\hb}_{2M}^{l\Delta t}]^{\mathrm{T}}$ denotes the (spatially discretized) hidden layer feature at the $l-$th iterative layer of the IFNO. Subsequently, we note that the second part of the feature vector, $\hat{\hb}_{2j}^{l\Delta t}\in\real^{M}$, satisfies
$$\hat{\hb}_{2j}^{(l+1)\Delta t}=\hat{\hb}_{2j}^{l\Delta t}=\cdots=\hat{\hb}_{2j}^{0}=\Fb, \quad \forall l=0,\cdots,L-1, \forall j = 1,\cdots,M$$
Hence, the first part of the feature vector, $\hat{\hb}_{2j-1}^{l\Delta t}\in\real^{(\tilde{d}-1)M}$, satisfies the following iterative rule:
$$\hat{\hb}_{2j-1}^{(l+1)\Delta t}=\hat{\hb}_{2j-1}^{l\Delta t}+S^+S\sigma(B[S\hat{\hb}_{2j-1}^{l\Delta t}, \Fb ]^{\mathrm{T}}+A), \quad \forall l=0,\cdots,L-1, \forall j = 1,\cdots,M,$$
and
$$\hat{\hb}_{1}^{(l+1)\Delta t}=\hat{\hb}_{3}^{(l+1)\Delta t}=\cdots=\hat{\hb}_{2M-1}^{(l+1)\Delta t}.$$
Finally, for the projection layer $\mathcal{Q}$, we set the activation function in the projection layer as the identity function, $Q_1 := I_{dM}$ (the identity matrix of size $dM$), $Q_2 := [S, \mathbf{0}] \in \real^{M \times dM}$, $\qb_1 := \mathbf{0}\in\real^{dM}$, and $\qb_2 := \mathbf{0}\in\real^{M}$. Denoting the output of IFNO as $\Ub_{IFNO}:=\mathcal{Q}\circ(\mathcal{L}^{IFNO})^L\circ \mathcal{P}([\Ub^0, \Fb ]^{\mathrm{T}})$, we now show that $\Ub_{IFNO}$ can approximate $\Ub^*$ with a desired accuracy $\varepsilon$:
\begin{align*}
    ||\Ub_{IFNO} - \Ub^*|| & \leq ||\Ub_{IFNO} - \Ub^{L}||_{l^2(\mathbb{R}^M)} + ||\Ub^L - \Ub^*||_{l^2(\mathbb{R}^M)}  \\
    & \leq ||S\hat{\hb}_1^{L\Delta t} - \Ub^{L}||_{l^2(\mathbb{R}^M)} + \frac{\varepsilon}{2} \quad (\textit{by Assumption \ref{asp:2}}) \\
    & \leq ||S\hat{\hb}_1^{(L-1)\Delta t} - \Ub^{L-1}||_{l^2(\mathbb{R}^M)} + ||\hat{\mathcal{R}}(S\hat{\hb}_1^{(L-1)\Delta t},\Fb) - \mathcal{R}(\Ub^{L-1},\Fb)||_{l^2(\mathbb{R}^M)}  + \frac{\varepsilon}{2} \\
    & \leq ||S\hat{\hb}_1^{(L-1)\Delta t} - \Ub^{L-1}||_{l^2(\mathbb{R}^M)} + 
     ||\hat{\mathcal{R}}(S\hat{\hb}_1^{(L-1)\Delta t},\Fb) - \mathcal{R}(S\hat{\hb}_1^{(L-1)\Delta t},\Fb)||_{l^2(\mathbb{R}^M)} \\
     & + ||\mathcal{R}(S\hat{\hb}_1^{(L-1)\Delta t},\Fb) - \mathcal{R}(\Ub^{L-1},\Fb)||_{l^2(\mathbb{R}^M)} + \frac{\varepsilon}{2} \\
     & \leq (1+m)||S\hat{\hb}_1^{(L-1)\Delta t} - \Ub^{L-1}||_{l^2(\mathbb{R}^M)} +  \frac{m\varepsilon}{2(1+m)^L}  + \frac{\varepsilon}{2}\quad \textit{(by Lemma \ref{lemma:1} and Assumption \ref{asp:1}})\\
     & \leq \frac{m\varepsilon}{2(1+m)^L}(1+(1+m)+(1+m)^2+\dots+(1+m)^{L-1}) + \frac{\varepsilon}{2} \\
     & \leq \frac{\varepsilon}{2} + \frac{\varepsilon}{2} = \varepsilon.
\end{align*}

\end{proof}

\section{Numerical Examples}\label{sec:experiments}

\begin{table}
\centering
\small
 \begin{tabular}{| c | c | c | c |} 
 \hline
 \textbf{Problem} & \textbf{Data collected from} & \textbf{Input function}& \textbf{Output function}\\
 \hline
Porous medium I & Darcy's equation & Permeability field& Pressure field \\
\hline
\multirow{2}{*}{Porous medium II} & \multirow{2}{*}{Darcy's equation} & Source field \&  & \multirow{2}{*}{Pressure field}\\
&&boundary condition&\\
 \hline
\multirow{2}{*}{Fiber-reinforced material} & Holzapfel-Gasser-Odgen  & \multirow{2}{*}{Boundary condition} & \multirow{2}{*}{Displacement field}\\
& (HGO) model&&\\
 \hline
%\multirow{2}{*}{Glass-ceramics fracture I} & Quasi-static linear & \multirow{2}{*}{Material microstructure}& \multirow{2}{*}{Final damage field}\\
%&peridynamic solid model&&\\
% \hline
\multirow{2}{*}{Glass-ceramics fracture} & Quasi-static linear & Boundary displacement \& & \multirow{2}{*}{Damage field}\\
&peridynamic solid model&previous damage field&\\
 \hline
Latex glove sample& Digital Image Correlation&Boundary condition \&&\multirow{2}{*}{Displacement field}\\
(palm region)&(DIC) displacement tracking&previous displacement field&\\
\hline
\end{tabular}
\caption{Setup for the three numerical examples in Section \ref{sec:experiments} and the application in Section \ref{sec:dic}.}
\label{table:prob_setup}
\end{table}

%In this section, we illustrate the superior performance of NKNs in both learning governing laws and image classification tasks, and compare it to baseline approaches. Our numerical experiments are performed on a machine with 2.8 GHZ 8-core CPU and a single Nvidia V100 GPU.

In this section, we illustrate the performance of the proposed IFNOs on three benchmark material modeling problems: (i) the flow through a porous medium, (ii) the deformation of a hyperelastic and anisotropic fiber-reinforced material, and (iii) the brittle fracture mechanics in glass-ceramics. The detailed settings of each example, including the choices of high-fidelity (ground-truth) training/testing data generation, input function, and output function, are provided in Table \ref{table:prob_setup}. For all numerical experiments, we compare the IFNO to the FNO as the baseline approach, since the other two integral neural operators (i.e., the GKNs and the NKNs) are computationally much more expensive and hence not feasible given the data size and our computational resources. All our numerical experiments were performed on a machine with 2.8 GHz 8-core CPU and a single Nvidia RTX 3060 GPU. For the implementation of the IFNOs and the FNOs, we used the Pytorch package provided in \cite{li2020fourier}. The optimization was performed with the Adam optimizer. To conduct a fair comparison, for each method, we tuned the hyperparameters, including the learning rates, the decay rates and the regularization parameters, to minimize the training loss. Furthermore, for each example and each method, we repeated the numerical experiment for five different random initializations, and reported the averaged relative mean squared errors and their standard errors. For a compact presentation of the results, we reported the relative mean squared errors in plots, as functions of the number of hidden layers ($L$), with error bars representing the standard errors over five simulations. A more detailed error comparison is provided in the appendix.

\subsection{The flow through a porous medium}

\begin{figure}[h!]
    \centering
    \includegraphics[width=.95\textwidth]{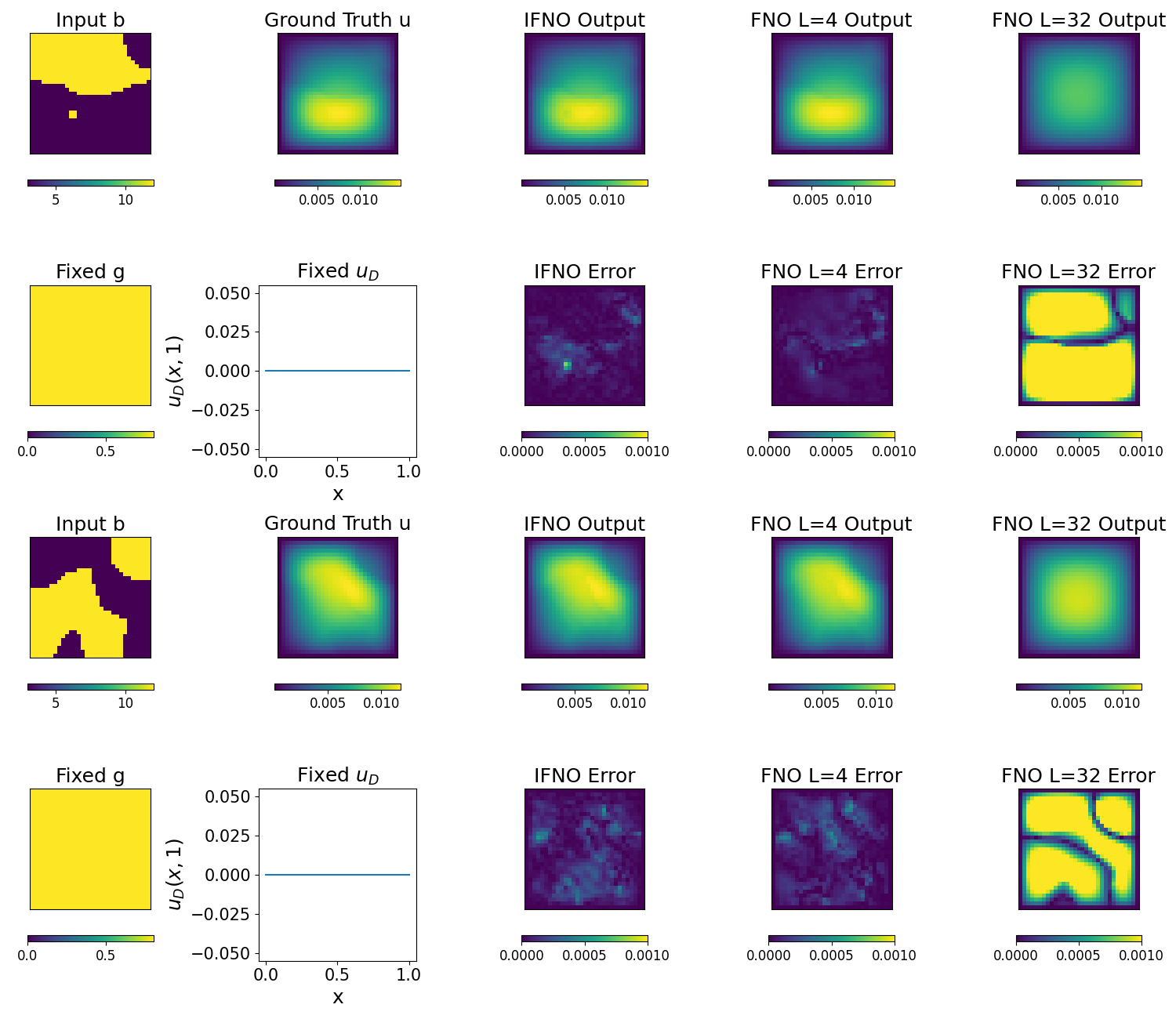}
    \caption{The flow of a fluid through a porous medium setting I, prediction of pressure field from different permeability field $b(\xb)$ and fixed source field $g(\xb)=1$ and boundary condition $u_D(\xb)=0$ (example 1).  A visualization of FNO and IFNO performances on two instances of permeability parameter $b(\xb)$. Here, the best IFNO results ($L=32$), best FNO results ($L=4$), and the deepest FNO results ($L=32$) are reported.}
    \label{fig:s16_test}
\end{figure}

\begin{figure}[h!]
    \centering
    \includegraphics[width=.95\textwidth]{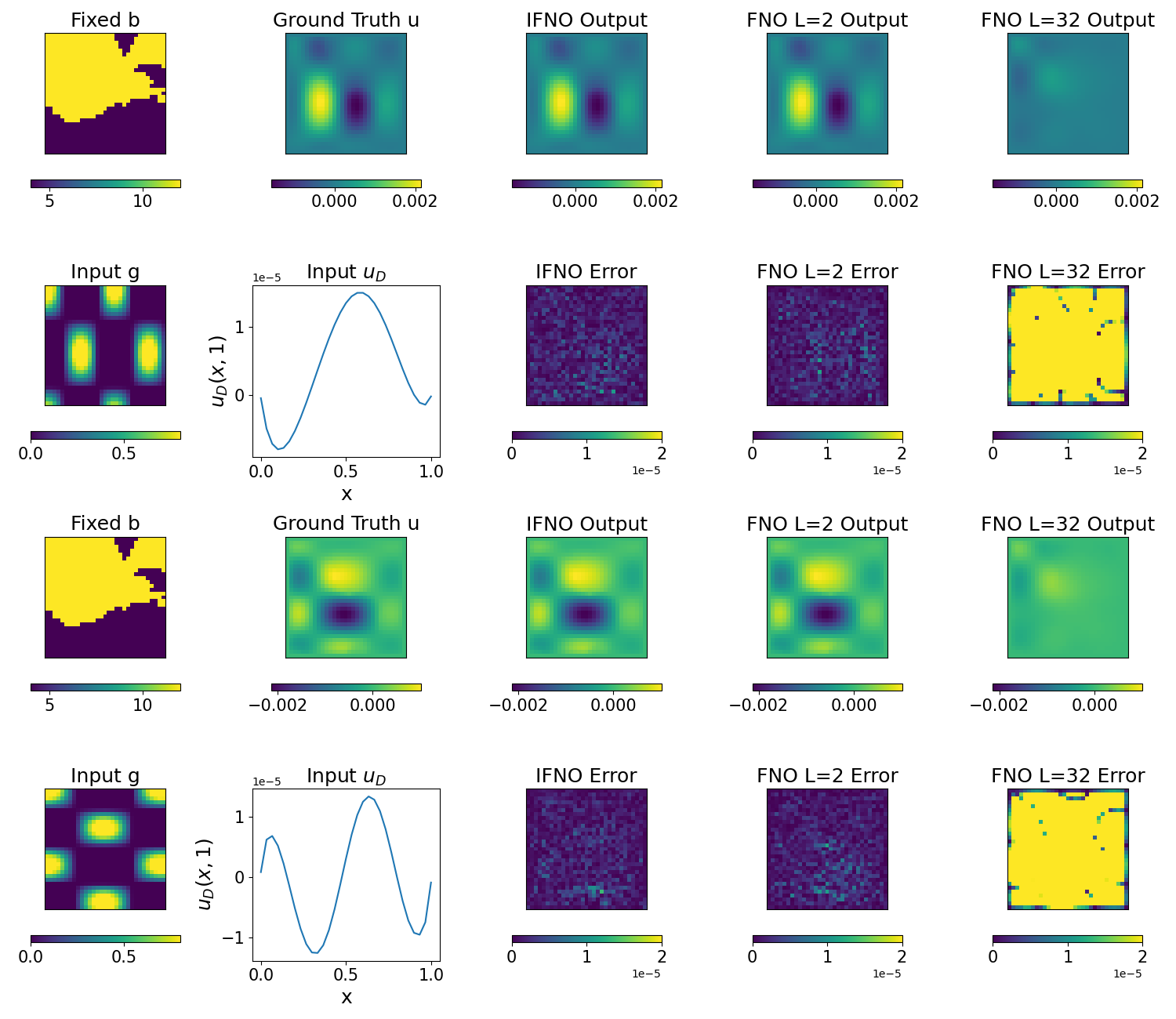}
    \caption{The flow of a fluid through a porous medium setting II, prediction of pressure field on a fixed permeability field $b(\xb)$ and different source field $g(\xb)$ and boundary condition $u_D(\xb)$ (example 1).  A visualization of FNO and IFNO performances on two instances of $g(\xb)$ and $u_D(\xb)$. Here, the best IFNO results ($L=32$), best FNO results ($L=2$), and the deepest FNO results ($L=32$) are reported.}
    \label{fig:s16_test_set2}
\end{figure}

We consider the modeling problem of two-dimensional sub-surface flows through a porous medium with heterogeneous permeability field. Following the settings in \cite{li2020neural}, the high-fidelity synthetic simulation data for this example are described by the Darcy's flow. Here, the physical domain is $D=[0,1]^2$, $b(\xb)$ is the permeability field, and the operator $\mcK_{b}$ is then an elliptic operator associated with $b(\xb)$. In particular, the Darcy's equation has the form:
\begin{align*}
-\nabla\cdot(b(\xb)\nabla u(\xb))=g(\xb),&\quad\xb\in \omg,\\
u(\xb)=u_D(\xb),&\quad\xb\in\partial \omg.
\end{align*}
In this context, our goal is to learn the solution operator of the Darcy's equation and compute the pressure field $u(\xb)$. In this example two study scenarios are considered, corresponding to two different real-world application scenarios:
\begin{enumerate}
    \item (Porous medium I, see Figure \ref{fig:s16_test}) Considering a fixed source field $g(\xb)=1$ and Dirichlet boundary condition $u_D(\xb)=0$, we aim to obtain the pressure field $u(\xb)$ for each permeability field $b(\xb)$. Therefore, the neural operators are employed to learn the mapping from $\fb(\xb):=[\xb,b(\xb)]$ to $u(\xb)$. This setting corresponds to a scenario that the same lab test protocols are applied to heterogeneous material samples with different microstructures, and our learning goal is to predict the material response for a new and unseen sample. Note that this setting is also the benchmark problem considered in a series of integral neural operator studies \cite{li2020neural,li2020multipole,li2020fourier,you2022nonlocal}.
    \item (Porous medium II, see Figure \ref{fig:s16_test_set2}) Considering a fixed permeability field $b(\xb)$, we aim to estimate the pressure field $u(\xb)$ subject to different source fields $g(\xb)$ and Dirichlet boundary conditions $u_D(\xb)$. That means, the neural operators are employed to learn the mapping from $\fb(\xb):=[\xb,g(\xb),\tilde{u}_D(\xb)]$ to $u(\xb)$. This setting corresponds to a scenario that different lab tests are available for a given material sample with unknown microstructure, and our learning goal is to predict the mechanical response of this sample under a new and unseen loading. We note that this setting reflects the typical material mechanical testing experiments, see, e.g., \cite{he2021manifold}, where one representative material sample is tested under several loading protocols and the responses, such as the displacement fields and/or stretch-stress curves, are provided. %\CL{[Chung-Hao, can you check here and add more details?]}
\end{enumerate}

\begin{table}[h]
    \centering
    \begin{tabular}{|c|c|c|c|c|c|c|}
    \hline
    Model & $L=1$ &$L=2$&$L=4$&$L=8$&$L=16$&$L=32$ \\
    \hline
    FNO, setting I &  171.42k & 338.37k & 672.26k & 1.34M & 2.68M & 5.35M\\
    IFNO, setting I &  171.42k & 171.42k & 171.42k & 171.42k & 171.42k & 171.42k\\
    FNO, setting II & 300.48k & 596.45k & 1.19M & 2.37M & 4.74M & 9.48M \\
    IFNO, setting II & 300.48k & 300.48k & 300.48k & 300.48k & 300.48k  & 300.48k \\
    \hline
    \end{tabular}
    \caption{Example 1: the flow of a fluid through a porous medium. Number of trainable parameters for each model.}
    \label{tab:2d_param}
\end{table}

\begin{figure}[h]
    \centering
    \includegraphics[width=0.4\textwidth]{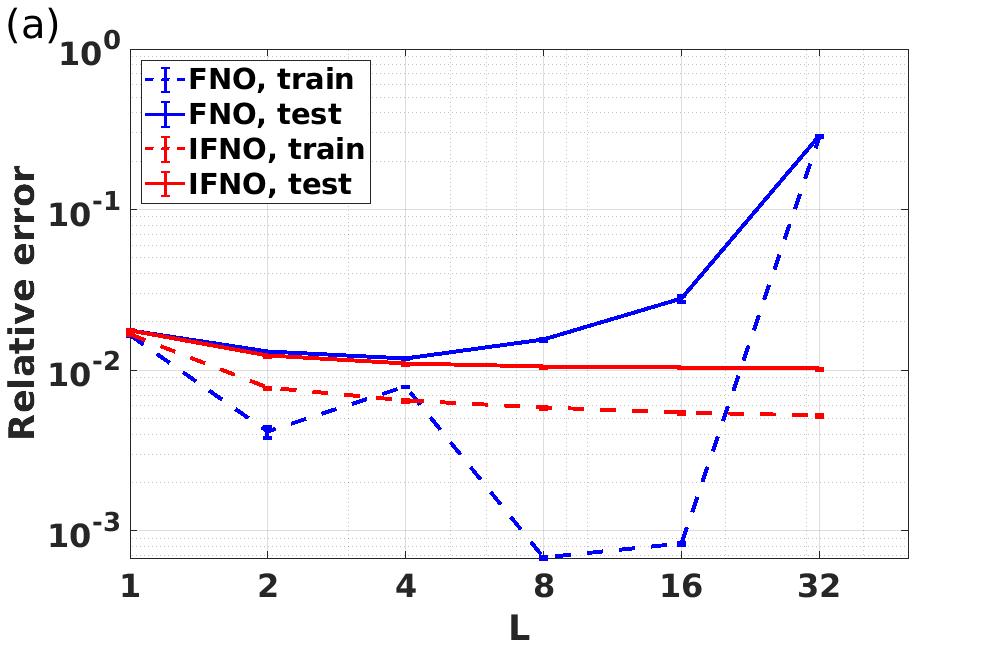}\quad
    \includegraphics[width=0.4\textwidth]{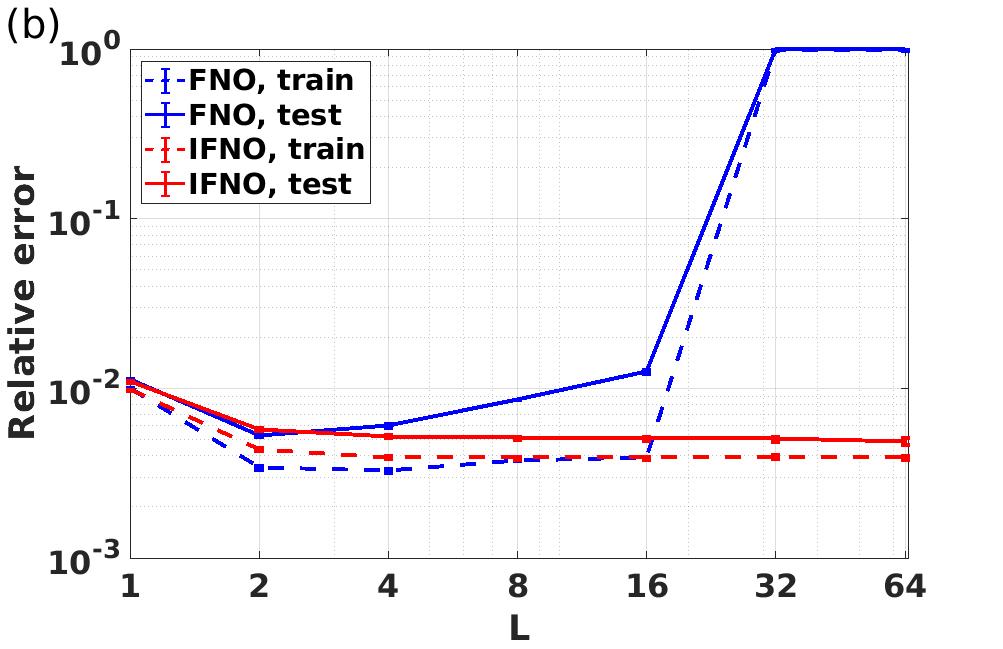}
    \caption{
    The flow of a fluid through a porous medium (example 1). Comparison of relative mean squared errors of pressure field from FNOs and IFNOs. (a) Results from setting I, where neural operators are employed to solve for the corresponding pressure field for each giving permeability field. (b) Results from setting II, where neural operators are employed to solve for the corresponding pressure field with each given pair of source field $g(\xb)$ and Dirichlet boundary condition $u_D(\xb)$.}
    \label{fig:loss_2ddarcy_16}
\end{figure}

\paragraph{Setting and results of porous medium I} As standard simulations of subsurface flow, the permeability $b(\xb)$ is modeled as a two-valued piecewise constant function with random geometry such that the two values have a ratio of 4. Specifically, we generated $1,100$ samples of $b(\xb)$ according to $b\sim \psi_{\#}\mcN(0,(-\Delta+9I)^{-2})$, where $\psi$ takes a value of 12 on the positive part of the real line and a value of 3 on the negative. For cross-validation, the total dataset were divided into a training set with $1,000$ samples and a test dataset with $100$ samples. Then, for each sample the high-fidelity solution of $u$ was generated by using a second-order finite difference scheme to solve the Darcy's equation on a $241\times 241$ grid solution, and both the input and output functions were down-sampled to a structured grid $\chi$ with grid size $\Delta x=1/30$. In this experiment, for both the FNOs and IFNOs, we set the dimension of $\hb$ as {$d=32$}, and the number of truncated Fourier modes as $k=9\times 9$. For each depth $L$, we trained the network for {500 epochs with a learning rate of $1e{-3}$, then decrease the learning rate with a ratio $0.5$ every 100 epochs.} For the IFNOs, the network was trained with the shallow-to-deep training procedure: we initialized the $L-$layer network parameters from the $(L/2)-$layer IFNOs model. This strategy was also employed for other examples in this paper.

In Figure \ref{fig:loss_2ddarcy_16}(a) we report the averaged relative mean squared errors from setting I as a function of iterative layer number $L$; the number of trainable parameters for each model is provided in Table \ref{tab:2d_param}. We can observe that as we increases $L$ to $8$ and $16$, the FNO reaches a relatively low level of error on the training dataset ($O(10^{-4})$). However, the test error of the FNOs deteriorates with the increase of $L$, and reaches $O(10^{-2})$ when $L=8$ or $16$. This indicates that the network is overfitting the training data. Moreover, for $L\geq 32$, the training of the FNOs becomes challenging due to the vanishing gradient phenomenon \cite{hochreiter1998vanishing}. In contrast, the IFNOs trained with the shallow-to-deep initialization are robust and not subject to the overfitting issues: the test error improves as one increases $L$, and stays at a similar magnitude as the training error. Comparing with the FNOs with the same number of layers, the IFNOs have a much smaller number of trainable parameters and lower test errors in all $L>1$ cases. Specifically, the IFNO reaches its best performance when $L=32$, where the averaged (relative) test error is $1.02\%$. On the other hand, the lowest error for the FNO is $1.19\%$, achieved when $L=4$. In Figure \ref{fig:s16_test}, we show the plots of solutions obtained with the best IFNO, the best FNO,  and the deepest FNO, in correspondence of two instances of permeability parameter $b(\xb)$. Both the solutions and the errors are plotted, showing that the FNO loses accuracy when the layer gets deeper ($L=32$), while all other solutions are visually consistent with the ground-truth solutions.

\paragraph{Setting and results of porous medium II} In setting II, we considered a fixed realization of permeability field $b(\xb)$, which was generated following the same procedure as described in setting I. In this context, our goal is to predict the pressure field driven by different source fields $g(\xb)$ and Dirichlet boundary conditions $u_D$. To generate each sample, we set the source field as $g(\xb) = \cos(2\pi a_x x)\cos(2\pi a_y y)$. Here, $a_x$ and $a_y$ are the constant coefficients randomly generated as $a_x,a_y\sim \mathcal{U}(0.5,2)$, the uniform distribution on $[0.5,2]$. To generate the boundary condition $u_D$, we set the pressure on the top edge of the domain as $u_D(x,1) = U_0(t_1\sin(2\pi x)+t_2\sin(4\pi x))/(t_1+t_2)$, where $U_0 \sim \mathcal{U}(-0.001,0.001)$, and $t_1,t_2 \sim \mathcal{U}(0,1)$. On the rest of boundaries, the pressure was prescribed as {$u_D(x,y)=U_0$}.
%The dirichlet boundary conditions are prescribed as $u(x,0) = u(0,y) = u(1,y) = u_0$, and  
For training and cross-validation, we generated 600 samples in total and split it as a training set with 500 samples and a test set with 100 samples. Similar to setting I, the training and test measurements of the pressure fields $u$ were also generated by solving the Darcy's equation and down-sampling to a $M=31\times 31$ grid. In this experiment, for both the FNOs and IFNOs, we set the dimension of $\hb$ as $d=32$, and the number of truncated Fourier modes as $k=12\times 12$. For each depth $L$, we trained the network for 500 epochs with a learning rate of $3e{-3}$, then decrease the learning rate with a ratio $0.5$ every 100 epochs.

In Figures \ref{fig:loss_2ddarcy_16}(b), we report the relative mean squared errors from each model, with hidden layer number $L$ from $1$ to $64$. The number of trainable parameters for each model is provided in Table \ref{tab:2d_param}. Similarly to the porous medium setting I, when increasing the number of layers, the relative test errors of the FNOs deteriorates for $L>2$, after initially decreasing. In contrast, the accuracy of the IFNOs monotonically improves for increasing values of $L$. Also in this case, the FNOs suffer from the vanishing gradient: the training becomes challenging when $L>16$. In Figure \ref{fig:s16_test_set2}, we depict both solutions and prediction errors obtained with the best IFNO, the best FNO, and the deepest FNO, in correspondence of two pairs of source field $g(\xb)$ and boundary condition $u_D(\xb)$. In particular, in this setting, the IFNO reaches its best test error, $0.49\%$, when $L=32$. For the FNO, the best performance is achieved when $L=2$, where the test error is $0.53\%$. When $L>2$, The IFNOs consistently outperforms the FNOs in the testing experiments.

%\paragraph{Porous medium setting II}

\subsection{The deformation of a hyperelastic and anisotropic fiber-reinforced material}

\begin{figure}[h!]
\centering
\subfigure{\includegraphics[width=.55\columnwidth]{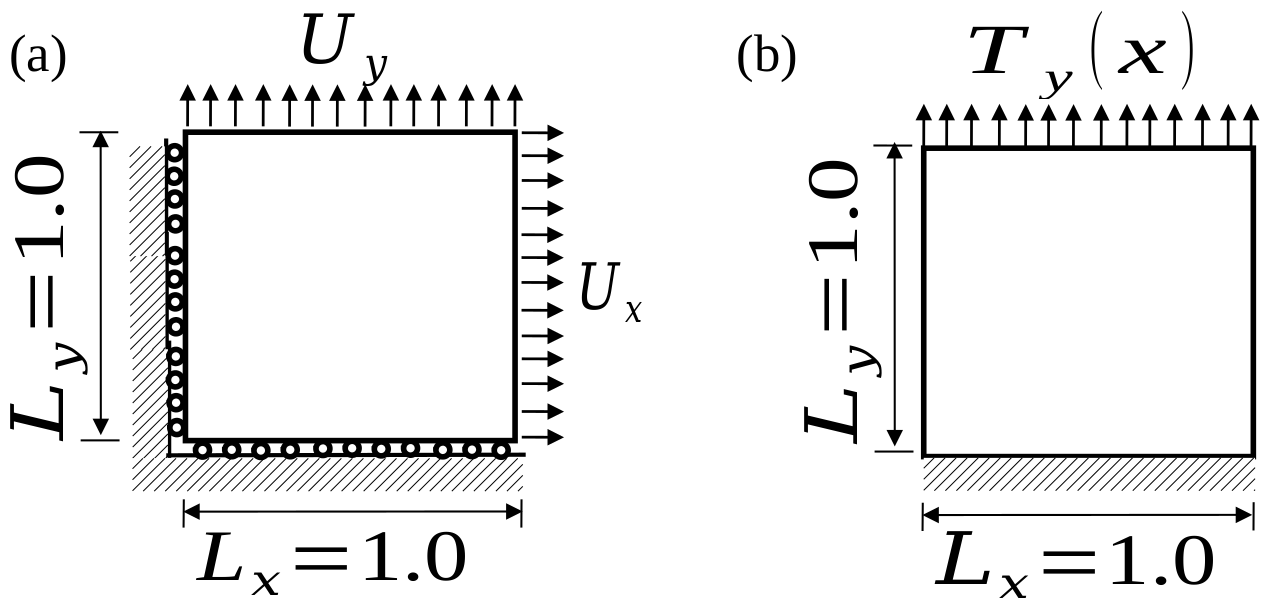}}
\caption{Problem setup of example 2: the deformation of a hyperelastic and anisotropic fiber-reinforced material. (a) A unit square subject to biaxial stretching with Dirichlet-type boundary conditions. (b) A unit square subject to uniaxial tension with Neumann-type boundary condition.}
\label{fig:hgosetup}
\end{figure}

%% constitutive
\begin{table}
\centering
%\begin{adjustbox}{}%{width=0.4\textwidth}
 \begin{tabular}{|c | c ccccc|} 
 \hline
 Parameter &  $c_{10}$ & $K$ & $k_1$& $k_2$&$\alpha$ &$\kappa$\\
 \hline
 Value & 0.3846 & 0.8333& 0.1& 1.5& $\pi$/2&0\\
 \hline
\end{tabular}
%\end{adjustbox}
\caption{Parameter values of the HGO model for data generation in example 2.}
\label{table:hyper_param}
\end{table}

    \begin{table}[]
        \centering
        \begin{tabular}{|c|c|c|c|}
        \hline
        Set ID& Protocol   & max $U_x$ on the right edge & max $U_y$ on the top edge\\
        \hline
        1& Biaxial Stretch $1:1$ & 0.4 & 0.4 \\
        2& Biaxial Stretch $0.66:1$ & 0.4 & 0.6  \\
        3& Biaxial Stretch $0.5:1$  & 0.2 & 0.4  \\
        4& Biaxial Stretch $0.33:1$ & 0.2 & 0.6  \\
        5& Biaxial Stretch $1:0.66$ & 0.6 & 0.4  \\
        6& Biaxial Stretch $1:0.5$ & 0.4 & 0.2 \\
        7& Biaxial Stretch $1:0.33$ & 0.6 & 0.2  \\
        8& Uniaxial Stretch in $x$ & 0.4 & 0 \\
        9& Uniaxial Stretch in $y$ & 0 & 0.4 \\
        \hline 
        \end{tabular}
        \caption{Nine protocols of the synthetic biaxial mechanical testing on a $1\times 1$ hyperelastic and anisotropic fiber-reinforced material sample. All simulations are generated with FEniCS \cite{alnaes2015fenics} based on the HGO model.}
        \label{tab:HGO_setting}
    \end{table}

% \begin{figure}
%     \centering
%     \includegraphics[width=0.46\textwidth]{figures/hgo_train_fnoboth_neu.jpg}
%     \includegraphics[width=0.52\textwidth]{figures/HGO_neumann_sample.png}
%     \caption{
%     Example 2: the deformation of a hyperelastic and anisotropic fiber-reinforced material driving by traction loadings. Left: comparison of relative mean squared errors of displacement field from FNOs and IFNOs. Right: a visualization of $L=32$ IFNO performances on two instances of traction loads on the top edge.}
%     \label{fig:loss_hgo_neu}
% \end{figure}

We now consider the modeling problem of a hyperelastic, anisotropic, fiber-reinforced material, and seek to find its displacement field $\ub:[0,1]^2\rightarrow\real^2$ under different boundary loadings. To generate training and test samples, the Holzapfel-Gasser-Odgen (HGO) model~\cite{holzapfel2000new} was employed to describe the constitutive behavior of the material in this example, with its strain energy density function given as:
\begin{align*}
    \eta & = \frac{c_{10}}{2}(\overline{I}_{1} - 3) - c_{10}\ln(J) + \frac{k_{1}}{2k_{2}}\sum^{2}_{i=1}(\exp{(k_{2}\langle E_{i} \rangle^{2}}) - 1) + \frac{K}{2}\left( \frac{J^{2} - 1}{2} - \ln{J} \right).
\end{align*}
Here, $\langle \cdot \rangle$ denotes the Macaulay bracket, and the fiber strain of the two fiber groups is defined as:
\begin{equation*}\label{eqn:fiberstrain}
    E_{i} = \kappa (\overline{I}_{1} - 3) + (1 - 3\kappa)(\overline{I}_{4i} - 1),\quad i=1,2,
\end{equation*}
where $k_{1}$ and $k_{2}$ are fiber modulus and the exponential coefficient, respectively, $c_{10}$ is the moduli for the non-fibrous ground matrix, $K$ is the bulk modulus, and $\kappa$ is the fiber dispersion parameter. Moreover, $\overline{I}_{1}=\text{tr}(\mathbf{C})$ is the is the first invariant of the right Cauchy-Green tensor $\mathbf{C}=\mathbf{F}^T\mathbf{F}$, $\mathbf{F}$ is the deformation gradient, and $J$ is related with $\mathbf{F}$ such that $J = \det \mathbf{F}$. For the $i-$th fiber group with angle direction $\alpha_i$ from the reference direction, $\overline{I}_{4i}=\mathbf{n}_i^T\mathbf{C}\mathbf{n}_i$ is the fourth invariant of the right Cauchy-Green tensor $\mathbf{C}$, where $\mathbf{n}_i=[\cos(\alpha_{i}), \sin(\alpha_{i})]^{T}$. In our simulations, we considered a material with fiber reinforcement in the vertical direction, and set the orientation for both fiber groups as $\alpha_{i}=\pi/2$. All parameter values are summarized in Table~\ref{table:hyper_param}.

In this example, our goal is to learn the solution operator of the HGO 
model, and predict the displacement field $\ub(\xb)$ subject to 
different boundary conditions. As depicted in Figure \ref{fig:hgosetup}, two types of boundary conditions are considered: (i) the 
Dirichlet-type boundary condition where a uniform uniaxial displacement
loading was applied on the right and top edges of the plate (see Figure \ref{fig:hgosetup}(a)); and (ii) the Neumann-type boundary loading 
where we applied a uniaxial tension $\tb(\xb)$ on the top edge (see Figure \ref{fig:hgosetup}(b)). For both cases, to generate the high-fidelity (ground-truth) dataset, we solved the displacement field on the entire domain by minimizing potential energy using the finite element method implemented in FEniCS \cite{alnaes2015fenics}. In particular, the displacement filed was approximated by continuous piecewise linear finite elements with triangular mesh, and the grid size was taken as $0.025$. Then, the finite element solution was interpolated onto $\chi$, a structured $41 \times 41$ grid which will be employed as the discretization in our neural operators.
%. We take the interpolated finite element solution as the ground-truth solution.}

\begin{figure}[h!]
    \centering
    \includegraphics[width=0.4\textwidth]{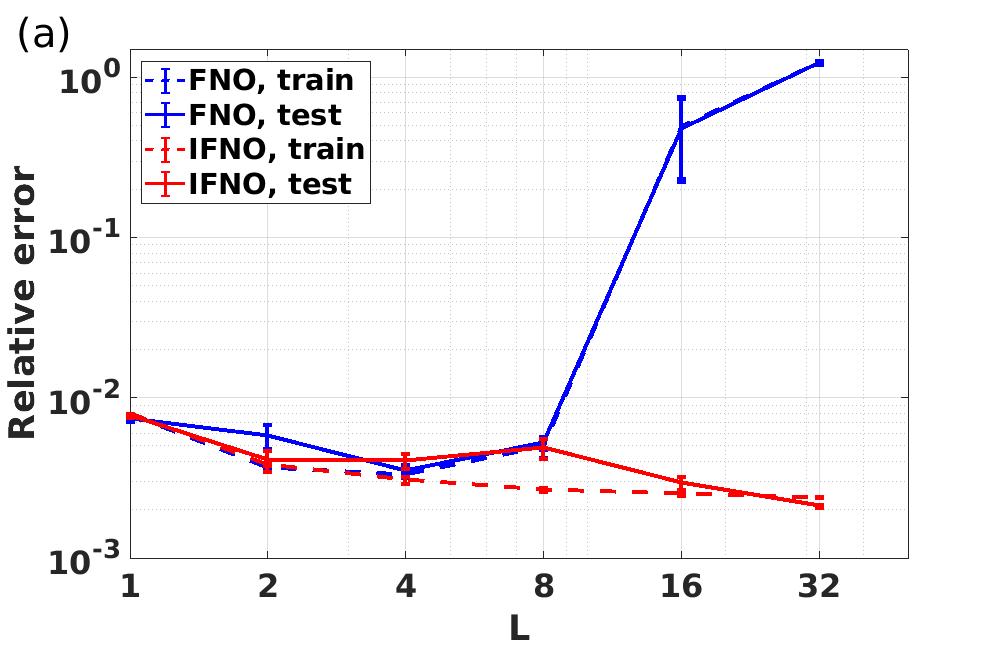}\quad
    \includegraphics[width=0.4\textwidth]{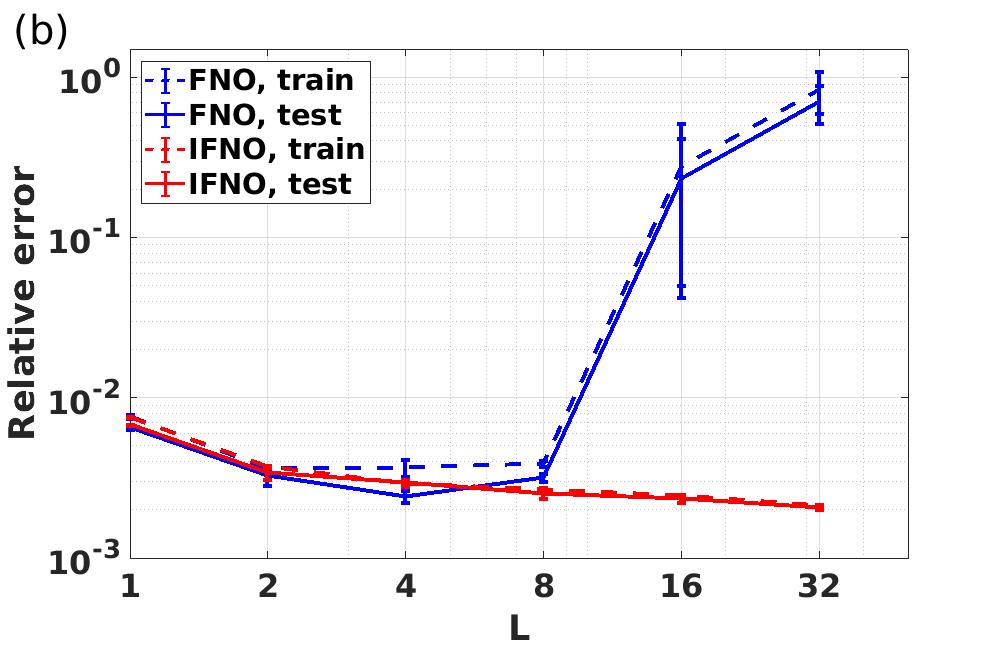}
    \includegraphics[width=0.4\textwidth]{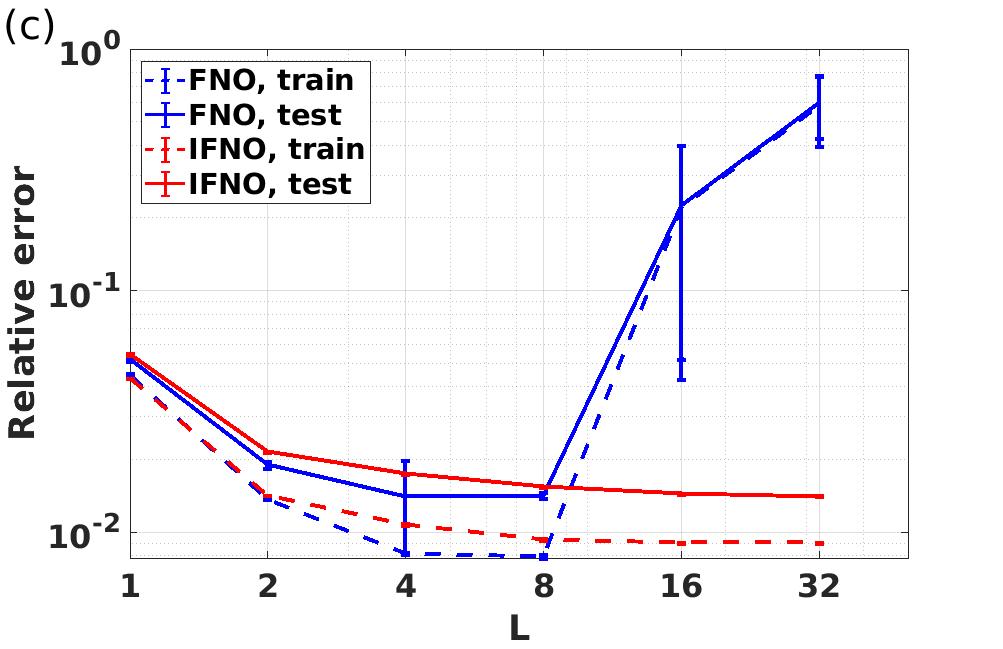}
    \includegraphics[width=0.58\textwidth]{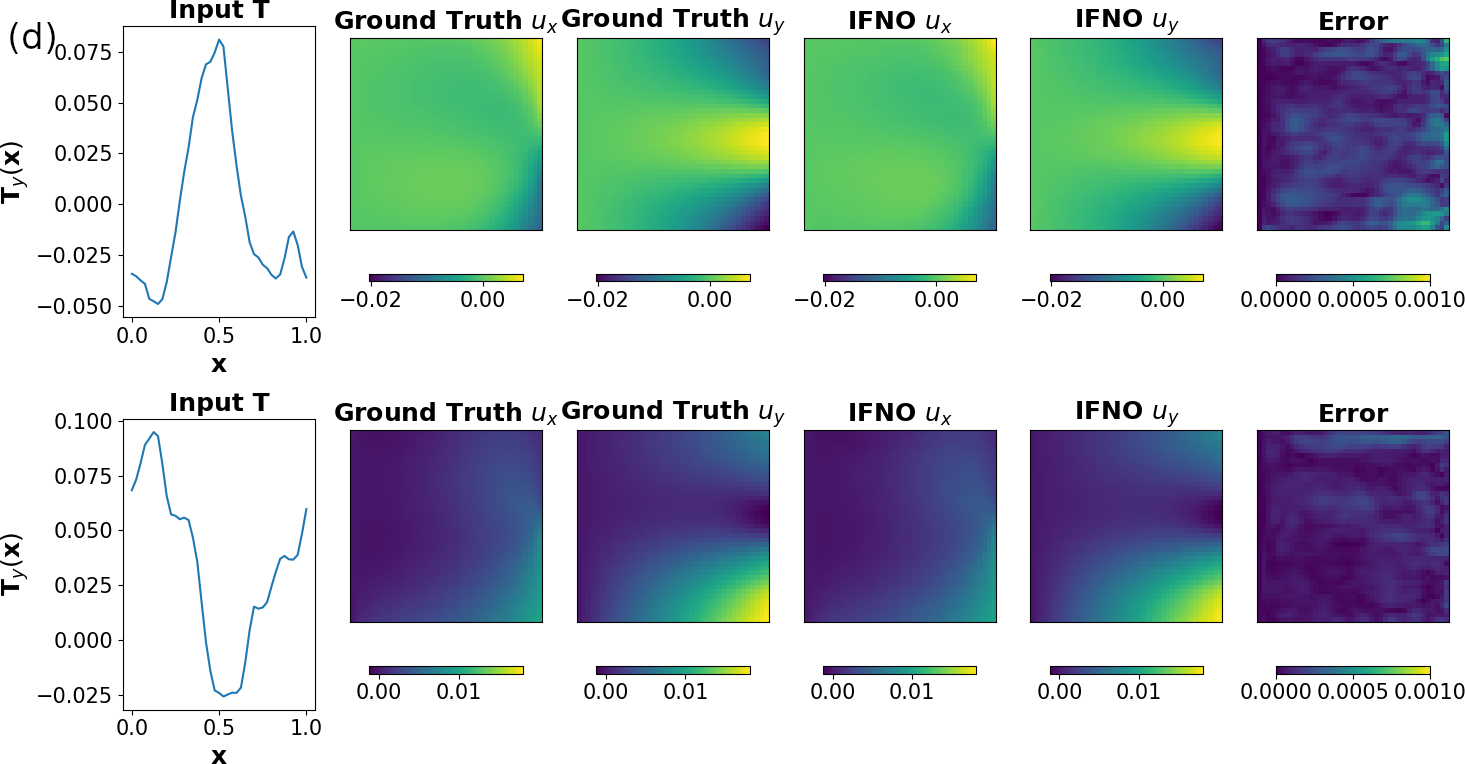}
    \caption{
    The deformation of a hyperelastic and anisotropic fiber-reinforced material (example 2). (a-b) Comparison of relative mean squared errors of displacement field predictions driving by displacement boundary conditions. (a) Results from in-distribution tests, where the testing boundary conditions are \textbf{inside} the training region. (b) Results from out-of-distribution tests, where the testing boundary conditions are \textbf{outside} the training region. (c-d) Results of displacement field predictions driving by traction conditions. (c) Comparison of relative mean squared errors. (d) A visualization of $L=32$ IFNO performances on two instances of traction loads on the top edge.}
    \label{fig:loss_hgo}
\end{figure}

\paragraph{Learning material responses from displacement boundary conditions} We first studied the performance of the IFNOs as a solution operator under Dirichlet-type boundary conditions. To mimic the real-world mechanical test settings (see, e.g. \cite{he2021manifold}), we generated $9$ different biaxial loading protocol sets as listed in Table \ref{tab:HGO_setting}, with $100$ samples for each set. For each sample, a uniform uniaxial displacement boundary condition $\ub_D=(U_x,0)$ was applied on the right edge of the plate, and another uniform uniaxial displacement $\ub_D=(0,U_y)$ was prescribed on the top edge. The other two edges were set as clamped on the tangential direction. Based on this boundary condition, we generated the displacement field solution $\ub(\xb)$ using FEniCS, serving as the high-fidelity solution. Then, the neural operators were employed to learn the mapping from $\fb(\xb):=[\xb,\tilde{U}_x,\tilde{U}_y]$ to $\ub(\xb):=[u_x(\xb),u_y(\xb)]$, where $\tilde{U}_x$ and $\tilde{U}_y$ are the padded boundary conditions, as described in \eqref{eqn:pad}. Two study scenarios were considered to evaluate the in-distribution prediction capability and the out-of-distribution generalizability of the proposed neural operators:
\begin{enumerate}
    \item We randomly selected $100$ samples as the test dataset from the $900$ total samples, and used all other samples to form the training dataset. In this scenario, we note that the boundary conditions of test samples are inside the training region.
    \item We used protocol set \#4 ($0.33:1$ biaxial tension) as the test dataset, and all other sets as the training dataset. Note that that the $0.33:1$ biaxial tension protocol is not covered in any of other sets. Therefore, with this scenario we aim to study the generalizability of the proposed method by testing with boundary conditions outside the training region.
\end{enumerate}
For both study scenarios, we set the dimension of $\hb$ as $d=32$ and the number of truncated Fourier modes as $k=8\times 8$ in all neural operator models. For this example, we trained the network for 500 epochs with a learning rate of $5e-3$, then decreased the learning rate with a ratio of 0.5 every 100 epochs.

In Figure \ref{fig:loss_hgo}(a-b), we provide the averaged relative mean squared errors as functions of hidden layer numbers $L$. In Figure \ref{fig:loss_hgo}(a), we depict the results from scenario 1. We observed that when $L>4$, both the training and testing errors from the FNOs start to increase, due to the vanishing gradient issue. A similar phenomenon is observed in Figure \ref{fig:loss_hgo}(b), where the results from scenario 2 are provided. In contrast, the accuracy of the IFNOs monotonically improves for increasing values of $L$. For the in-distribution test scenario, the IFNOs reaches its lowest test error ($0.21\%$) at $L=32$, which almost halved the optimal error from FNOs ($0.35\%$ at $L=4$). Similarly, for the out-of-distribution scenario, the best performance for the IFNO is obtained at $L=32$, and the test error is $0.21\%$. In the mean time, the optimal FNO is still with $L=4$, and achieved a slightly larger test error ($0.24\%$). When comparing between the IFNO and the FNO with the same depth, the IFNO again achieves a better accuracy whenever the network is deeper than $4$. Different from example 1, in this example we did not observe much overfitting problem, possibly due to the fact that the material microstructure and loading settings have low complexity: the material is assumed to be homogeneous, and the displacement-type boundary conditions are uniform. All these facts are anticipated to reduce the complexity of this learning task, so the material responses in testing datasets do not vary much from the responses in the training dataset, even in the out-of-distributing prediction scenario.

\paragraph{Learning material responses from traction boundary conditions} In the previous examples and experiments, we have investigated the performance of integral neural operators on predicting material responses driven by Dirichlet-type boundary conditions. Here, we further studied the material deformation driven by Neumann-type boundary conditions, as depicted in Figure \ref{fig:hgosetup}(b). 
%Specifically, we considered a plate which is clamped on the bottom edge, with a vertical traction $\tb(\xb)$ distributed on the top edge. 
With the IFNOs, we aim to learn the solution operator which predicts the resultant displacement field $\ub(\xb)$ driven by different traction boundary conditions $\tb(\xb)=[0,T_y(\xb)]$. In this context, the input function is $\fb(\xb):=[\xb,\tilde{T}_y(\xb)]$, where $\tilde{T}_y(x,y):={T}_y(x,1)$ is the padded function of $T_y(\xb)$ onto the whole domain $\omg$. The output function is the displacement field. To generate the training/testing dataset, we sampled $1,000$ different vertical traction conditions $T_y(\xb)$ on the top edge from a random field, following the algorithm in \cite{LangPotthoff2011,yin2022interfacing}. In particular, $T_y(\xb)$ is taken as the restriction of a 2D random field, $\phi(\xb) = \mathcal{F}^{-1}(\gamma^{1/2}\mathcal{F}(\Gamma))(\xb)$, on the top edge. Here, $\Gamma(\xb)$ is a Gaussian white noise random field on $\real^2$, $\gamma=(w_1^2+w^2_2)^{-\frac{5}{4}}$ represents a correlation function, and $w_1$, $w_2$ are the wave numbers on $x$ and $y$ directions, respectively. Then, for each sampled traction loading, we performed a FEniCS simulation based on the HGO model, to obtain the solutions in the entire domain and collect the corresponding solutions of displacement fields of in $\Omega$. Among these $1,000$ samples, $800$ cases were employed as the training data while the rest was kept as testing data. {In this setting, we train the network for 500 epochs with a learning rate of 5e-3, then decreased the learning rate with a ratio of 0.5 every 100 epochs.}

In Figure \ref{fig:loss_hgo}(c), we show the relative mean squared errors from each neural operator model with respect to different hidden layer numbers $L$. Similarly to the Dirichlet-type boundary cases, the IFNO achieves its best performance at $L=32$, while the FNO suffers from vanishing gradient when $L>8$. In Figure \ref{fig:loss_hgo}(d), we compare the horizontal displacement $u_x$ and vertical displacement $u_y$ in $\Omega$ between the FEniCS ground truth and the $L=32$ IFNO prediction, together with the prediction errors. To illustrate the network generalizability to different traction boundary conditions, the results on two instances of $T_y(\xb)$ among the test samples are illustrated. We observe that the IFNO predictions match well with the ground truth solution, demonstrating the capability of our proposed method in predicting material responses driven by unseen traction conditions.

\subsection{The brittle fracture mechanics in glass-ceramics}

\begin{figure}[h!]
\centering
\subfigure{\includegraphics[width=.5\columnwidth]{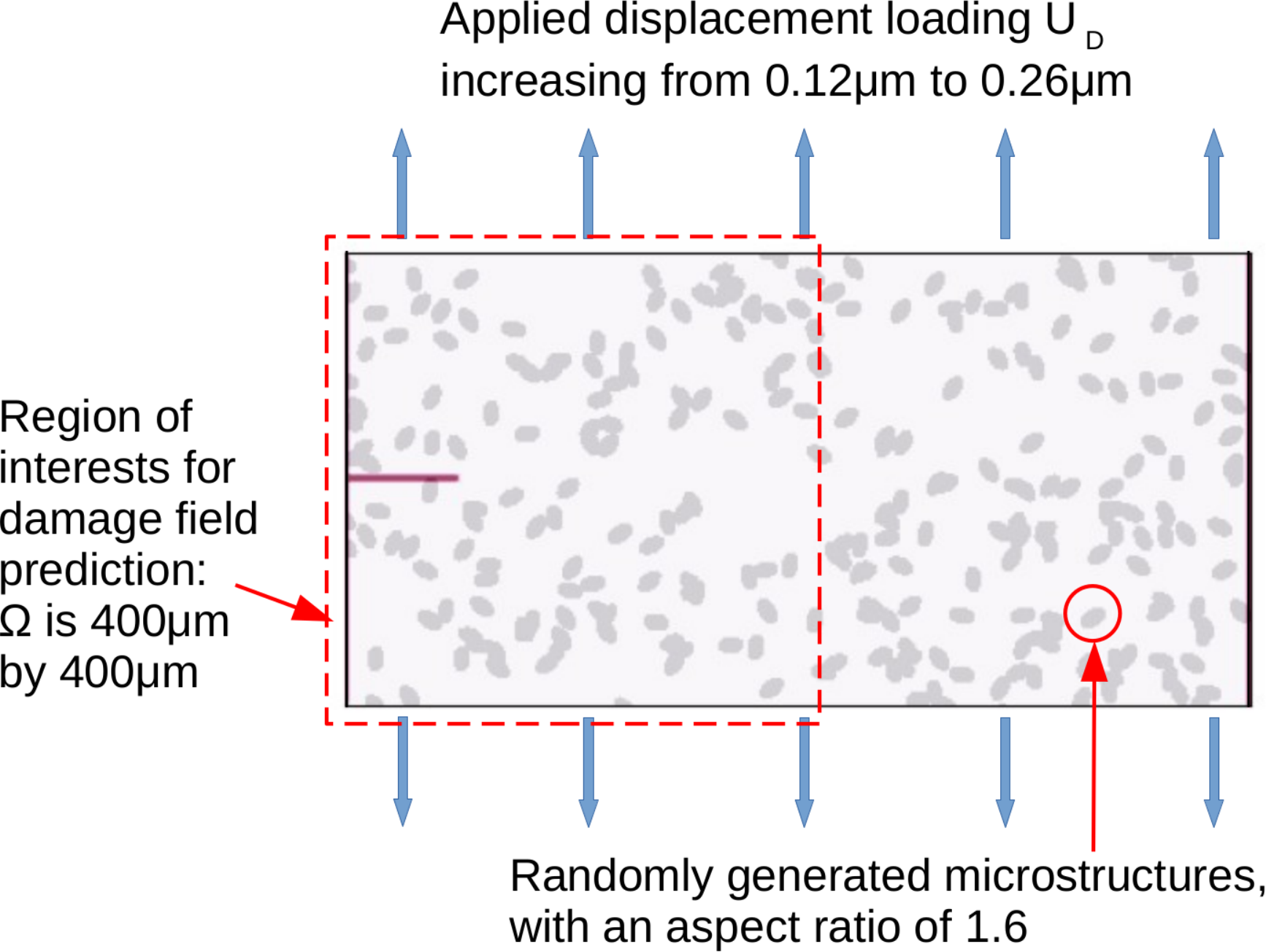}}
\caption{Problem setup of pre-cracked glass-ceramics experiment with randomly distributed material property fields and the plate microstructure considered in example 3, following \cite{serbena2015crystallization}. Here, dark grey represents the crystalline and light grey represents the glassy matrix. This microstructure represents a glass-ceramic sample where the crystals occupy $20\%$ of the volume.}
\label{fig:glasssetup}
\end{figure}

\begin{table}[h]
\center
\begin{tabular}{|c|cccc|}
\hline
&Young's modulus& Poisson ratio& Fracture energy& Fracture Toughness\\
\hline
Glass&$E_1=$80\,GPa&0.25&$G_1=$6.59\,J/m$^2$ & 0.75\,MPa$\cdot \sqrt{\rm {m}}$\\
Crystal&$E_2=$133\,GPa&0.25&$G_2=$86.35\,J/m$^2$ & 3.5\,MPa$\cdot \sqrt{{\rm m}}$\\
\hline
\end{tabular}
\caption{Material parameters used for generating the high-fidelity solution in the pre-cracked glass-ceramics experiment, following \cite{serbena2015crystallization}.}
\label{tab:glass}
\end{table}

In this example, we study the problem of brittle fracture in a glass-ceramic material, as a prototypical exemplar on the heterogeneous material damage field prediction. A glass-ceramic material is the product of controlled crystallization of a specialized glass composition, which results in the creation of a microstructure composing of one of more crystalline phases within the residual amorphous glass \cite{prakash2022investigation,serbena2012internal,holand2019glass,fu2017nature,fan2022meshfree}. In glass-ceramics, the material has enhanced strength and toughness compared to pure glass, while the microstructure and phase assemblage of each material sample play a vital role in determining material strength and toughness. 
%Therefore, it is important to investigate the microstructure of these materials and their relation to damage metrics of interests to gain fundamental insight \cite{serbena2015crystallization}. In particular, we employed the proposed approach to study the {fracture toughness of a model glass-ceramic material (lithium disilicate) as a function of crystal volume fraction} \cite{serbena2015crystallization}.

We considered a pre-notched idealized microstructural realization which is subject to displacement {boundary conditions} on its top and bottom {boundaries}. As demonstrated in Figure \ref{fig:glasssetup}, a plate of dimensions $800\,\mu$m by $400\,\mu$m was considered, with an initial crack of length $100\,\mu$m, and a gradually increasing uniform displacement loading $U_D$ applied on the top and bottom of the sample. All other boundaries, including the new boundaries created by cracks, were treated as free surfaces. This microstructure realization is composed of randomly distributed crystals embedded in a glassy matrix, such that the crystals occupy $20\%$ of the volume. Similarly to \cite{serbena2015crystallization,prakash2022investigation,fan2022meshfree}, we generated the center location  $(C_x,C_y)$ and rotation angle $C_\eta$ of each crystal as random variables, satisfying $C_x\sim \mathcal{U}(0,800)$, $C_y\sim \mathcal{U}(0,400)$, and $C_\eta\sim \mathcal{U}(0,2\pi)$. All crystals are identical ellipses with semi-major and semi-minor axes being $12\,\mu$m and $7.5\,\mu$m, respectively, with an aspect ratio of $1.6$. The mechanical properties of glass and crystalline phases are summarized in Table \ref{tab:glass}. This material was studied experimentally in \cite{serbena2015crystallization} and numerically in \cite{prakash2022investigation,fan2022meshfree} for different crystallized volume fractions. Here, we adopted the setting in \cite{fan2022meshfree} and employed the quasi-static linear peridynamic solid (LPS) model to generate the high-fidelity simulation data. In particular, for each microstructure realization, we used $R(\xb)$ to denote the microstructure, such that 
\begin{equation}
R(\xb)=\left\{\begin{array}{cc}
0     &\text{ if the material point $\xb$ is glass,}  \\
1     &\text{ if the material point $\xb$ is crystal.} 
\end{array}\right.
\end{equation}
For this microstructure sample, the field of Young's modulus $E(\xb)$ and fracture energy $G(\xb)$ can be represented as linear transformations of $R$:
$$E(\xb)=R(\xb)(E_2-E_1)+E_1,\quad G(\xb)=R(\xb)(G_2-G_1)+G_1,$$
where $E_1$, $E_2$ are the Young's modulus of glass and crystal, respectively, and $G_1$, $G_2$ are their repsective fracture energy. The high-fidelity material responses and crack propagation simulations in this sample are calculated using the LPS model proposed in \cite{fan2022meshfree}:
\begin{equation}\label{eqn:probn_rand}
\begin{array}{ll}
\mcK_{R}\ub(\xb,\tau) = 0,&\quad \xb\in\omg\\
%\theta(\xb,t)=\int_{B_\delta (\xb)} \gamma(\xb,\yb,t)K(\left|\yb-\xb\right|) (\yb-\xb)^T \mathbf{M}(\xb,t)\left(\ub(\yb,t) - \ub(\xb,t) \right)d\yb,&\quad \text{ in }\omg\cup\omgbb_D\\
\ub(\xb,\tau)=\ub_D(\xb,\tau), &\quad \xb\in\omgbb_D
\end{array}
\end{equation}
where $\omgbb_D$ denotes the nonlocal boundary layer on the top and the bottom edges of the plate, the instant $\tau$ denotes the indexes for (incrementally increasing) loading. In particular, we set $\ub_D(\xb,\tau)=[0,U_D(\tau)]$ on the top edge, and $\ub_D(\xb,\tau)=[0,-U_D(\tau)]$ on the bottom edge. To perform quasi-static simulations of crack propagation, we gradually increase $U_D$ from $0.12\,\mu$m to $0.26\,\mu$m, and simulate the propagation of the crack starting from the pre-crack tip till it reaches the right boundary of the domain. At each quasi-static step, we increased $U_D$ by $0.002\,\mu$m, performed subiterations until no new broken bonds are detected, and then proceeded to the next step. For spatial discretization, we employed uniform grids with grid size $\Delta x=2\,\mu$m. Therefore, the whole computational domain $\omg\cup\omgbb_D$ has $87,969$ grid points in total. To generate the training and testing samples, we employed the meshfree method proposed in \cite{fan2022meshfree} to solve for the displacement field $\ub(\xb,\tau)$ and the damage field $d(\xb,\tau)$. For the detailed formulation of the LPS operator $\mcK_{R}$ and the numerical method, we refer interested readers to \cite{fan2022meshfree}.

\begin{figure}
    \centering
\subfigure{\includegraphics[width=.4\columnwidth]{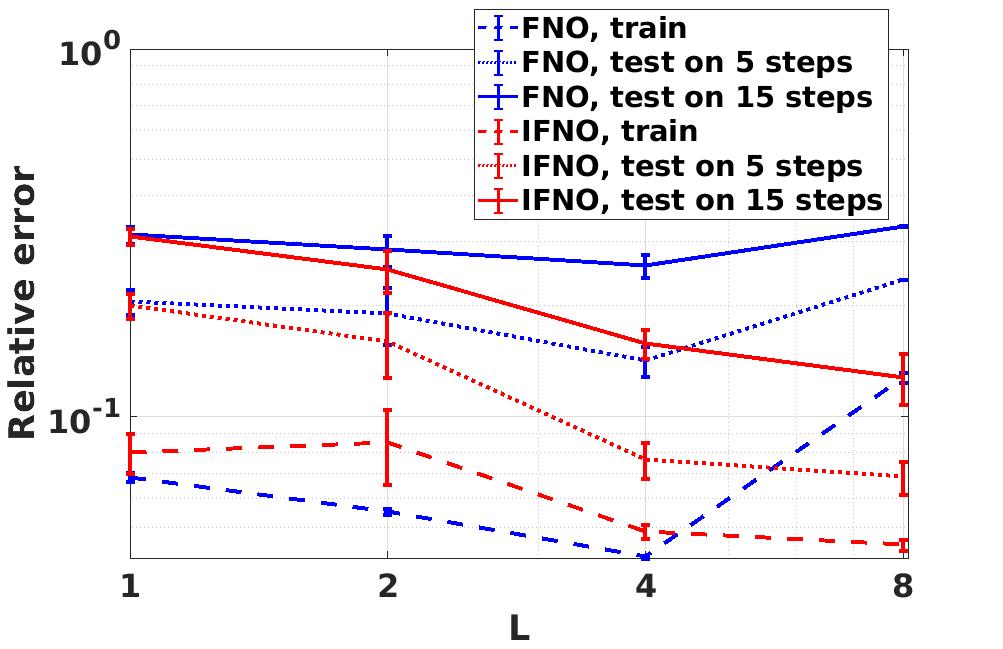}}
    \caption{
    The glass-ceramic crack propagation problem (example 3). Comparison of relative mean squared errors for quasi-static damage field prediction on a fixed microstructure field and increasing boundary displacement loading.}
    \label{fig:loss_lps}
\end{figure}

\begin{figure}
    \centering
    \includegraphics[width=0.96\textwidth]{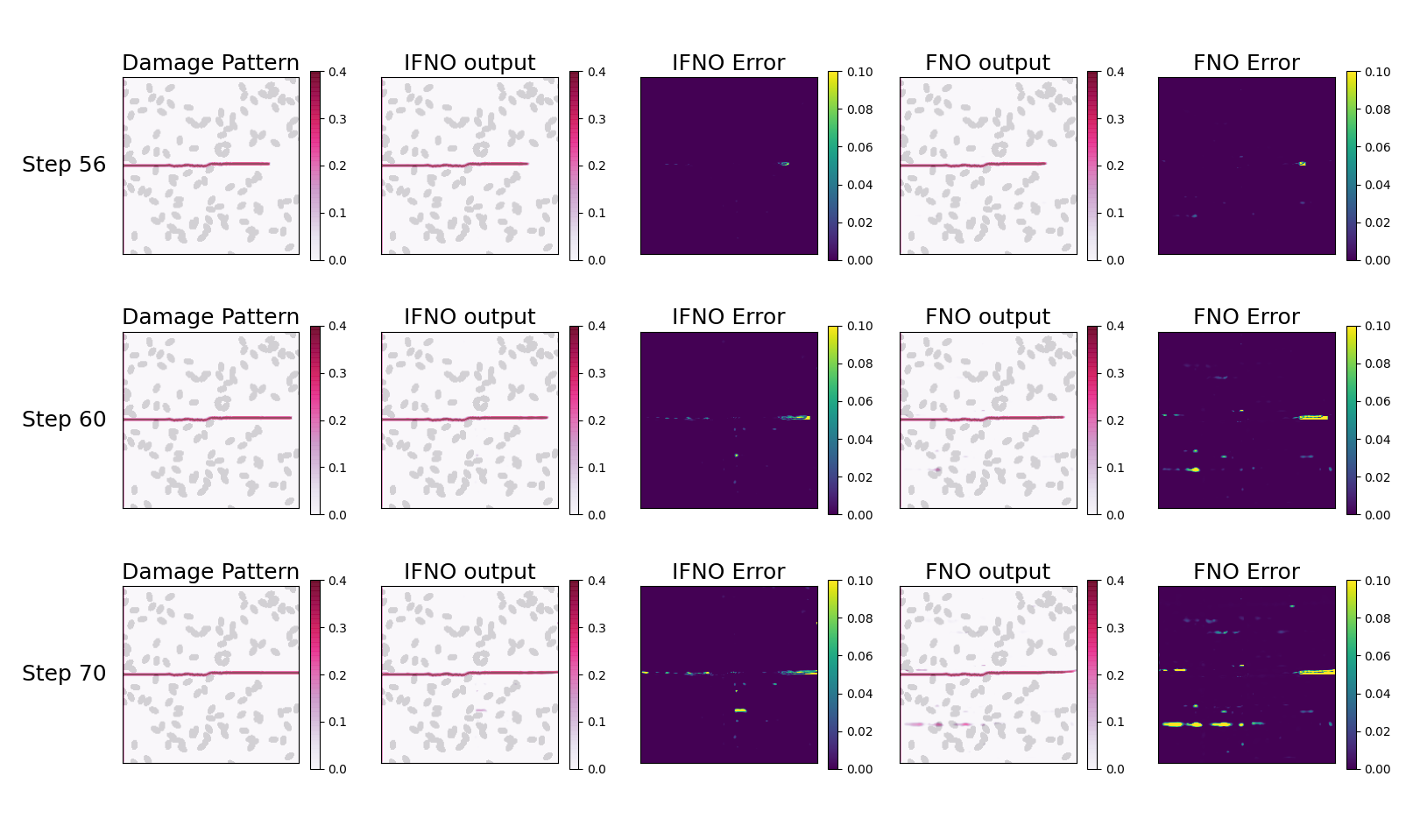}
    \caption{The glass-ceramic crack propagation problem (example 3).  A visualization of FNO and IFNO performances on the 1st, 5th and 15th prediction steps. Here, the best IFNO results ($L=8$) and best FNO results ($L=4$) are reported.}
    \label{fig:lps_test}
\end{figure}

\paragraph{Setting and results of the glass-ceramics fracture problem} 
In this context, our goal is to learn the solution operator of the quasi-static LPS equation, and compute the damage field for a given plate. As shown in Figure \ref{fig:glasssetup}, we considered a plate with a fixed microstructure field $R(\xb)$, and the goal is to estimate the evolution of crack in the left half of this plate, by predicting the damage field $d(\xb,\tau)$ subject to increasing Dirichlet-type boundary loadings $U_D(\tau)$. That means, the neural operator were employed to learn the mapping from $\fb(\xb):=[\xb,d(\xb,\tau-\Delta\tau),\tilde{U}_D(\tau)]$ to {$d(\xb,\tau$)}, where $d(\xb,\tau-\Delta\tau)$ stands for the damage field corresponding to the last quasi-static loading step. With this setting, we aim to predict the crack propagation of one particular material sample under a new and unseen loading scenario. In particular, we generate 70 numbers of samples corresponding to $U_D(\tau)\in[0.12\,\mu$m$,0.26\,\mu$m$]$ with an increment of $\Delta U_D=0.002\,\mu$m for each quasi-static step, such that the first 55 steps/samples (corresponding to $U_D(\tau)\in[0.12\,\mu$m$,0.23\,\mu$m$]$) are employed for training, and the last 15 steps/samples (corresponding to $U_D(\tau)\in[0.232\,\mu$m$,0.26\,\mu$m$]$) are for testing. This setting reflects to the material defect monitoring scenario where some cracks are detected on a material sample with a unknown microstructure, and the learning goal is to predict and monitor the future crack growth. Therefore, this is an out-of-distribution test problem: the longer the prediction period (corresponding to larger $U_D$) is, the harder the prediction task will be. 
%\end{enumerate}
{For the purpose of training, we choose the loss function as the accumulated error of the damage field $d(\xb,\tau)$ within five successive quasi-static steps. Specifically, we used the neural operator to map $[\xb,d(\xb,\tau-\Delta \tau),\tilde{\Ub}_D(\tau)]$ to $\overline{d}(\xb,\tau)$, then used $[\xb,\overline{d}(\xb,\tau),\tilde{\Ub}_D(\tau+\Delta \tau)]$ as the input to obtain $\overline{d}(\tau +\Delta \tau )$, and repeat till an approximated damage field for the next five steps are obtained. Then, we train the network by minimizing the averaged error of $\overline{d}(\xb,\tau+k\Delta \tau)$, $k=0,\cdots,4$. A similar setting can be found, e.g., in \cite{li2020fourier}. In this example, a structured $200\times200$ grid is employed as the discretization in our neural operators. For both FNO and IFNO, we set the dimension of $\hb$ as $48$, and the truncated fourier modes $k = 20\times20$. For each depth $L$, we trained the neural network for $500$ epochs with a learning rate of $5e-3$, then the learning rate was decreased by $0.5$ every $100$ epochs.}

In Figure \ref{fig:loss_lps} we report the averaged relative mean squared errors as a function of iterative layer number $L$. In particular, we show the averaged prediction errors for a relatively short term (over $5$ prediction steps) and longer term (over $15$ prediction steps), respectively. When considering the short term prediction error, we observe that as we increase $L$ from $1$ to $8$, the prediction error from IFNO has a monotonic and drastic decrease from $20.1\%$ to $6.8\%$, reaching a similar level as the averaged training error. Hence, in this example using deeper layer is necessary to obtain a sufficiently expressive IFNO. In contrast, FNO reaches its best performance at $L=4$, and obtains only $14.2\%$ prediction error. For the longer term prediction error, a $12.8\%$ averaged prediction error is obtained for IFNO at $L=8$, while the error from the best FNO is only $25.7\%$. Similarly to previous examples, %while the test error in IFNO monotonically and significantly decreases as the network gets deeper, the FNO reaches its best performance at $L=4$. 
when increasing $L$ the performance of FNOs starts to get polluted by the network instability issue caused by overfitting and vanishing gradient problems, which limits FNOs performance on deeper layers. In Figure \ref{fig:lps_test}, we show the predicted damage fields obtained with the best IFNO and the best FNO, in correspondence of the 1$^{\rm st}$, 5$^{\rm th}$ and 15$^{\rm th}$ prediction steps, which are the 56$^{\rm th}$, 60$^{\rm th}$ and 70$^{\rm th}$ steps among all samples. Both the solutions and the errors are plotted, showing that the FNO starts to mistakenly predict a subcrack since the 1st prediction step, and this crack grows over time, eventually leads to the large long-term prediction error in FNOs. From this example, we find that developing a stable and deep NN is important, especially for a complex learning task like the heterogeneous material damage problem.

\section{Application: Learning From Digital Image Correction (DIC) Measurements}\label{sec:dic}

\begin{figure}
    \centering
\subfigure{\includegraphics[width=.35\columnwidth]{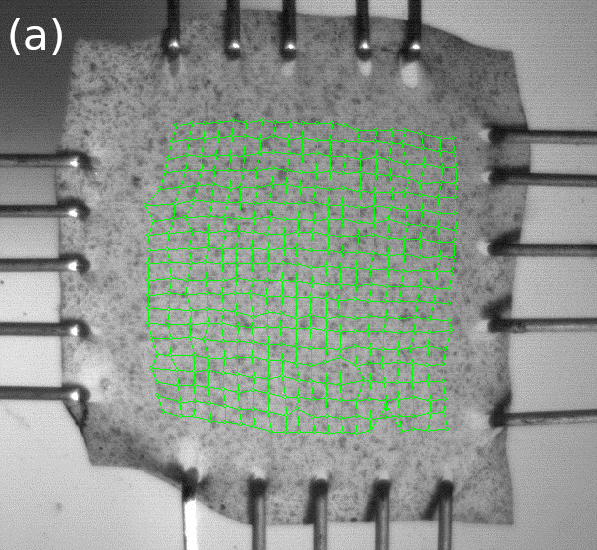}}\qquad\qquad
\subfigure{\includegraphics[width=.35\columnwidth]{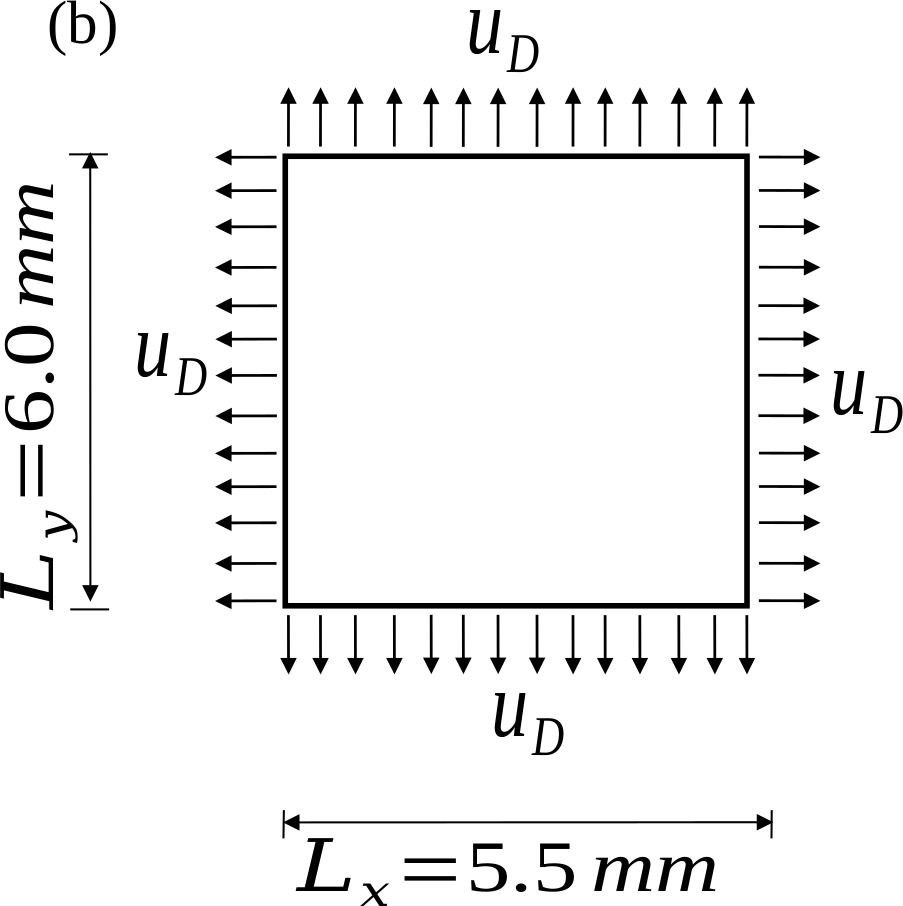}}
    \caption{
    Problem setup of the DIC data acquisition in the latex glove sample modeling problem. (a) An image of the speckle-patterned specimen subject to biaxial stretch loading. (b) A sample subject to Dirichlet-type boundary conditions, as the corresponding numerical setting of (a).}
    \label{fig:dic_setting}
\end{figure}

Having illustrated the performances of our learned neural operators on high-fidelity synthetic simulation datasets in Section \ref{sec:experiments}, we now consider a problem of learning the material response of a latex glove sample from DIC displacement tracking measurements as a prototypical exemplar. The main objective of this section is to provide a proof-of-principle demonstration that the framework introduced thus far applies to learning tasks where the constitutive equations and material microstructure are both unknown, and the dataset has unavoidable measurement noise. Besides the FNOs, in this application we further compare our proposed IFNO against two conventional approaches that use constitutive modeling with parameter fitting to demonstrate the advantages of neural operator models and the importance of considering the heterogeneity of material microstructures.
%In the following, we first introduce the experimental data acquisition in Section \ref{sec:DICexp} and the constitutive models for comparisons with the proposed neural operator learning method in Section \ref{sec:conventional}. Then in Section \ref{sec:dic_result} we illustrate the prediction results of IFNO and three baseline methods.

\subsection{Digital Image Correction (DIC) and biaxial mechanical testing}\label{sec:DICexp}

In this section, we first introduce the experimental sample and data acquisition procedure. For material sample acquisition, the central, palm region of a standard nitrile glove (Dealmed, NY, USA) was sectioned into a $7.5$ mm $\times 7.5$ mm specimen. Then, an optics-based laser thickness measurement device (Keyence, IL, USA) was used to measure the thickness of the specimen before application of a speckle pattern. Following the common procedures from previous research works \cite{zhang2004applications,lionello2014practical,palanca2016use}, %we used an airbrush filled with $2-3$\,mL of India ink to generate a random speckling texture on the surface of the specimen (20-30~cm operating distance, 3 spraying passes).
we used an airbrush to generate a random speckling texture on the surface of the specimen. Then, speckle-patterned specimens were mounted to a biaxial mechanical testing device (CellScale Biomaterials Testing Co., Canada) using five BioRake tines that pierced the specimen at each edge (Figure \ref{fig:dic_setting}(a)). Biaxial characterizations of the specimen was conducted with 3 loading/unloading cycles, targeting an arbitrary force of 750 mN in each direction. Throughout the test, the load cell force readings and actuator positions were recorded at a frequency of 5 Hz, which were subsequently used to calculate the stresses and the stretches for the constitutive model fitting approach as one of the baselines. Meanwhile, a CCD camera captured images throughout the biaxial test at a frequency of 5 Hz. The recorded images were tracked using the digital image correlation (DIC) module of the CellScale LabJoy software. The central $6$ mm $\times 5.5$ mm region of the specimen was selected for the DIC tracking, as the speckling pattern was more random and less susceptible to tracking errors. A $20\times20$ node grid was constructed, and the tracked coordinates were exported.

% \begin{figure}
%     \centering
% \subfigure{\includegraphics[width=.4\columnwidth]{./figures/dicunsmooth_train_fnoboth.jpg}}\quad
% \subfigure{\includegraphics[width=.4\columnwidth]{./figures/dicsmooth_train_fnoboth.jpg}}
%     \caption{Application: a latex glove sample modeling from DIC measurements. Sample-wise error comparison of each model. Left: results for the original dataset. Right: results for the smoothed dataset}
%     \label{fig:loss_dic}
% \end{figure}

% \begin{figure}
%     \centering
% \subfigure{\includegraphics[width=.48\columnwidth]{./figures/unsmooth_sample_error.eps}}
% \subfigure{\includegraphics[width=.48\columnwidth]{./figures/smooth_sample_error.eps}}
%     \caption{Application: a latex glove sample modeling from DIC measurements. Sample-wise error comparison of each model. Left: results for the original dataset. Right: results for the smoothed dataset}
%     \label{fig:dic_samplewise}
% \end{figure}

Based on the tracked coordinates, we constructed two datasets: (i) an original dataset obtained directly from the experimental measurement, and (ii) a smoothed dataset where a moving least-squares (MLS) algorithm was used to calculate the smoothed nodal displacements. To generate the displacement field $\ub^{ori}(\xb)$ for original samples, we subtracted each material point location with its initial location on the first sample, and the boundary displacement loading was obtained by restricting $\ub^{ori}(\xb)$ on the boundary nodes. To create a structured grid for FNOs and IFNOs, we further applied a cubic spline interpolation to the displacement field on a structured $21\times 21$ node grid. Our goal was then to predict the displacement field in the current loading step, given the displacement on the previous step and the current boundary displacement. To construct the smoothed samples for the $j^{\rm th}$ material point, $\xb_j=(x_j,y_j)$, we employed a two-dimensional MLS shape function $\Psi_{j}$ to reconstruct the smoothed displacement field:
$$\mathbf{u}(x,y)=\sum_{j=1}^{NP}\Psi_{j}(x,y)\mathbf{u}_j=\sum_{j=1}^{NP}\phi(x-x_j,y-y_j;w) \mathbf{H}^{T}(0,0)\mathbf{M}^{-1}(x,y)\mathbf{H}(x-x_j,y-y_j)\mathbf{u}_j,$$
where $\mathbf{u}_j=[u_{xj}, u_{yj}]^T$ is the displacement vector of the j$^{\rm th}$ point, $\phi(x,y;w)$ is the window function with a support of $w$, $\mathbf{H}(x,y)=[1,x,y]^T$ is the monomial basis function of linear order,  $\mathbf{M}(x,y):=\sum_{k=1}^{NP}\phi_k(x-x_k,y-y_k)\mathbf{H}(x-x_k,y-y_k)\mathbf{H}^{T}(x-x_k,y-y_k)$ is the moment matrix, and $NP$ is the set of discrete points used to represent the region of interest \cite{belytschko1996meshless,chen1996reproducing}. 

For this study, we chose $NP=9$, a cubic B-spline function with a support of $w=5$  for $\phi(x,y;w)$, and a $14\times 14$ query point grid. The MLS shape functions were used to obtain the smoothed nodal displacements $\ub^{sm}$. Both the smoothed and the original datasets have $877$ total time instants (samples), denoted as $\mcD^{sm}=\{(\ub_D)^{sm}_j,\ub^{sm}_j\}_{j=1}^{877}$ and $\mcD^{un}=\{(\ub_D)^{ori}_j,\ub^{ori}_j\}_{j=1}^{877}$, respectively. For training and cross-validation, we randomly select $177$ samples from the each dataset as test samples, and use the rest as training samples. On each dataset, these training samples were employed for parameter fitting in the constitutive modeling approaches, and used to train for the best neural operators for the IFNO and FNO. In that context, we used common datasets for the constitutive modeling approaches and neural operator learning approaches, to provide a fair comparison between the two different approaches.

\subsection{Constitutive modeling for comparisons with the IFNO}\label{sec:conventional}

\begin{figure}[h]
    \centering
\subfigure{\includegraphics[width=.48\columnwidth]{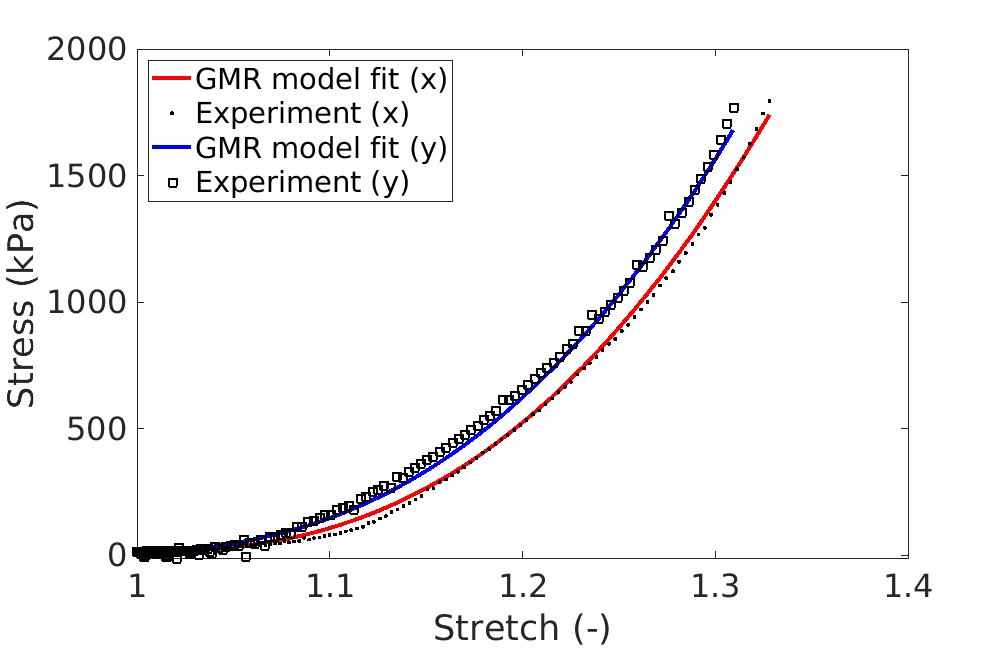}}
    \caption{An illustration of the constitutive model fitting approach which optimizes the generalized Mooney Rivlin (GMR) model parameters from the stress-stretch curve for a latex glove sample.}
    \label{fig:dic_gmr}
\end{figure}

In this section, we provide details for two constitutive modeling approaches, one uses constitutive model fitting to the stress-stretch data and the other uses finite element modeling of the DIC-tracked node displacements, for comparisons with the proposed IFNO method. For both approaches, a generalized Mooney-Rivlin (GMR) hyperelastic model was considered, with its strain energy density function given by:
$$\eta(\overline{I}_1,\overline{I}_2)=c_{10}(\overline{I}_{1}-3) + c_{01}(\overline{I}_{2}-3) + c_{20}(\overline{I}_{1}-3)^{2} + c_{02}(\overline{I}_{2}-3)^{2} + c_{11}(\overline{I}_{1}-3)(\overline{I}_{2}-3).$$
Here, $\overline{I}_{1}=\text{tr}(\mathbf{C})$ and $\overline{I}_{2}=\frac{1}{2}[\text{tr}(\mathbf{C})^2-\text{tr}(\mathbf{C}^2)]$ represent the first and second invariants of the right Cauchy-Green deformation tensor $\mathbf{C}$, and $c_{ij}$ are the model-specific parameters. Based on this pre-assumed constitutive model, we aim to find the optimal parameters of $c_{ij}$ from the training samples, and these parameters will then be used for displacement field predictions on the test samples.

In the first modeling approach, constitutive model parameters were obtained by fitting the final unloading portion of the biaxial stress-stretch data. In particular, the first Piola-Kirchhoff stresses in the $x$- and $y$-directions were determined using the specimen thickness $t$, the undeformed edge lengths $L_x$ and $L_y$, and the measured forces $F_x$ and $F_y$ as $P_{xx}=F_x/tL_y$ and $P_{yy}=F_y/tL_x$. Meanwhile, the stretches in the two directions were calculated as the ratio of the deformed edge lengths to the undeformed length. Both stress-stretch curves in the $x$- and $y$-directions are shown in Figure \ref{fig:dic_gmr}. To obtain the optimal parameters for the GMR model, we used a differential evolution optimization framework to minimize the residual errors in stress predictions between the experimental and model predicted data. Then, using the determined model parameters, finite element modeling was performed using the DIC-tracked nodes and the relative errors of displacement fields are evaluated by comparing the result from this finite element solver and the displacement measurements from DIC. In the following contents, we will refer to this approach as the ``GMR model fitting'' method.

As the second modeling approach, we optimized the constitutive model parameters by minimizing the displacement error from the finite element solver directly. In particular, the structured nodal locations were imported to Abaqus \cite{abaqus2011abaqus} to construct a $21\times 21$ node domain composed of plane stress elements. 
%The GMR models was implemented with a custom user subroutine. 
Then, we solved for the displacement field based on the GMR model using Abaqus, and calculated its relative error with respect to the experimentally-retrieved displacements of each node. The optimal model parameters were obtained by minimizing the total relative displacement error on all training samples. In the following contents, we refer to this approach as the ``GMR inverse analysis'' method.

\subsection{Results and discussion}\label{sec:dic_result}

\begin{figure}[h!]
    \centering
\subfigure{\includegraphics[width=1.\columnwidth]{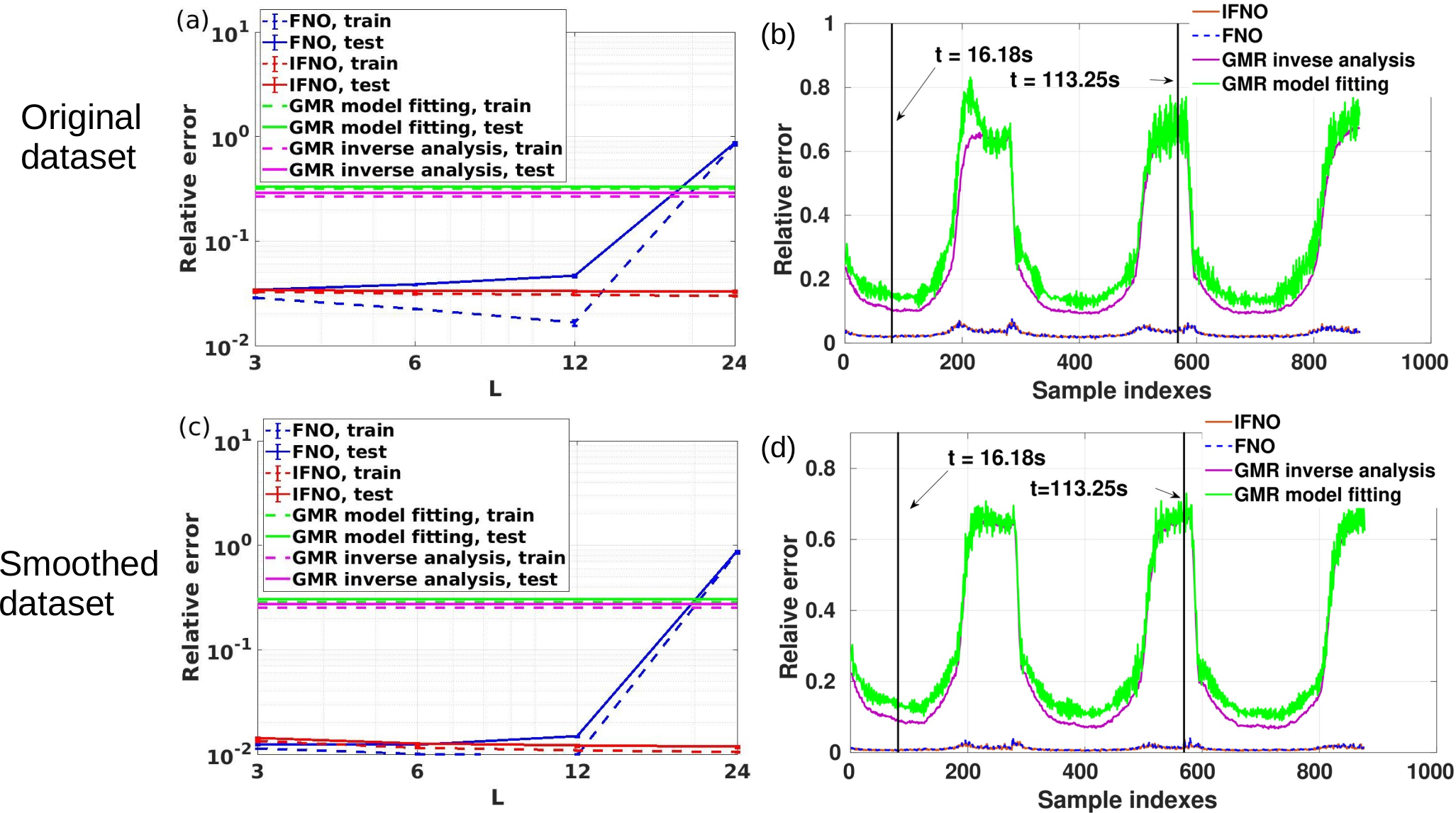}}
    \caption{A latex glove sample modeling from DIC measurements. Error comparisons of each model. Upper plots: results from the original dataset. Bottom plots: results from the smoothed dataset. Left column: relative mean squared errors for quasi-static displacement field prediction on the training and test datasets. Right column: sample-wise error comparison on all samples.}
    %Left: results for the original dataset. Right: results for the smoothed dataset}
    \label{fig:loss_dic}
\end{figure}

\begin{figure}[h!]
    \centering
    \includegraphics[width=.8\textwidth]{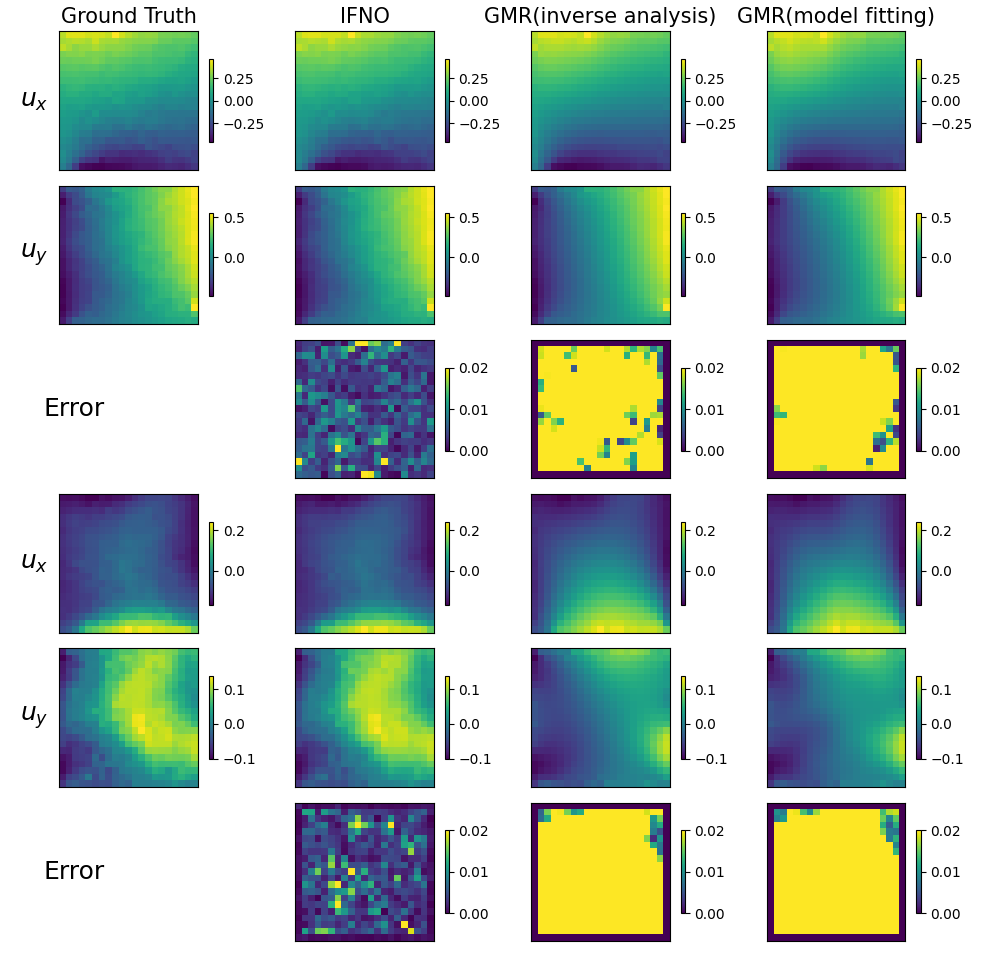}
    \caption{A latex glove sample modeling from DIC measurements. A visualization of GMR and IFNO performances on a test sample in the {\bf original} dataset.}
    \label{fig:unsmoothdata}
\end{figure}

\begin{figure}[h!]
    \centering
    \includegraphics[width=.8\textwidth]{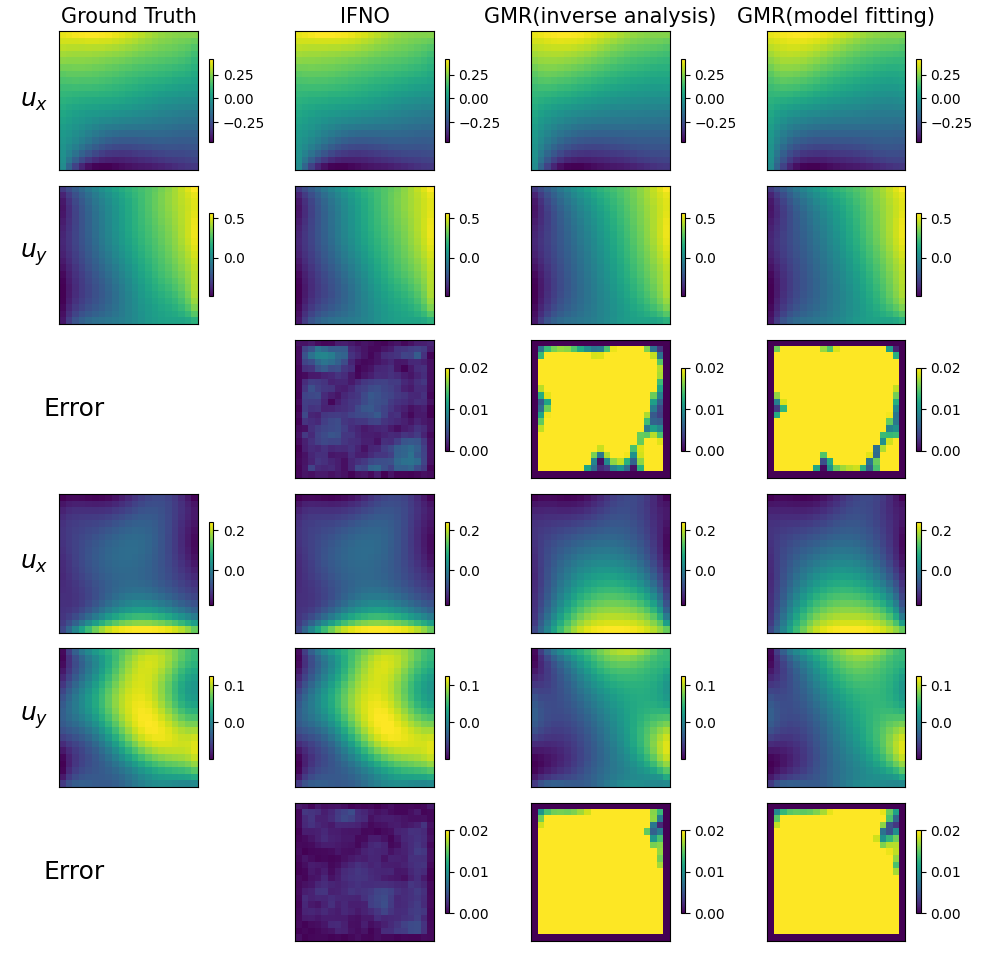}
    \caption{A latex glove sample modeling from DIC measurements. A visualization of GMR and IFNO performances on a test sample in the {\bf smoothed} dataset.}
    \label{fig:smoothdata}
\end{figure}

In this section, we introduce the settings of our neural operator learning models and report the comparison results. Because the time instance between two subsequent loading steps is relatively long, we employed a quasi-static model. In this context, we aim to predict the displacement field $\ub(\xb)$ based on a given boundary displacement loading $\ub_D(\xb)$ and the displacement field from the last loading step (denoted as $\ub^{last}(\xb)$). Therefore, the neural operators were employed to learn the mapping from $\fb(\xb):=[\xb,\ub^{last}(\xb),\tilde{\ub}_D(\xb)]$ to $\ub(\xb)$, where $\tilde{\ub}_D$ is the zero-padded boundary condition, as described in \eqref{eqn:pad}. {In this example, for both the FNOs and IFNOs, we set the dimension of $\hb$ as $d=16$, and the number of truncated Fourier modes as $k = 8\times8$. For each depth $L$, we train the neural network for 1,000 epochs with a learning rate of $1e-3$, then decrease the learning rate with a ratio of 0.7 every 100 epochs.}

In Figure \ref{fig:loss_dic}, we report the relative mean squared errors from both the original dataset (see Figure \ref{fig:loss_dic}(a)) and the smoothed dataset (see Figure \ref{fig:loss_dic}(c)), as functions of the number of hidden layers $L$ from $3$ to $24$. The sample-wise error for each model are also provided in Figure \ref{fig:loss_dic}(b) for the original dataset and in Figure \ref{fig:loss_dic}(d) for the smoothed dataset. Unsurprisingly, when comparing the results from the original dataset and the smoothed dataset, one can observe that the smoothing procedure improves the prediction accuracy for all models. That is because the DIC measurements may contain noise-induced errors, and the nonlocal smoothing procedure we employed performs as an effective filter \cite{lehoucq2015novel} for the measurement noise. When comparing the prediction accuracy from different models, similar to the previous examples, the FNO suffers from overfitting and vanishing gradient issues when $L>2$, especially in the original (more noisy) dataset. This finding is consistent with the results reported in \cite{lu2021comprehensive,kissas2022learning}, where the performance of the FNOs was found to be deteriorated on noisy datasets. In contrast, the accuracy of the IFNOs monotonically improves with the increase of $L$. Both neural operator models outperforms the conventional constitutive modeling approaches by around one order of magnitude. Among all the models, the deep IFNO ($L=24$) performs the best in both datasets. On the original dataset which features noise, it achieves a $3.3\%$ prediction error. On the smoothed dataset, the IFNO has an $1.18\%$ prediction error. On the other hand, the GMR model fitting and GMR inverse analysis approaches have obtained $33.0\%$ and $29.1\%$ prediction errors on the original dataset, respectively. On the smoothed dataset, the prediction error for these two GMR models are slightly smaller, as $30.5\%$ and $27.3\%$, respectively. To provide further insights into this comparison, in Figures \ref{fig:unsmoothdata}-\ref{fig:smoothdata} we depict both solutions and prediction errors obtained with the best IFNO and the two GMR models on two test samples which correspond to the large deformation ($t=113.25\,s$) and small deformation ($t=16.18\,s$) representatives, respectively. From the ground-truth data pattern of $u_y(\xb)$, we can see that the glove sample is in fact heterogeneous, since a large deformation region is observed in the middle of the sample. Both GMR models fail to capture the material heterogeneity and hence obtained large prediction errors. This observation again confirms the importance of capturing the material heterogeneity and verifies the capability of IFNOs in heterogeneous material modeling.

\section{Conclusion}\label{sec:conclusion}

With the objective of predicting material responses under unseen loading conditions, in this work we have proposed a novel data-driven computing paradigm for material modeling, which integrates material identification, modeling procedures, and material responses prediction into one unified learning framework. In particular, a data-driven model has been developed, which learns the mapping from loading conditions to the corresponding material responses as a solution operator. To this end, a new integral neural operator has been proposed, which we refer to as the implicit Fourier neural operator (IFNO). In the IFNO, the increment between layers are modeled by integral operators, so the resultant architecture can be interpreted as a fixed point method for the unknown governing laws. Furthermore, by identifying its layers with time instants, the IFNO can be reinterpreted as time-dependent equations, which enables the use of efficient initialization techniques that enhances the network stability in the deep layer limit. Our results have shown that, in all learning tasks, the IFNOs outperform baseline methods in stability and prediction accuracy for unseen loading conditions. Both the universal approximation theorem and numerical results demonstrate that, in complex learning tasks, a stable deep layer architecture is necessary to achieve a satisfactory prediction accuracy. Last but not least, we have, \textit{for the first time}, leveraged the application of neural operators to learning the material responses directly from DIC displacement tracking measurements, where the constitutive equations and material microstructure are both unknown, and measurement noise is present. Numerical results have confirmed the advantage of neural operator learning approaches against the conventional constitutive modeling approaches: the former does not require a pre-assumed material model, and is able to capture the material heterogeneity. Hence, the proposed neural operator models have outperformed the conventional generalized Mooney Rivlin (GMR) model in prediction accuracy by at least one order of magnitude. When comparing with another neural operator model, i.e., the FNOs, our proposed IFNOs have been shown to be less prone to the overfitting issue and hence achieve a better performance on noisy experimental datasets.

Although the IFNO requires a much smaller number of trainable parameters comparing with its counterpart, FNOs, we did not observe a decrease of the computational time because the fixed point procedure of the IFNO comes with the price of using an iterative algorithm. Therefore, an important next step is to combine the IFNO with faster training techniques of implicit networks \cite{fung2021jfb} to improve its efficiency. Moreover, we point out that the IFNO provides a general and flexible solution operator for unknown governing laws, which is not restricted to material modeling tasks. As another natural extension, we will consider the application of the IFNO on other complex learning tasks, such as image classification problems. The implicit neural operator architecture we proposed here can also be combined with other recent integral neural operator architectures, e.g., the multiwavelet-based operator \cite{gupta2021multiwaveletbased} and the integral autoencoder-based network (IAE-Net) \cite{Ong2022}, which would be another interesting future direction.

\section*{Acknowledgements}

The authors would like to thank Mr. Minglang Yin for sharing his FEniCS codes and for the helpful discussions. H. You and Y. Yu would like to acknowledge support by the National Science Foundation under award DMS 1753031. Portions of this research were conducted on Lehigh University's Research Computing infrastructure partially supported by NSF Award 2019035. We also thank the Presbyterian Health Foundation Team Science Grant, and the National Science Foundation Graduate Research Fellowship Program (GRF2020307284).

\appendix
\renewcommand{\thefigure}{A\arabic{figure}}

\setcounter{figure}{0}

\renewcommand{\thetable}{A\arabic{table}}

\setcounter{table}{0}

\section{Detailed Numeric Results}\label{sec:newapp_pde}

In this section we provide the detailed numerical results of each task in Sections \ref{sec:experiments}-\ref{sec:dic}, as the supplementary results of the training and test errors plotted in Figures \ref{fig:loss_2ddarcy_16}, \ref{fig:loss_hgo}, \ref{fig:loss_lps} and \ref{fig:loss_dic} of the main text. The full results for porous medium pressure field learning I, porous medium pressure field learning II, fiber-reinforced material displacement field learning, glass-ceramics damage field learning, and DIC measurements of latex glove displacement filed learning are provided in Tables \ref{tab:2DDarcy_1}, \ref{tab:2DDarcy_2}, \ref{tab:hgo}, \ref{tab:lps} and \ref{tab:dic}, respectively. {To reduce the impact of initialization in neural operator models, for each task we run five simulations for each network using different random seeds, and report the mean and the standard error among these five simulations.} For each model, we use the bold case to highlight the architecture with the best prediction accuracy.

\begin{table}[ht!]
\small
    \centering
    \begin{tabular}{|>{\hspace{-4pt}}c<{\hspace{-4pt}}|>{\hspace{-4pt}}c<{\hspace{-4pt}}|>{\hspace{-4pt}}c<{\hspace{-4pt}}|>{\hspace{-4pt}}c<{\hspace{-4pt}}|>{\hspace{-4pt}}c<{\hspace{-4pt}}|>{\hspace{-4pt}}c<{\hspace{-4pt}}|>{\hspace{-4pt}}c<{\hspace{-4pt}}|>{\hspace{-4pt}}c<{\hspace{-4pt}}|}
    \hline
    \multicolumn{2}{|>{\hspace{-4pt}}c<{\hspace{-4pt}}|}{Model/dataset} & $L=1$ & $L=2$ & $L=4$ & $L=8$ & $L=16$ & $L=32$\\
    \hline 
    \multirow{2}{*}{IFNO} &train& 1.67e-2$\pm$1.16e-4 & 7.79e-3$\pm$5.58e-5 & 6.48e-3$\pm$6.16e-5 & 5.84e-3$\pm$6.58e-5 & 5.46e-3$\pm$6.79e-5 & 5.21e-3$\pm$6.98e-5 \\
    &test & 1.77e-2$\pm$1.18e-4 & 1.23e-2$\pm$9.46e-5 & 1.10e-2$\pm$6.90e-5 & 1.05e-2$\pm$5.72e-5 & 1.04e-2$\pm$3.72e-5 & {\bf 1.02e-2$\pm$5.77e-5} \\
    \hline 
    \multirow{2}{*}{FNO} &train& 1.65e-2$\pm$4.94e-5 & 4.13e-3$\pm$3.16e-4 & 7.94e-3$\pm$1.65e-5 & 6.83e-4$\pm$5.09e-6 & 8.34e-4$\pm$1.55e-5 & 2.84e-1$\pm$2.57e-6 \\
    &test & 1.76e-2$\pm$9.40e-5 & 1.30e-2$\pm$5.11e-5 & {\bf 1.19e-2$\pm$8.98e-5} & 1.56e-2$\pm$1.85e-4 & 2.80e-2$\pm$1.19e-3 & 2.90e-1$\pm$1.07e-4\\
    \hline 
    \end{tabular}
    \caption{Numerical results for the learning task of porous medium I. Bold numbers highlight the case with the best error for each model.}
     \label{tab:2DDarcy_1}  
\end{table}

\begin{table}[ht!]
\footnotesize
    \centering
    \begin{tabular}{|>{\hspace{-4pt}}c<{\hspace{-4pt}}|>{\hspace{-4pt}}c<{\hspace{-4pt}}|>{\hspace{-4pt}}c<{\hspace{-4pt}}|>{\hspace{-4pt}}c<{\hspace{-4pt}}|>{\hspace{-4pt}}c<{\hspace{-4pt}}|>{\hspace{-4pt}}c<{\hspace{-4pt}}|>{\hspace{-4pt}}c<{\hspace{-4pt}}|>{\hspace{-4pt}}c<{\hspace{-4pt}}|>{\hspace{-4pt}}c<{\hspace{-4pt}}|}
    \hline 
    \multicolumn{2}{|>{\hspace{-4pt}}c<{\hspace{-4pt}}|}{Model/dataset} &  $L=1$ & $L=2$ & $L=4$ & $L=8$ & $L=16$ & $L=32$ & $L=64$ \\
    \hline
    \multirow{2}{*}{IFNO} &train& 9.81e-3$\pm$9.90e-5 & 4.38e-3$\pm$9.30e-5 & 3.93e-3$\pm$7.19e-5 & 3.89e-3$\pm$7.60e-5 & 3.90e-3$\pm$8.36e-5 & 3.98e-3$\pm$9.57e-5 &3.93e-3$\pm$1.01e-4 \\
    &test & 1.10e-2$\pm$1.16e-4 & 5.75e-3$\pm$1.17e-4 & 5.23e-3$\pm$1.05e-4& 5.10e-3$\pm$1.16e-4 & 5.07e-3$\pm$1.51e-4 & 5.04e-3$\pm$1.61e-4 & {\bf 4.89e-3$\pm$2.22e-4}\\
    \hline 
    \multirow{2}{*}{FNO} &train& 1.00e-2$\pm$9.52e-5 & 3.43e-3$\pm$8.54e-5 & 3.27e-3$\pm$8.53e-5 & 3.77e-3$\pm$2.65e-5 & 3.91e-3$\pm$2.54e-5 & 9.86e-1$\pm$2.20e-5 & 9.86e-1$\pm$2.19e-5\\
    &test & 1.13e-2$\pm$1.05e-4 & {\bf 5.26e-3$\pm$8.21e-5} & 6.07e-3$\pm$1.68e-4 & 8.59e-3$\pm$5.14e-5 & 1.26e-2$\pm$3.26e-4 & 9.89e-1$\pm$2.03e-4 & 9.89e-1$\pm$2.03e-4\\
    \hline 
    \end{tabular}
    \caption{Numerical results for the learning task of porous medium II. Bold numbers highlight the case with the best error for each model.}
    \label{tab:2DDarcy_2}
\end{table}

\begin{table}[ht!]
\small
    \centering
    \begin{tabular}{|>{\hspace{-4pt}}c<{\hspace{-4pt}}|>{\hspace{-4pt}}c<{\hspace{-4pt}}|>{\hspace{-4pt}}c<{\hspace{-4pt}}|>{\hspace{-4pt}}c<{\hspace{-4pt}}|>{\hspace{-4pt}}c<{\hspace{-4pt}}|>{\hspace{-4pt}}c<{\hspace{-4pt}}|>{\hspace{-4pt}}c<{\hspace{-4pt}}|>{\hspace{-4pt}}c<{\hspace{-4pt}}|}
    \hline
    \multicolumn{8}{|c|}{Dirichlet boundary condition with in-distribution test}\\
    \hline
    \multicolumn{2}{|>{\hspace{-4pt}}c<{\hspace{-4pt}}|}{Model/dataset} & $L=1$ & $L=2$ & $L=4$ & $L=8$ & $L=16$ & $L=32$\\
    \hline 
    \multirow{2}{*}{IFNO} &train& 7.96e-3$\pm$6.60e-5 & 3.82e-3$\pm$6.10e-5&3.07e-3$\pm$1.36e-4 & 2.68e-3$\pm$6.40e-5 & 2.52e-3$\pm$4.91e-5 & 2.40e-3$\pm$2.42e-5 \\
    &test & 7.70e-3$\pm$1.66e-4 & 4.10e-3$\pm$6.17e-4 & 4.05e-3$\pm$4.25e-4 & 4.86e-3$\pm$7.12e-4 & 2.96e-3$\pm$2.78e-4 & {\bf 2.12e-3$\pm$4.42e-5} \\
    \hline 
    \multirow{2}{*}{FNO} &train& 7.81e-3$\pm$8.77e-5 & 3.65e-3$\pm$8.59e-5 & 3.34e-3$\pm$4.58e-5 &4.98e-3$\pm$7.51e-4 & 4.93e-1$\pm$2.60e-1 & 1.23e0$\pm$2.73e-3 \\
    &test & 7.46e-3$\pm$3.79e-4 & 5.78e-3$\pm$9.90e-4 & {\bf 3.50e-3$\pm$3.26e-4} & 5.24e-3$\pm$4.70e-4 & 4.78e-1$\pm$2.56e-1 & 1.24e0$\pm$3.67e-2\\
    \hline 
    \multicolumn{8}{|c|}{Dirichlet boundary condition with out-of-distribution test}\\
    \hline
    \multicolumn{2}{|>{\hspace{-4pt}}c<{\hspace{-4pt}}|}{Model/dataset} & $L=1$ & $L=2$ & $L=4$ & $L=8$ & $L=16$ & $L=32$\\
    \hline 
    \multirow{2}{*}{IFNO} &train& 7.58e-3$\pm$8.20e-5 & 3.74e-3$\pm$6.09e-5 & 2.92e-3$\pm$6.52e-5 & 2.69e-3$\pm$7.10e-5 & 2.44e-3$\pm$3.88e-5 & 2.14e-3$\pm$2.33e-5 \\
    &test &  6.78e-3$\pm$7.18e-5 & 3.42e-3$\pm$3.44e-4 & 2.95e-3$\pm$1.60e-4 & 2.51e-3$\pm$1.74e-4 & 2.34e-3$\pm$1.23e-4 & {\bf 2.06e-3$\pm$5.92e-5} \\
    \hline 
    \multirow{2}{*}{FNO} &train& 7.61e-3$\pm$2.07e-4 & 3.63e-3$\pm$1.13e-4 & 3.68e-3$\pm$4.46e-4 & 3.89e-3$\pm$1.77e-4 & 2.78e-1$\pm$2.36e-1 & 8.34e-1$\pm$2.44e-1 \\
    &test & 6.56e-3$\pm$2.62e-4 & 3.25e-3$\pm$4.33e-4 & {\bf 2.43e-3$\pm$2.09e-4} & 3.18e-3$\pm$1.98e-4 & 2.33e-1$\pm$1.83e-1 & 7.03e-1$\pm$1.85e-1\\
    \hline 
    \multicolumn{8}{|c|}{Neumann boundary condition}\\
    \hline
    \multicolumn{2}{|>{\hspace{-4pt}}c<{\hspace{-4pt}}|}{Model/dataset} & $L=1$ & $L=2$ & $L=4$ & $L=8$ & $L=16$ & $L=32$\\
    \hline
    \multirow{2}{*}{IFNO} &train& 4.33e-2$\pm$1.26e-4 &  1.42e-2$\pm$2.40e-5 & 1.08e-2$\pm$8.91e-5 & 9.32e-3$\pm$7.34e-5 & 9.03e-3$\pm$4.48e-5 & 9.04e-3$\pm$7.01e-5 \\
    &test & 5.45e-2$\pm$6.35e-4 & 2.15e-2$\pm$1.44e-4 & 1.75e-2$\pm$8.42e-5 & 1.54e-2$\pm$8.60e-5 & 1.44e-2$\pm$1.30e-4 & {\bf 1.41e-2$\pm$3.70e-5} \\
    \hline 
    \multirow{2}{*}{FNO} &train& 4.47e-2$\pm$5.61e-4 & 1.37e-2$\pm$4.88e-5 & 8.18e-3$\pm$7.51e-5 & 7.96e-3$\pm$1.36e-4 & 2.21e-1$\pm$1.78e-1 & 5.81e-1$\pm$1.87e-1 \\
    &test & 5.19e-2$\pm$8.17e-4 & 1.90e-2$\pm$6.55e-4 & {\bf 1.40e-2$\pm$5.80e-3} & 1.42e-2$\pm$3.86e-4 & 2.26e-1$\pm$1.74e-1 & 6.01e-1$\pm$1.76e-1\\
    \hline 
    \end{tabular}
    \caption{Numerical results for the learning task of fiber-reinforced material displacement field. Bold numbers highlight the case with the best error for each model.}
       \label{tab:hgo}  
\end{table}

\begin{table}[ht!]
    \centering
    \begin{tabular}{|>{\hspace{-4pt}}c<{\hspace{-4pt}}|>{\hspace{-4pt}}c<{\hspace{-4pt}}|>{\hspace{-4pt}}c<{\hspace{-4pt}}|>{\hspace{-4pt}}c<{\hspace{-4pt}}|>{\hspace{-4pt}}c<{\hspace{-4pt}}|>{\hspace{-4pt}}c<{\hspace{-4pt}}|}
    \hline
    \multicolumn{2}{|c|}{Model/dataset} & $L=1$ & $L=2$ & $L=4$ & $L=8$ \\
    \hline 
    \multirow{3}{*}{IFNO} &train& 8.01e-2$\pm$9.96e-3  & 8.48e-2$\pm$1.96e-2 & 4.87e-2$\pm$2.05e-3 & 4.48e-2$\pm$1.62e-3 \\
    &5-step test &2.01e-1$\pm$1.58e-2 & 1.60e-1$\pm$3.25e-2 & 7.65e-2$\pm$8.46e-3 & {\bf 6.86e-2$\pm$7.08e-3} \\
    & 15-step test &3.10e-1$\pm$1.54e-2 & 2.51e-1$\pm$3.28e-2 & 1.58e-1$\pm$1.42e-2 & {\bf 1.28e-1$\pm$2.04e-2} \\
    \hline 
    \multirow{3}{*}{FNO} &train& 6.85e-2$\pm$1.85e-3 & 5.52e-2$\pm$9.74e-4 & 4.16e-2$\pm$3.33e-4 & 1.28e-1$\pm$4.11e-3\\
     &5-step test & 2.06e-1$\pm$1.64e-2 & 1.91e-1$\pm$3.41e-2 & {\bf 1.42e-1$\pm$1.34e-2} & 2.36e-1$\pm$3.68e-4 \\
    &15-step test & 3.12e-1$\pm$1.69e-2 & 2.84e-1$\pm$2.72e-2 & {\bf 2.57e-1$\pm$1.85e-2} & 3.30e-1$\pm$1.74e-3\\
    \hline 
    \end{tabular}
    \caption{Numerical results for the learning task of glass-ceramics damage field. Bold numbers highlight the case with the best error for each model.}
    \label{tab:lps}  
\end{table}

\begin{table}[ht!]
\small
    \centering
    \begin{tabular}{|>{\hspace{-4pt}}c<{\hspace{-4pt}}|>{\hspace{-4pt}}c<{\hspace{-4pt}}|>{\hspace{-4pt}}c<{\hspace{-4pt}}|>{\hspace{-4pt}}c<{\hspace{-4pt}}|>{\hspace{-4pt}}c<{\hspace{-4pt}}|>{\hspace{-4pt}}c<{\hspace{-4pt}}|}
    \hline
    \multicolumn{2}{|c|}{Model/dataset} & $L=3$ & $L=6$ & $L=12$ & $L=24$ \\
    \hline 
    \multirow{2}{*}{IFNO, original} &train& 3.26e-2$\pm$1.08e-4 & 3.13e-2$\pm$1.30e-4 & 3.06e-2$\pm$1.08e-4 & 3.00e-2$\pm$1.24e-4 \\
    &test &3.43e-2$\pm$4.96e-4 & 3.34e-2$\pm$4.53e-4 & 3.32e-2$\pm$4.41e-4 & {\bf 3.30e-2$\pm$4.63e-4} \\
    \hline 
    \multirow{2}{*}{FNO, original} &train& 2.88e-2$\pm$1.23e-4 & 2.25e-2$\pm$8.68e-5 & 1.66e-2$\pm$9.94e-4 & 8.47e-1$\pm$4.72e-3\\
     &test & {\bf 3.40e-2$\pm$4.09e-4} & 3.84e-2$\pm$4.21e-4 & 4.66e-2$\pm$1.47e-3 & 8.61e-1$\pm$2.70e-2 \\
     \hline
    \multirow{2}{*}{GMR model fitting, original} &train &\multicolumn{4}{c|}{3.16e-1}  \\
    &test & \multicolumn{4}{c|}{\bf 3.30e-1}  \\
    \hline 
     \multirow{2}{*}{GMR inverse analysis, original} &train &\multicolumn{4}{c|}{2.66e-1}  \\
    &test & \multicolumn{4}{c|}{\bf 2.91e-1}  \\
    \hline
     \multirow{2}{*}{IFNO, smoothed} &train& 1.33e-2$\pm$1.61e-4 & 1.16e-2$\pm$8.10e-5 & 1.09e-2$\pm$4.91e-5 & 1.05e-2$\pm$6.01e-5 \\
    &test &1.43e-2$\pm$2.99e-4 & 1.26e-2$\pm$2.20e-4 & 1.21e-2$\pm$2.28e-4 & {\bf 1.18e-2$\pm$2.21e-4}\\
    \hline 
    \multirow{2}{*}{FNO, smoothed} &train& 1.14e-2$\pm$3.28e-5 & 1.01e-2$\pm$9.28e-5 & 9.83e-3$\pm$3.02e-4 & 8.49e-1$\pm$3.50e-3\\
     &test & 1.25e-2$\pm$2.25e-4 & {\bf 1.23e-2$\pm$1.98e-4} & 1.49e-2$\pm$1.15e-4 & 8.73e-1$\pm$1.87e-2 \\
     \hline
    \multirow{2}{*}{GMR model fitting, smoothed} &train &\multicolumn{4}{c|}{2.87e-1}  \\
    &test & \multicolumn{4}{c|}{\bf 3.05e-1}  \\
    \hline
     \multirow{2}{*}{GMR inverse analysis, smoothed} &train &\multicolumn{4}{c|}{2.52e-1}  \\
    &test & \multicolumn{4}{c|}{\bf 2.73e-1}  \\
    \hline
    \end{tabular}
    \caption{Numerical results for the learning task of DIC measurements of latex glove displacement filed, compared with the generalized Mooney-Rivlin (GMR) model. Bold numbers highlight the case with the best error for each model.}
    \label{tab:dic}  
\end{table}

\bibliographystyle{elsarticle-num}
\bibliography{snl}

\end{document}